\documentclass[10pt, journal]{IEEEtran}

\IEEEoverridecommandlockouts

\usepackage{gensymb}
\usepackage{dsfont}
\usepackage{amsmath,amssymb,amsfonts,amsthm}

\usepackage{epsfig}
\usepackage{cite}
\usepackage{hhline}
\usepackage{multirow}
\usepackage{xcolor}
\usepackage{makecell}
\usepackage{subcaption}
\usepackage{capt-of}
\usepackage[labelformat=simple]{subcaption}

\usepackage{enumerate}
\usepackage{enumitem}
\usepackage{color}
\usepackage{graphicx}
\usepackage{algpseudocode}
\usepackage[ruled]{algorithm}
\makeatletter
\newcounter{parentalgorithm}

\makeatother

\usepackage[nolist,printonlyused]{acronym}      
\usepackage{booktabs}
\usepackage{bm}
\usepackage{pgf}
\usepackage{pgfplots}
\usepackage{tikz}
\usepackage{grffile}
\pgfplotsset{compat=newest}
\usetikzlibrary{plotmarks}
\usetikzlibrary{shapes.geometric,quotes,angles,automata,arrows,positioning,calc,3d,matrix,decorations.markings,matrix}

\tikzstyle{startstop} = [rectangle, rounded corners, minimum width=3cm, minimum height=1cm,text centered, draw=black, fill=white!30]
\tikzstyle{io} = [trapezium, trapezium left angle=70, trapezium right angle=110, minimum width=3cm, minimum height=1cm, text centered, draw=black, fill=white!30]
\tikzstyle{process} = [rectangle, minimum width=2cm, minimum height=1cm, text centered, draw=black, fill=white!30]
\tikzstyle{decision} = [diamond, minimum width=3cm, minimum height=1cm, text centered, draw=black, fill=white!30]
\tikzstyle{arrow} = [thick,->,>=stealth]

\makeatletter
\newcommand{\vast}{\bBigg@{3}}
\newcommand{\Vast}{\bBigg@{4}}
\makeatother
\renewcommand{\IEEEQED}{\IEEEQEDopen}

\showoutput
\makeatletter
\tikzset{%
  remember picture with id/.style={%
    remember picture,
    overlay,
    save picture id=#1,
  },
  save picture id/.code={%
    \edef\pgf@temp{#1}%
    \immediate\write\pgfutil@auxout{%
      \noexpand\savepointas{\pgf@temp}{\pgfpictureid}}%
  },
  if picture id/.code args={#1#2#3}{%
    \@ifundefined{save@pt@#1}{%
      \pgfkeysalso{#3}%
    }{
      \pgfkeysalso{#2}%
    }
  }
}

\def\savepointas#1#2{%
  \expandafter\gdef\csname save@pt@#1\endcsname{#2}%
}

\def\tmk@labeldef#1,#2\@nil{%
  \def\tmk@label{#1}%
  \def\tmk@def{#2}%
}

\tikzdeclarecoordinatesystem{pic}{%
  \pgfutil@in@,{#1}%
  \ifpgfutil@in@%
    \tmk@labeldef#1\@nil
  \else
    \tmk@labeldef#1,(0pt,0pt)\@nil
  \fi
  \@ifundefined{save@pt@\tmk@label}{%
    \tikz@scan@one@point\pgfutil@firstofone\tmk@def
  }{%
  \pgfsys@getposition{\csname save@pt@\tmk@label\endcsname}\save@orig@pic%
  \pgfsys@getposition{\pgfpictureid}\save@this@pic%
  \pgf@process{\pgfpointorigin\save@this@pic}%
  \pgf@xa=\pgf@x
  \pgf@ya=\pgf@y
  \pgf@process{\pgfpointorigin\save@orig@pic}%
  \advance\pgf@x by -\pgf@xa
  \advance\pgf@y by -\pgf@ya
  }%
}

\makeatother

\usepackage{comment}
\usepackage{xpatch}
\makeatletter
\xpatchcmd{\algorithmic}{\itemsep\z@}{\itemsep=-0.25mm}{}{}
\makeatother

\usepackage{matlab-prettifier}

\usepackage{multicol}
\usepackage{multirow}

\usepackage{lipsum} 
\usepackage{multicol}
\usepackage{amsmath}
\usepackage{algorithm}

\algrenewcommand\algorithmicforall{\textbf{foreach}}
\algrenewcommand\algorithmicindent{.8em}

\usepackage{mathtools}

\DeclarePairedDelimiter\floor{\lfloor}{\rfloor}

\usepackage{pgfplots}
\usetikzlibrary{pgfplots.groupplots}

\usepackage{colortbl}
\usepackage{colortbl}  
\usepackage{xcolor}    
\usepackage{multirow}  

\usepackage{tikz}
\usetikzlibrary{shapes.geometric, arrows, positioning, fit, backgrounds}

\tikzstyle{process} = [rectangle, rounded corners, minimum width=1cm, minimum height=1.2cm, text centered, draw=black, fill=white]
\tikzstyle{server} = [rectangle, rounded corners, minimum width=2cm, minimum height=1.2cm, text centered, draw=black, fill=yellow!30]
\tikzstyle{arrow} = [thick,->,>=stealth]

\begin{document}

\title{ Accelerating Energy-Efficient Federated Learning in Cell-Free Networks with Adaptive Quantization}

\author{Afsaneh Mahmoudi, Ming Xiao, Emil Björnson
\thanks{ School of Electrical Engineering and Computer Science, KTH Royal Institute of Technology, Stockholm, Sweden. \{afmb, mingx, emilbjo\}@kth.se }
\thanks{The work was supported by the SUCCESS grant from the Swedish Foundation for Strategic Research.}}



\newcommand{\n}{\nabla}
\newcommand{\nrm}[1]{\left \| #1 \right \|}

\newcommand{\E}{\mathds{E}}
\newcommand{\inner}[1]{\left\langle #1 \right \rangle}
\newtheorem{theorem}{Theorem}
\newtheorem{defin}{Definition}
\newtheorem{prop}{Proposition}
\newtheorem{lemma}{Lemma}
\newtheorem{corollary}{Corollary}
\newtheorem{alg}{Algorithm}
\newtheorem{remark}{Remark}
\newtheorem{result}{Result}
\newtheorem{conjecture}{Conjecture}
\newtheorem{example}{Example}
\newtheorem{notations}{Notations}
\newtheorem{assumption}{Assumption}
\newcommand{\combin}[2]{\ensuremath{ \left( \ba{c} #1 \\ #2 \ea \right) }}
\newcommand{\diag}{{\mbox{diag}}}
\newcommand{\rank}{{\mbox{rank}}}
\newcommand{\dom}{{\mbox{dom{\color{white!100!black}.}}}}
\newcommand{\range}{{\mbox{range{\color{white!100!black}.}}}}
\newcommand{\image}{{\mbox{image{\color{white!100!black}.}}}}
\newcommand{\herm}{^{\mbox{\scriptsize H}}}  
\newcommand{\sherm}{^{\mbox{\tiny H}}}       
\newcommand{\tran}{^{\mbox{\scriptsize T}}}  
\newcommand{\tranIn}{^{\mbox{-\scriptsize T}}}  
\newcommand{\card}{{\mbox{\textbf{card}}}}
\newcommand{\asign}{{\mbox{$\colon\hspace{-2mm}=\hspace{1mm}$}}}
\newcommand{\ssum}[1]{\mathop{ \textstyle{\sum}}_{#1}}

\newcommand{\vbar}{\raisebox{.17ex}{\rule{.04em}{1.35ex}}}
\newcommand{\vbarind}{\raisebox{.01ex}{\rule{.04em}{1.1ex}}}
\newcommand{\D}{\ifmmode {\rm I}\hspace{-.2em}{\rm D} \else ${\rm I}\hspace{-.2em}{\rm D}$ \fi}
\newcommand{\T}{\ifmmode {\rm I}\hspace{-.2em}{\rm T} \else ${\rm I}\hspace{-.2em}{\rm T}$ \fi}
\newcommand{\B}{\ifmmode {\rm I}\hspace{-.2em}{\rm B} \else \mbox{${\rm I}\hspace{-.2em}{\rm B}$} \fi}
\newcommand{\Hil}{\ifmmode {\rm I}\hspace{-.2em}{\rm H} \else \mbox{${\rm I}\hspace{-.2em}{\rm H}$} \fi}
\newcommand{\C}{\ifmmode \hspace{.2em}\vbar\hspace{-.31em}{\rm C} \else \mbox{$\hspace{.2em}\vbar\hspace{-.31em}{\rm C}$} \fi}
\newcommand{\Cind}{\ifmmode \hspace{.2em}\vbarind\hspace{-.25em}{\rm C} \else \mbox{$\hspace{.2em}\vbarind\hspace{-.25em}{\rm C}$} \fi}
\newcommand{\Q}{\ifmmode \hspace{.2em}\vbar\hspace{-.31em}{\rm Q} \else \mbox{$\hspace{.2em}\vbar\hspace{-.31em}{\rm Q}$} \fi}
\newcommand{\Z}{\ifmmode {\rm Z}\hspace{-.28em}{\rm Z} \else ${\rm Z}\hspace{-.38em}{\rm Z}$ \fi}

\newcommand{\sgn}{\mbox {sgn}}
\newcommand{\var}{\mbox {var}}
\newcommand{\cov}{\mbox {cov}}
\renewcommand{\Re}{\mbox {Re}}
\renewcommand{\Im}{\mbox {Im}}
\newcommand{\cum}{\mbox {cum}}

\renewcommand{\vec}[1]{{\bf{#1}}}     

\newcommand{\vecsc}[1]{\mbox {\boldmath \scriptsize $#1$}}     
\newcommand{\itvec}[1]{\mbox {\boldmath $#1$}}
\newcommand{\itvecsc}[1]{\mbox {\boldmath $\scriptstyle #1$}}
\newcommand{\gvec}[1]{\mbox{\boldmath $#1$}}

\newcommand{\balpha}{\mbox {\boldmath $\alpha$}}
\newcommand{\bbeta}{\mbox {\boldmath $\beta$}}
\newcommand{\bgamma}{\mbox {\boldmath $\gamma$}}
\newcommand{\bdelta}{\mbox {\boldmath $\delta$}}
\newcommand{\bepsilon}{\mbox {\boldmath $\epsilon$}}
\newcommand{\bvarepsilon}{\mbox {\boldmath $\varepsilon$}}
\newcommand{\bzeta}{\mbox {\boldmath $\zeta$}}
\newcommand{\boldeta}{\mbox {\boldmath $\eta$}}
\newcommand{\btheta}{\mbox {\boldmath $\theta$}}
\newcommand{\bvartheta}{\mbox {\boldmath $\vartheta$}}
\newcommand{\biota}{\mbox {\boldmath $\iota$}}
\newcommand{\blambda}{\mbox {\boldmath $\lambda$}}
\newcommand{\bmu}{\mbox {\boldmath $\mu$}}
\newcommand{\bnu}{\mbox {\boldmath $\nu$}}
\newcommand{\bxi}{\mbox {\boldmath $\xi$}}
\newcommand{\bpi}{\mbox {\boldmath $\pi$}}
\newcommand{\bvarpi}{\mbox {\boldmath $\varpi$}}
\newcommand{\brho}{\mbox {\boldmath $\rho$}}
\newcommand{\bvarrho}{\mbox {\boldmath $\varrho$}}
\newcommand{\bsigma}{\mbox {\boldmath $\sigma$}}
\newcommand{\bvarsigma}{\mbox {\boldmath $\varsigma$}}
\newcommand{\btau}{\mbox {\boldmath $\tau$}}
\newcommand{\bupsilon}{\mbox {\boldmath $\upsilon$}}
\newcommand{\bphi}{\mbox {\boldmath $\phi$}}
\newcommand{\bvarphi}{\mbox {\boldmath $\varphi$}}
\newcommand{\bchi}{\mbox {\boldmath $\chi$}}
\newcommand{\bpsi}{\mbox {\boldmath $\psi$}}
\newcommand{\bomega}{\mbox {\boldmath $\omega$}}

\newcommand{\R}{\mathbb{R}}
\newcommand{\N}{\mathbb{N}}

\def\calA{{\mathcal A}}
\def\calB{{\mathcal B}}
\def\calC{{\mathcal C}}
\def\calD{{\mathcal D}}
\def\calE{{\mathcal E}}
\def\calF{{\mathcal F}}
\def\calG{{\mathcal G}}
\def\calH{{\mathcal H}}
\def\calI{{\mathcal I}}
\def\calJ{{\mathcal J}}
\def\calK{{\mathcal K}}
\def\calL{{\mathcal L}}
\def\calM{{\mathcal M}}
\def\calN{{\mathcal N}}
\def\calO{{\mathcal O}}
\def\calP{{\mathcal P}}
\def\calQ{{\mathcal Q}}
\def\calR{{\mathcal R}}
\def\calS{{\mathcal S}}
\def\calT{{\mathcal T}}
\def\calU{{\mathcal U}}
\def\calV{{\mathcal V}}
\def\calW{{\mathcal W}}
\def\calX{{\mathcal X}}
\def\calY{{\mathcal Y}}
\def\calZ{{\mathcal Z}}

\def\bA{\mbox {\boldmath $A$}}
\def\bB{\mbox {\boldmath $B$}}
\def\bC{\mbox {\boldmath $C$}}
\def\bD{\mbox {\boldmath $D$}}
\def\bE{\mbox {\boldmath $E$}}
\def\bF{\mbox {\boldmath $F$}}
\def\bG{\mbox {\boldmath $G$}}
\def\bH{\mbox {\boldmath $H$}}
\def\bI{\mbox {\boldmath $I$}}
\def\bJ{\mbox {\boldmath $J$}}
\def\bK{\mbox {\boldmath $K$}}
\def\bL{\mbox {\boldmath $L$}}
\def\bM{\mbox {\boldmath $M$}}
\def\bN{\mbox {\boldmath $N$}}
\def\bO{\mbox {\boldmath $O$}}
\def\bP{\mbox {\boldmath $P$}}
\def\bQ{\mbox {\boldmath $Q$}}
\def\bR{\mbox {\boldmath $R$}}
\def\bS{\mbox {\boldmath $S$}}
\def\bT{\mbox {\boldmath $T$}}
\def\bU{\mbox {\boldmath $U$}}
\def\bV{\mbox {\boldmath $V$}}
\def\bW{\mbox {\boldmath $W$}}
\def\bX{\mbox {\boldmath $X$}}
\def\bY{\mbox {\boldmath $Y$}}
\def\bZ{\mbox {\boldmath $Z$}}

\def\ba{\mbox {$\bf{a}$}}
\def\bb{\mbox {\boldmath $b$}}
\def\bc{\mbox {\boldmath $c$}}
\def\bd{\mbox {\boldmath $d$}}
\def\be{\mbox {\boldmath $e$}}
\def\bg{\mbox {\boldmath $g$}}
\def\bh{\mbox {\boldmath $h$}}
\def\bi{\mbox {\boldmath $i$}}
\def\bj{\mbox {\boldmath $j$}}
\def\bk{\mbox {\boldmath $k$}}
\def\bl{\mbox {\boldmath $l$}}
\def\bm{\mbox {\boldmath $m$}}
\def\bn{\mbox {\boldmath $n$}}
\def\bo{\mbox {\boldmath $o$}}
\def\bp{\mbox {\boldmath $p$}}
\def\bq{\mbox {\boldmath $q$}}
\def\br{\mbox {\boldmath $r$}}
\def\bs{\mbox {\boldmath $s$}}
\def\bt{\mbox {\boldmath $t$}}
\def\bu{\mbox {\boldmath $u$}}
\def\bv{\mbox {\boldmath $v$}}
\def\bw{\mbox {\boldmath $w$}}
\def\bx{\mbox {\boldmath $x$}}
\def\by{\mbox {\boldmath $y$}}
\def\bz{\mbox {\boldmath $z$}}

\newcommand{\snr}{\textup{SNR}}
\newcommand{\UE}{\mathrm{UE}}
\newcommand{\BS}{\mathrm{BS}}
\newcommand{\Passoc}{p_{_{I^{(1)}}}}
\newcommand{\Pintra}{p_{_{I^{(2)}}}}
\newcommand{\Pinter}{p_{_{I^{(3)}}}}

\newenvironment{Ex}
{\begin{adjustwidth}{0.04\linewidth}{0cm}
\begingroup\small
\vspace{-1.0em}
\raisebox{-.2em}{\rule{\linewidth}{0.3pt}}
\begin{example}
}
{
\end{example}
\vspace{-5mm}
\rule{\linewidth}{0.3pt}
\endgroup
\end{adjustwidth}}


\newcommand{\Hossein}[1]{{\textcolor{blue}{\emph{**Hossein: #1**}}}}
\newcommand{\Gabor}[1]{{\textcolor{cyan}{\emph{**Afsaneh: #1**}}}}
\newcommand{\Hadi}[1]{{\textcolor{red}{#1}}}
\newcommand{\gf}[1]{{\textcolor{cyan}{#1}}}
\newcommand{\REV}[1]{{\textcolor{blue}{#1}}}


\makeatletter
\let\save@mathaccent\mathaccent
\newcommand*\if@single[3]{%
 \setbox0\hbox{${\mathaccent"0362{#1}}^H$}%
  \setbox2\hbox{${\mathaccent"0362{\kern0pt#1}}^H$}%
  \ifdim\ht0=\ht2 #3\else #2\fi
  }
\newcommand*\rel@kern[1]{\kern#1\dimexpr\macc@kerna}
\newcommand*\widebar[1]{\@ifnextchar^{{\wide@bar{#1}{0}}}{\wide@bar{#1}{1}}}
\newcommand*\wide@bar[2]{\if@single{#1}{\wide@bar@{#1}{#2}{1}}{\wide@bar@{#1}{#2}{2}}}
\newcommand*\wide@bar@[3]{%
  \begingroup
  \def\mathaccent##1##2{%
    \let\mathaccent\save@mathaccent
    \if#32 \let\macc@nucleus\first@char \fi
    \setbox\z@\hbox{$\macc@style{\macc@nucleus}_{}$}%
    \setbox\tw@\hbox{$\macc@style{\macc@nucleus}{}_{}$}%
    \dimen@\wd\tw@
    \advance\dimen@-\wd\z@
    \divide\dimen@ 3
    \@tempdima\wd\tw@
    \advance\@tempdima-\scriptspace
    \divide\@tempdima 10
    \advance\dimen@-\@tempdima
    \ifdim\dimen@>\z@ \dimen@0pt\fi
    \rel@kern{0.6}\kern-\dimen@
    \if#31
      \overline{\rel@kern{-0.6}\kern\dimen@\macc@nucleus\rel@kern{0.4}\kern\dimen@}%
      \advance\dimen@0.4\dimexpr\macc@kerna
      \let\final@kern#2%
      \ifdim\dimen@<\z@ \let\final@kern1\fi
      \if\final@kern1 \kern-\dimen@\fi
    \else
      \overline{\rel@kern{-0.6}\kern\dimen@#1}%
    \fi
  }%
  \macc@depth\@ne
  \let\math@bgroup\@empty \let\math@egroup\macc@set@skewchar
  \mathsurround\z@ \frozen@everymath{\mathgroup\macc@group\relax}%
  \macc@set@skewchar\relax
  \let\mathaccentV\macc@nested@a
  \if#31
    \macc@nested@a\relax111{#1}%
  \else
    \def\gobble@till@marker##1\endmarker{}%
    \futurelet\first@char\gobble@till@marker#1\endmarker
    \ifcat\noexpand\first@char A\else
      \def\first@char{}%
    \fi
    \macc@nested@a\relax111{\first@char}%
  \fi
  \endgroup
}
\makeatother

\def\herm{\mathsf{H}}
\def\trans{\mathsf{T}}
\newcommand{\call}[1]{{\textsf{\small \textsc{#1}}}}
\newcommand{\callf}[1]{{\textsf{\footnotesize \textsc{#1}}}}

\def\argmax{\mathrm{arg}\max}
\def\argmin{\mathrm{arg}\min}
\renewcommand{\algorithmicrequire}{\textbf{Input:}}
\renewcommand{\algorithmicensure}{\textbf{Output:}}
\algdef{SE}[PROCEDURE]{Procedure}{EndProcedure}%
   [2]{\algorithmicprocedure\ \textproc{#1}\ifthenelse{\equal{#2}{}}{}{(#2)}}%
   {\algorithmicend\ \algorithmicprocedure}%
\algdef{SE}[FUNCTION]{Function}{EndFunction}%
   [2]{\algorithmicfunction\ \textproc{#1}\ifthenelse{\equal{#2}{}}{}{(#2)}}%
   {\algorithmicend\ \algorithmicfunction}%

\makeatletter
\newcommand\fs@betterruled{%
  \def\@fs@cfont{\bfseries}\let\@fs@capt\floatc@ruled
  \def\@fs@pre{\vspace*{5pt}\hrule height.8pt depth0pt \kern2pt}%
  \def\@fs@post{\kern2pt\hrule\relax}%
  \def\@fs@mid{\kern2pt\hrule\kern2pt}%
  \let\@fs@iftopcapt\iftrue}
\floatstyle{betterruled}
\restylefloat{algorithm}
\makeatother

\maketitle
\begin{abstract}

Federated Learning (FL) enables clients to share learning parameters instead of local data, reducing communication overhead. Traditional wireless networks face latency challenges with FL. In contrast, Cell-Free Massive MIMO (CFmMIMO) can serve multiple clients on shared resources, boosting spectral efficiency and reducing latency for large-scale FL. However, clients' communication resource limitations can hinder the completion of the FL training. To address this challenge, we propose an energy-efficient, low-latency FL framework featuring optimized uplink power allocation for seamless client-server collaboration. Our framework employs an adaptive quantization scheme, dynamically adjusting bit allocation for local gradient updates to reduce communication costs. We formulate a joint optimization problem covering FL model updates, local iterations, and power allocation, solved using sequential quadratic programming (SQP) to balance energy and latency. Additionally, clients use the AdaDelta method for local FL model updates, enhancing local model convergence compared to standard SGD, and we provide a comprehensive analysis of FL convergence with AdaDelta local updates. Numerical results show that, within the same energy and latency budgets, our power allocation scheme outperforms the Dinkelbach
and max-sum rate methods by increasing the test accuracy up
to $7$\% and $19$\%, respectively. Moreover, for the three power allocation methods, our proposed quantization scheme outperforms AQUILA and LAQ by increasing test accuracy by up to $36$\% and $35$\%, respectively. 

%


  
\end{abstract}
\section{Introduction}
Federated Learning (FL) is a machine learning framework that allows multiple clients or clients in a wireless network to collaboratively train a model without sharing their local datasets, preserving data privacy and reducing the communication overhead associated with transmitting raw data \cite{konevcny2016federated}. In each iteration, clients train local models on their private data and share model updates with a central server, which aggregates the updates and sends back a global model. This iterative process is highly dependent on efficient communication networks, as the timely and reliable transmission of high-dimensional local models is critical for successful FL. Cell-free massive multiple-input multiple-output (CFmMIMO) networks are particularly well-suited to support FL, providing a high-quality service for uplink transmission from numerous clients, even in the presence of imperfect channel knowledge \cite{cellfreebook}. These networks leverage macro-diversity and channel hardening to ensure consistent data rates and enable the simultaneous transmission of model updates from many clients.

While Federated Learning (FL) reduces the need for data sharing by focusing on the exchange of model updates, the transmission of large-scale gradient vectors remains resource-intensive~\cite{9311931}. Addressing this challenge requires efficient resource allocation strategies, including the optimization of time, frequency, space, and energy, to alleviate communication bottlenecks and ensure seamless FL training. The following subsection presents a thorough review of recent research efforts aimed at overcoming these challenges.

\subsection{Literature Review}

Recent research has focused on reducing the communication overhead in FL through adaptive quantization methods; for example, \cite{AQG} introduces an adaptive quantized gradient method that reduces communication costs without sacrificing convergence, while AdaQuantFL \cite{AdaQuantFL} employs dynamic quantization levels to improve communication efficiency and maintain low error rates during training. Other works, such as \cite{JCDO}, accelerate Federated Edge Learning (FEEL) by combining data compression with deadline-based straggler exclusion, optimizing these parameters to minimize training time. Similarly, \cite{FedDQ} proposes a descending quantization scheme to reduce the communication load, and AQUILA \cite{AQUILA} offers an adaptive quantization approach that enhances communication efficiency while ensuring model convergence. DAdaQuant \cite{DAdaQuant} also presents a dynamic quantization algorithm that adapts to varying conditions, reducing communication costs while maintaining convergence speed. Further advancements include LAQ \cite{LAQ}, which introduces a lazily aggregated quantized gradients technique, and A-LAQ \cite{ALAQ}, which refines this method with adaptive bit allocation to further reduce communication overhead.

Several studies have proposed joint optimization approaches that tackle client selection, transmit power, data rate, and processing frequency. For instance, the authors of \cite{ 9124715} propose frameworks to jointly optimize these factors to minimize latency. Adaptive quantization is also explored to reduce uplink and downlink communication overhead, as seen in \cite{FedAQ}. Additionally, the authors of \cite{wcnc} propose a joint optimization of uplink energy and latency in FL over CFmMIMO to strike a desirable balance between energy and latency. Furthermore, FL has been extensively studied for its ability to balance communication and computation efficiency. The decentralized nature of FL means that clients contribute to model training by computing local models and sharing updates with the central server while keeping their data private \cite{konevcny2016federated}. Given these benefits, researchers have investigated methods to improve FL's communication, computation, and energy efficiency \cite{9264742, 8917592, fedcau}. For example, \cite{9264742} addresses the challenge of minimizing energy consumption in FL under latency constraints. Similarly, \cite{8917592} focuses on joint power and resource allocation in ultra-reliable low-latency vehicular networks, proposing a distributed FL-based approach to estimate queue length distributions. Moreover, \cite{fedcau} proposes a framework that jointly optimizes the FL loss function and resource consumption, highlighting the importance of integrating communication protocol design with FL for resource-efficient and accurate training. 

The authors in \cite{FedCPF} propose customized local training, partial client participation, and flexible aggregation to improve communication efficiency and convergence speed in FL. Similarly,~\cite{wang2023communication} introduces adaptive sparsification, reducing communication overhead while maintaining accuracy and linear convergence Adaptive sampling is explored in~\cite{FLAS}, where dynamic filtering of training data and client selection effectively reduces communication costs and improves convergence rates. 
Meanwhile, \cite{ARFL} proposes an adaptive model update and robust aggregation strategy to optimize local training. Moreover, \cite{sun2023efficient} introduces the FedLADA optimizer, a momentum-based algorithm that corrects local offsets using global gradient estimates, addressing convergence challenges and improving training speed. Furthermore, \cite{GC2024} proposes an adaptive quantization resolution and dynamic power control to mitigate the straggler effect in FL over CFmMIMO. These innovations collectively address the key challenges of communication, efficiency, and fairness in FL. These innovations collectively address the key challenges of communication, efficiency, and fairness in FL.

While previous works have significantly improved communication efficiency in FL, further improvements are still needed. Most existing methods focus on reducing communication overhead through adaptive quantization or optimal resource allocation, such as adapting bits, power, or frequencies~\cite{SAFARI, 8849334}. However, these approaches apply uniform quantization across the entire local gradient vector, overlooking the sparsity of the elements in the gradients, which has a negligible contribution to the global model update and, thus, can be ignored. Therefore, in this paper, we propose an adaptive element-wise quantization scheme that assigns adaptive bits to each entry of the local gradient vectors based on its value in each iteration. This technique significantly reduces the number of bits required for transmission, adapting dynamically across clients and iterations. Additionally, motivated by~\cite{chor2023more}, we employ an adaptive number of local iterations per client to accelerate global FL convergence while applying the proposed quantization scheme. We then integrate the proposed approach into a CFmMIMO network, where uplink power is optimized based on adaptive bits, optimizing energy usage and mitigating the straggler effect during the FL training. The following subsection outlines the key contributions of this paper.


\subsection{Contributions}

The main contributions of this paper are as follows:
\begin{itemize}

    \item We investigate energy and latency-efficient FL in a CFmMIMO environment, incorporating an efficient uplink power allocation strategy for the clients. In our system, each client applies AdaDelta optimization algorithm~\cite{zeiler2012adadelta}, utilizing adaptive step sizes at every local iteration to update its local FL model. We also analyze the convergence of FedAvg with local AdaDelta updates, which is among the first to integrate FedAvg with AdaDelta, providing a detailed convergence analysis.
    
    \item We propose a novel quantization scheme that dynamically assigns an adaptive number of bits to each client's local FL gradients during every global iteration. Furthermore, we analyze the quantization errors introduced by this scheme when applied to the local gradient updates.

    \item An optimization problem is formulated to train FL models over CFmMIMO systems under constrained energy and latency conditions. The optimization variables include the global FL model, the number of local iterations, uplink transmission powers, and the number of global iterations.

    \item The primary optimization problem is divided into two sub-problems solved at each global iteration: one for determining the global FL model and local iterations and the other for uplink power allocation using max-min optimization via sequential quadratic programming (SQP). The remaining energy and latency budgets dictate the total number of global iterations.


    \item Numerical results on IID and non-IID data show that our quantization scheme, using an adaptive number of local iterations, achieves comparable test accuracy to the full precision FL with maximum local iterations, thereby reducing computational resources by at least $49$\%. Furthermore, 
    within equal energy and latency budgets, our power allocation scheme improves test accuracy by up to $7$\% and $19$\% over the Dinkelbach and max-sum rate methods, respectively. Additionally, across all three power allocation methods, our quantization scheme surpasses AQUILA and LAQ, enhancing test accuracy by up to $36$\% and $35$\%, respectively.

\end{itemize}

\subsection{Paper Organization and Notation}

Section~\ref{section: System Model and Problem Formulation} details the system model for EFCAQ, encompassing the FL architecture, the proposed quantization scheme, the uplink process in CFmMIMO, and the formulation of the optimization problem. Section~\ref{section: Solution Approach} introduces the methodology employed to solve the optimization problem. The convergence properties of EFCAQ are analyzed in Section~\ref{section: Convergence analysis}, while Section~\ref{section: numerical_results} provides numerical results and Section~\ref{Section: Conclusions} presents the study's key conclusions. Proofs of the theoretical results are provided in the Appendix.

\emph{Notation:} Normal font $w$, bold font lower-case $\bw$, bold-font capital letter $\bW$, and calligraphic font $\calW$ denote scalar, vector, matrix, and set, respectively. We define the index set $[N] = \{1,2,\ldots,N\}$ for any integer $N$. We denote by $\|\cdot\|$ the $l_2$-norm, by $\lceil. \rceil$ the ceiling value, by $|\calA|$ the cardinality of the set $\calA$, by $[\bw]_i$ the entry $i$ of vector $\bw$, by $[w]_{i,j}$ the entry $i,j$ of matrix $\bW$, by $\bw{\tran}$ the transpose of $\bw$, and $\mathds{1}_{x}$ is an indicator function taking $1$ if and only if $x$ is true and takes $0$ otherwise. Finally, we use $\odot$, $\circ a$, and $\oslash$ to denote the element-wise (Hadamard) product, Hadamard power by $a$, and Hadamard division, respectively. 

\begin{figure}[t]
\centering
 \includegraphics[width=\columnwidth]{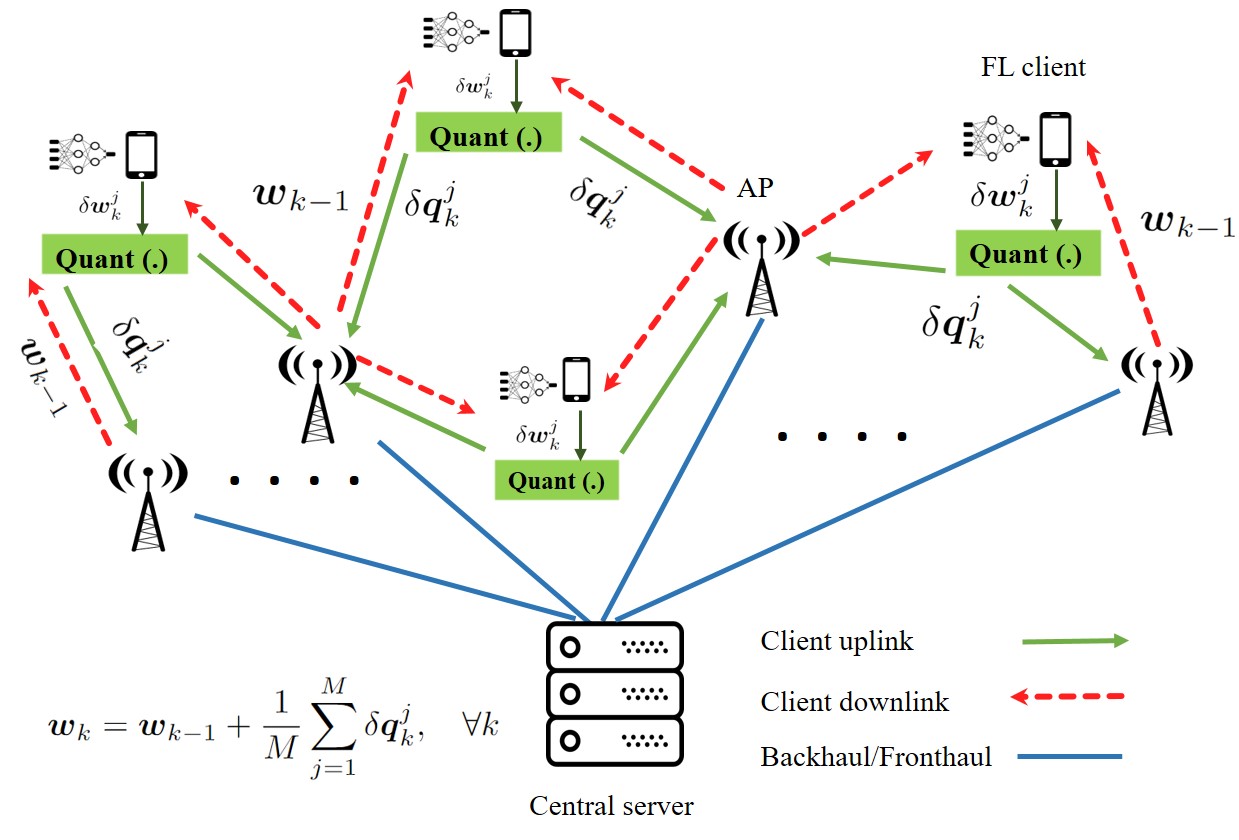}
  \caption{General architecture of FL over CFmMIMO with local model quantization.}\label{fig: CFmMIMO_FL}
\end{figure}
\section{System Model and Problem Formulation}\label{section: System Model and Problem Formulation}

In this section, we present the system model and problem formulation. First, we discuss the Federated Averaging~(FedAvg) algorithm with local AdaDelta updates. Then, we introduce our proposed quantization scheme, followed by the uplink process in CFmMIMO networks. Finally, we present the main optimization problem we aim to solve in this paper. 

\subsection{Federated Averaging with Local AdaDelta Updates
}\label{subsection: FedAvg}

We consider a CFmMIMO network with $M$ clients that cooperatively solve a distributed training problem involving a loss function $f(\bw)$. We consider $\calD$ as the dataset distributed among each client $j \in [M]$ with disjoint subset $\calD_j$ satisfying $\calD_j \cap \calD_{j'}=\emptyset$ for $j \neq j'$ and $\sum_{j =1}^{M} |\calD_j| = |\calD|$. We let the tuple $(\bx_{ij}, y_{ij})$ denote data sample $i$ of the $|\calD_j|$ samples available at client $j$.
We let $\bw \in \R^d$ denote the global model with dimension~$d$ and define $\rho_j := {|\calD_j|}/{|\calD|}$ as the fraction of data available at client $j$. Then, we formulate the following training problem 

\begin{equation}\label{eq: training}
\bw^* \in \argmin_{\bw\in\R^d} f(\bw)=\sum_{j =1}^{ M}
{\rho_j f_j(\bw)},
\end{equation}
where~$f, f_j: \R^d\rightarrow\R$, $f_j(\bw) := \sum_{i =1}^{|\calD_j|} {f(\bw; \bx_{ij}, y_{ij})}/{|\calD_j|}$. 

For a differentiable loss function~$f(\bw)$, we can solve~\eqref{eq: training} by the FedAvg algorithm~\cite{li2019convergence}. Initializing the training process with~$\bw_0$, FedAvg is a distributed learning algorithm in which the server sends $\bw_{k-1}$ to the clients at the beginning of each iteration~$k = 1,2, \ldots$. Every client $j\in[M]$ performs a number $l_k^j$ of local iterations, $i=1, \ldots,l_k^j$ 
 of stochastic gradient descent with AdaDelta updates~\cite{zeiler2012adadelta} with randomly-chosen data subset of~$\xi$, with $\left|\xi \right| \le |\calD_j|$. AdaDelta is a stochastic optimization technique that allows for a per-dimension learning rate method for SGD. It uses a running average of squared gradients, eliminating the need to manually set the learning rates. This approach is particularly beneficial for deep neural networks, as it handles varying gradient scales and prevents large learning rates, enhancing convergence even with sparse gradients~\cite{zeiler2012adadelta}. Moreover, integrating AdaDelta with FL enhances the benefits of an adaptive quantization scheme, which assigns more bits to higher-magnitude gradient elements and fewer bits to near-zero elements. AdaDelta's per-parameter adaptive learning rates naturally highlight the importance of higher-magnitude gradients by scaling updates based on each gradient's historical significance. This alignment with adaptive quantization ensures that crucial gradient information receives higher bit precision while minor updates are efficiently compressed. Such synergy is particularly valuable in FL, where communication constraints limit iterative gradient transfer between clients and the server.


 \begin{figure*}[t]
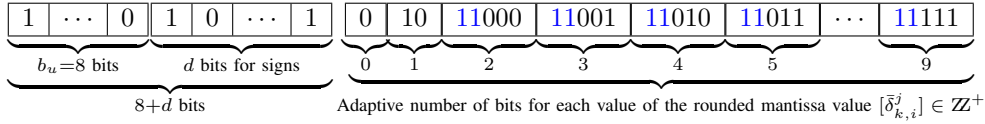
 
    \centering
    \[
     \underbrace{
    \underbrace{
    \begin{array}{|c|c|c|}
    \hline
    1 & \cdots & 0 \\ 
    \hline
    \end{array}
    }_{b_u = 8 \text{ bits}}
    \hspace{-0.3mm}
    \underbrace{
    \begin{array}{|c|c|c|c|}
    \hline
    1 & 0 & \cdots & 1  \\ 
    \hline
    \end{array}
    }_{d \text{ bits for signs}}
    }_{ 8 + d \text{ bits}}
    \underbrace{
    \hspace{-0.95mm}
    \underbrace{
    \begin{array}{|c|}
    \hline
    0  \\ 
    \hline
    \end{array}
    }_{0}
    \hspace{-0.95mm}
    \underbrace{
    \begin{array}{|c|}
    \hline
    10  \\ 
    \hline
    \end{array}
    }_{1}
    \hspace{-0.6mm}
    \underbrace{
    \begin{array}{|c|}
    \hline
     {\color{blue}11}000  \\ 
    \hline
    \end{array}
    }_{2}
    \hspace{-0.6mm}
    \underbrace{
    \begin{array}{|c|}
    \hline
     {\color{blue}11}001  \\ 
    \hline
    \end{array}
    }_{3}
    \hspace{-0.6mm}
    \underbrace{
    \begin{array}{|c|}
    \hline
     {\color{blue}11}010  \\ 
    \hline
    \end{array}
    }_{4}
    \hspace{-0.6mm}
    \underbrace{
    \begin{array}{|c|}
    \hline
     {\color{blue}11}011  \\ 
    \hline
    \end{array}
    }_{5}
    \hspace{-0.6mm}
    \begin{array}{|c|}
    \hline
    \cdots  \\ 
    \hline
    \end{array} 
    \hspace{-0.6mm}
    \underbrace{
    \begin{array}{|c|}
    \hline
    {\color{blue}11}111  \\ 
    \hline
    \end{array}
    }_{9}
    }_{\text{Adaptive number of bits for each value of the rounded mantissa value $[\Bar{\delta}_{k,i}^j] \in \Z^+$} }
    \]
    \caption{Illustration of the EMQ bit sequence assigned by each client $j \in [M]$ to $\delta\bq_k^j$ at each global iteration $k$. Colored bits indicate the prefix used for quantizing higher-magnitude rounded mantissa $[\Bar{\delta}_{k,i}^j]$.  }
    
    \label{fig: bits}
\end{figure*}



 We define~$E[\bg^2]_{l,k}^j$ as the running average at the local iteration~$l \le l_k^j$, for each client~$j$ at global iteration~$k$, and compute it as 

\begin{equation}\label{eq: Eg2}
E[\bg^2]_{l,k}^j = \rho E[\bg^2]_{l-1,k}^j + (1 - \rho) \bg_{l-1,k}^{j}\odot \bg_{l-1,k}^{j},
\end{equation}
where~$\rho~\in(0,1)$~is a decay constant 
and $\bg_{l-1,k}^{j}$ is the local gradient vector, as
\begin{equation*}
    \bg_{l-1,k}^{j}:= \frac{1}{| \xi |}\sum_{n~\in [ | \xi | ] }\nabla_{w} f(\bw_{l-1,k}^{j}; \bx_{nj}, y_{nj}).
\end{equation*}
To refine the step size for updating $\bw_{l,k}^{j}$, the AdaDelta procedure requires the element-wise square root of~$E[\bg^2]_{l,k}^j$, and it effectively becomes the RMS of the previous squared gradients up to the local iteration~$l$, as
\begin{equation}\label{eq: RMS_g}
    \textsc{RMS}[\bg]_{l,k}^j = \left( E[\bg^2]_{l,k}^j + \epsilon_a \right)^{\circ \frac{1}{2}},
\end{equation}
where $\epsilon_a > 0$ is a small constant to prevent division by zero. 
Considering $\bw_{0,k}^j = \bw_{k-1}$, we obtain
\begin{equation}\label{eq: w_lkj_update}
     \bw_{l, k}^j = \bw_{l-1, k}^j - {\alpha}~\bg_{l-1,k}^{j} \oslash { \textsc{RMS}[\bg]_{l,k}^j}  ,~l=1,\ldots,l_k^j,
\end{equation}
where~$\alpha > 0$ is the preliminary step size 
and we set~$\bw_k^j= \bw_{l, k}^j\big|_{l=l_k^j} $.

After obtaining $\bw_k^j$, each client~$j \in [M]$ computes~$\delta \bw_k^j$ defined as the difference between the current local model $\bw_k^j$ and the last global model~$\bw_{k-1}$, as
\begin{equation}\label{eq: deltaw}
    \delta \bw_k^j:= \bw_k^j - \bw_{k-1},~\hspace{2mm}\forall k, \forall j.
\end{equation}
Then, each client~$j$ quantizes each element of $\delta \bw_k^j$ according to the quantization scheme we will present in the next subsection, as
\begin{equation}\label{eq: deltaq}
    \delta \bq_k^j:= \text{Quant}(\delta \bw_k^j),~\hspace{2mm}\forall k, \forall j,
\end{equation}
and sends $\delta \bq_k^j$ to the server with~$b_k^j$ bits. Finally, considering~$\rho_j = 1/M$, the server updates the global model~$\bw_k$ as 
\begin{equation}\label{eq: EFCAQW}
    \bw_k = \bw_{k-1} +\frac{1}{M} \sum_{j=1}^{M} {\delta \bq_k^j},~\hspace{2mm} \forall k.
\end{equation}
Fig.~\ref{fig: CFmMIMO_FL} shows the general architecture of FL over 
CFmMIMO networks with the local model quantization. In the next subsection, we explain the novel quantization scheme we proposed in this paper to quantize $\delta \bw_k^j$ and send it to the server with a limited number of bits.



\subsection{Exponent-Mantissa Quantization~(EMQ) Scheme}
This subsection presents our proposed quantization scheme to compress~$\delta \bw_k^j$ and obtain~$\delta \bq_k^j$ that can be represented using a limited number of bits.

The updated gradient~$\delta \bw_k^j$ is a high-dimensional vector with many near-zero elements that contribute little to the global model but incur substantial communication overhead. To mitigate this, we propose an adaptive quantization scheme that addresses the communication bottleneck by dynamically allocating fewer bits to near-zero elements and more bits to significant ones, thereby reducing overall communication costs. Therefore, building on the inherent sparsity of local gradients~\cite{SAFARI, 8849334} and inspired by~\cite{floating-point-quant}, we propose an element-wise quantization scheme called Exponent-Mantissa Quantization (EMQ). This scheme can quantize any vector by identifying a shared exponent among all elements and unique significant digits for each element, referred to as the mantissa.\footnote{Note that in scientific or floating-point notation, the mantissa refers to the significant digits of a number, typically expressed between $1$ and $10$, while the exponent represents the order of magnitude. This definition is distinct from the logarithmic mantissa.} In the following, we describe the EMQ approach, using~$\delta \bw_k^j$ as the target vector for quantization.

 The quantization scheme expresses every element~$i$ of the vector~$\delta\bw_k^j$ as $[\delta \bw_k^j]_i:=~s_{k,i}^j\cdot\Bar{\delta}_{k,i}^j\cdot~10^{u_k^j}$, where~$s_{k,i}^j~\cdot~\Bar{\delta}_{k,i}^j$ represents the mantissa of the~$i^{\text{th}}$ element of$~\delta\bw_k^j$ with~$s_{k,i}^j~\in\{-1, +1\}$ indicating the sign of~$[\delta \bw_k^j]_i$ and~$\Bar{\delta}_{k,i}^j~\in [0,10)$. The term $u_k^j$ denotes the exponent, which satisfies~$|u_k^j|~\in~\Z^+$. To compute~$u_k^j$, we consider the infinite norm of~each vector~$\delta\bw_k^j$ as~$\|\delta \bw_k^j\|_{\infty}$ and compute the placement of the decimal point of~$\|\delta\bw_k^j\|_{\infty}$, as follows
\begin{equation}\label{eq: ukj}
   u_k^j := \floor{ \log_{10} \|\delta\bw_k^j\|_{\infty}}.
\end{equation}
 To obtain~$\delta \bq_k^j$, $[ 
\bb_k^j ]_i$ and $b_k^j$, the following procedure is performed: 
\begin{enumerate}
    \item  Each client calculates $u_k^j$ using \eqref{eq: ukj}, highlighting that $u_k^j$ denotes the exponent for the vector $\delta \bw_k^j$. Consequently, every client $j$ quantizes this value to represent the exponent for all elements within the vector, eliminating the need to consider it for each individual element. Therefore, each client allocates a maximum of $b_u = 8$ bits for $u_k^j$, similar to the single-precision floating-point format of the IEEE 754 standard for binary32.

    \item After computing~$u_k^j$, each client~$j$ assigns a single bit for~$s_{k,i}^j$, where $0$ is the negative and $1$ the positive sign. 


  \item Finally, each client $j$ computes the rounded mantissa value $[\Bar{\delta}_{k,i}^j] \in \{0, 1, \ldots, 9\}$ and encodes it using an adaptive number of bits as shown in Fig.~\ref{fig: bits}. The design principle is to use very few bit values for $0$ and $1$ (commonly appearing in gradients with near-zero elements) and $5$ bits for uncommon larger values. Note that we used a prefix-free encoding where larger values begin with $11$.  
  
  \end{enumerate}
  
  Fig.~\ref{fig: bits} illustrates the bit sequence each client uses to transmit $\delta \bq_k^j$ to the server. This sequence starts with $b_u + d = 8 + d$ bits for encoding the exponent and the sign of each mantissa. 
  The adaptive nature of EMQ, particularly the common exponent and quantization of the rounded mantissa value, differentiates the EMQ scheme from the single-precision floating-point format of the IEEE 754 standard for binary32.

We define~$\bb_k^j~\in~\R^d$ denoting the element-wise bit vector corresponding to the mantissa values of the elements of $ \delta \bq_k^j$ and the scalar~$b_k^j:= b_u + d\cdot 1 + \sum_{i=1}^d [\bb_k^j ]_i$ showing the total number of bits each client~$j$ sends to the server at every FL iteration~$k$. Thus, every client~$j$ spends~$b_k^j = b_u + d + \sum_{i=1}^d [\bb_k^j]_i \le 8 + d + 5d = 8 + 6d \approx 6d$ for transmitting the quantized version of~$\delta \bw_k^j$, defined as 
\begin{equation}
 \delta \bq_k^j = \text{Quant}(\delta \bw_k^j) = \bs_{k}^j~\odot~\Bar{\boldsymbol{\delta}}_{k}^j~\odot~\boldsymbol{1}_k^j,   
\end{equation}
where~$\boldsymbol{1}_k^j:= 10^{u_k^j}\cdot\boldsymbol{1}_d$ is a~$d$-dimensional with every element equal to~$10^{u_k^j}$, the vectorized version of rounded mantissa as $[\Bar{\boldsymbol{\delta}}_{k}^j]_i = [\Bar{\delta}_{k,i}^j]$, and the sign vector~$[\bs_{k}^j]_i:= s_{k,i}^j$ for~$i=1,\ldots,d$. Defining ${\boldsymbol \varepsilon}_k^j:=\delta \bw_k^j - \delta \bq_k^j$ as the local quantization error, the aggregated quantization error is obtained as ${{\boldsymbol \varepsilon}}_k:= \sum_{j=1}^M {\boldsymbol \varepsilon}_k^j/M$. Consequently, the following lemma shows how to evaluate the local quantization errors.

\begin{lemma}\label{lemma: local_error}
    Let~${{\boldsymbol \varepsilon}}_k^j:=\delta \bw_k^j - \delta \bq_k^j$ be the local quantization error for client~$j$ at iteration~$k$, and rewrite every element~$i$ of the vector~$\delta\bw_k^j$ as $\left[\delta \bw_k^j\right]_i:=~s_{k,i}^j\cdot\Bar{\delta}_{k,i}^j\cdot10^{u_k^j}$. Then, we obtain~${\boldsymbol~\varepsilon}_k^j := \bar{\boldsymbol{\varepsilon}}_k^j~\cdot~10^{u_k^j-1}$, where~$\|\bar{\boldsymbol{\varepsilon}}_k^j\|_{\infty} < 5$.
\end{lemma}
\begin{IEEEproof}
The proof is given in Appendix~\ref{P:lemma: local_error}.
\end{IEEEproof}
Lemma~\ref{lemma: local_error} establishes that the quantization error~$\varepsilon_k^j$ inherently depends on the value of~$u_k^j$, corresponding to the magnitude of the local FL model. This result is particularly important, as it implies that reducing the scale of the local FL model can effectively diminish or even eliminate the quantization error. Consequently, the following lemma articulates how this relationship impacts the convergence of the global FL model with the EMQ scheme.
\begin{lemma}\label{lemma: global_error}
  Let~${{\boldsymbol \varepsilon}}_k:= \sum_{j=1}^M {\boldsymbol \varepsilon}_k^j/M$ be the aggregated quantization error at iteration~$k$. Define~$\bw_k^F$ as the FedAvg global model with full precision. Then,
\begin{equation}
  \|\bw_k\|_{\infty}  \le \|\bw_k^F\|_{\infty} + 0.5\sum_{k'=1}^k 10^{u_{k'}^{\max}},
\end{equation}
where~$u_{k'}^{\max}:= \max_{j} u_{k'}^j$.
  
\end{lemma}
\begin{IEEEproof}
The proof is given in Appendix~\ref{P:lemma: global_error}
\end{IEEEproof}
Next, in the following subsection, we explain the uplink process in CFmMIMO that we utilize to facilitate the iterative process in EFCAQ.

\subsection{Uplink Process in CFmMIMO }
We consider a CFmMIMO system consisting of $A_p$ APs, each with $N$ antennas and $M$ single-antenna FL clients that will transmit their local models. We consider the standard block-fading model where each channel takes one realization per time-frequency coherence block of~$\tau_c$ channel uses and independent realizations across blocks~\cite{cellfreebook}. 
We assume independent Rayleigh fading where $g_{m,n}^{j} \sim \mathcal{N}_{\mathbb{C}}(0, \beta_{m}^{j})$ is the channel coefficient between client~$j$ and the~$n${{th}} antenna of AP~$m$, 
where~$\beta_{m}^{j}$ is the large-scale fading coefficient.
We focus on the uplink where the coherence block length $\tau_c$ is divided into $\tau_p < \tau_c$ pilot channel uses and $\tau_c-\tau_p$ data channel uses.

\subsubsection{Channel Estimation}

Each client is assigned a $\tau_p$-length pilot from a set of $\tau_p$ mutually orthogonal pilot sequences. 
The pilot of client $j$ is denoted $\sqrt{\tau_p} \boldsymbol{\varphi}_j~\in~\mathbb{C}^{\tau_p\times 1}$, where $p^u$ is the maximum transmit power, and $\|\boldsymbol{\varphi}_j\|^2 = 1$.
The clients send these sequences simultaneously and then the received vector~$\by_{m,n}^{p} \in \mathbb{C}^{\tau_p \times 1}$ at antenna~$n$ of the AP $m$ is
\begin{equation*}\label{eq: ypnl}
  \by_{m,n}^{p} = \sqrt{p_p}\sum_{j=1}^M g_{m,n}^{j} \boldsymbol{\varphi}_j + \boldsymbol{\omega}_{m,n}^{p}, 
\end{equation*}
where $p_p := \tau_p p^u$ is the pilot energy and~$\boldsymbol{\omega}_{m,n}^{p}$ is the noise with the noise power of~$\sigma^2$. By projecting~$\by_{m,n}^{p}$ onto the pilot~$\boldsymbol{\varphi}_j$, we obtain~$\hat{y}_{m,n}^{p,j} = \boldsymbol{\varphi}_j^H \by_{m,n}^{p} $. Then, the MMSE estimate of~$g_{m,n}^{j}$ is obtained as~\cite{cellfreebook}
\begin{equation*}\label{eq: ghat}
   \hat{g}_{m,n}^{j} = \frac{\mathds{E}\left\{ {\hat{y}_{m,n}^{p,j}}{g^*}_{m,n}^{j}\right\}}{\mathds{E}\left\{ \left|\hat{y}_{m,n}^{p,j}\right|^2\right\}} = c_{m}^{j} \hat{y}_{m,n}^{p,j},
\end{equation*}
where
\begin{equation*}\label{eq: cnlk}
    c_{m}^{j} := \frac{\sqrt{p_p} \beta_{m}^{j}}{p_p \sum_{j'=1}^{K} \beta_{m}^{j'} | \boldsymbol{\varphi}_{j'}^H \boldsymbol{\varphi}_j |^2 + \sigma^2 }. 
\end{equation*}
\subsubsection{Uplink Data Transmission}
We consider that all the $M$ clients simultaneously send their local models to the APs in the uplink. We define~$q_j$, $j=1,\ldots,K$, as the independent unit-power data symbol associated with the $j$th client (i.e., $\mathds{E}\left\{ |q_j|^2 \right\} = 1$). Client~$j$ sends its signal using the power~{\color{black}$p^u p_j$}, where $p_j$, $0 \le p_j \le 1$, is the power control coefficient we will optimize later. 
Then, the received signal~$ y_{m,n}^{u,j}$ at antenna~$n$ of AP~$m$ is
\begin{equation*}\label{eq: ymnk}
    y_{m,n}^{u} = \sum_{j=1}^M g_{m,n}^{j} q_j \sqrt{{\color{black}{p^u}} p_j} + \omega_{m,n}^u,
\end{equation*}
where~$\omega_{m,n}^u \sim \mathcal{N}_{\mathbb{C}}(0,{\color{black}\sigma^2})$ is the additive
noise at the $n$th antenna of AP~$m$. After the APs receive the uplink signal, they use maximum-ratio processing. Similar to~\cite[Th.~2]{7827017}, the achievable uplink data rate (in bit/s) at client $j$ is
\begin{alignat}{3}\label{eq: Rkj}
    &r_k^j = B_{\tau} \log_2\left(1+\text{SINR}_k^j\right),
    \vspace{-0.001\textheight}
\end{alignat}
where~${\text{SINR}}_k^j$ is the signal-to-interference+noise ratio of client~$j$ at iteration~$k$, $B_\tau := B(1 - { \tau_p}/{\tau_c})$ is the pre-log factor, $B$ is the bandwidth~(in Hz),~${\gamma_{m}^{j}} := \mathds{E}\left\{ |\hat{g}_{m,n}^j|^2\right\} = \sqrt{p_p} \beta_{m}^j c_{m}^j $,
\begin{alignat}{3}\label{eq: definitions}
\nonumber
\text{SINR}_k^j &:= \frac{\bar{A}_j {p}_k^j}{\bar{B}_j {p}_k^j + \sum_{j' \neq j}^M p_k^{j'} \Tilde{B}_{j}^{j'} + I_M^j },\: \\
      \nonumber
      \bar{B}_j &:= \sum_{m=1}^{A_p} N {\gamma_{m}^{j}} \beta_{m}^{j},\quad \bar{A}_j := \left( \sum_{m=1}^{A_p} N{\color{black}{\gamma_{m}^{j}}}\right)^{\! 2} , \: \\ 
      I_M^{j} &:= \sum_{m=1}^{A_p}~N{\sigma^2{\gamma_{m}^{j}}}/{{p^u}},\: \\
      \nonumber
      \Tilde{B}_{j}^{j'} &:= \sum_{m=1}^{A_p} N {\gamma_{m}^{j}} \beta_{m}^{j'} + |\boldsymbol{\varphi}_{j}^{H} \boldsymbol{\varphi}_{j'}|^2\left(\sum_{m=1}^{A_p} N {\gamma_{m}^{j}}\frac{\beta_{m}^{j'}}{\beta_{m}^{j}}\right)^2, \: \\
      \nonumber
        {\gamma_{m}^{j}}&:= \frac{p_p \left(\beta_{m}^{j}\right)^2}{p_p \sum_{j'=1}^{M} \beta_{m}^{j'} | \boldsymbol{\varphi}_{j'}^H \boldsymbol{\varphi}_j |^2 + \sigma^2}.
      \end{alignat}
Now, we provide the optimization problem formulation in the following subsection.    
\subsection{Problem Formulation}\label{subsection: Problem Formulation}
In this subsection, we consider the FedAvg algorithm with EFCAQ and propose an optimization problem to obtain the communication parameters. After computing the local model $\bw_k^j$, each client $j \in [M]$ computes $\delta \bw_k^j$ and obtains $\delta \bq_k^j$ with $b_k^j$ number of bits and sends the quantized vector $\delta \bq_k^j$ to the server. We consider $K$ as the number of global iterations, and define $\bW \in \R^{d\times K}$ as the matrix version of all the global models~$\bw \in \R^d$, $  \bP, \bL \in \R^{K \times M}$ as the matrix versions of  power coefficients~$\bP:=[p_k^j ]_{k,j}$, and number of local iterations~$\bL:=[l_k^j ]_{k,j}$, respectively. Then, we state the following optimization problem:
\begin{subequations}\label{eq: optimization}
\begin{alignat}{3}
\label{genral1}
  \underset{{{{\color{black}\bP}, \bw_K, \bL}, K}}{\mathrm{minimize}} & \quad f(\bw_K) \: \\ 
  \text{subject to} 
\label{EFCAQ w_k}
 &\quad \bw_k = \bw_{k-1} + \frac{1}{M}\sum_{j=1}^{M}\delta \bq_k^j ,\quad k\in[K]\: ,\\
\label{l_kj}
&\quad l_k^j \ge 1,\quad k\in[K], j\in [M]  \: ,\\
\label{power}
&\quad p_k^j \in [0,1] ,\quad k\in[K], j\in [M]  \: ,\\
 \label{Energy}
 &\quad {\color{black}\sum_{k=1}^{K} \sum_{j=1}^{M} E_k^j \le \mathcal{E}} \: ,\\
\label{latency}
&\quad {\color{black}\sum_{k=1}^{K}  \ell_k^{\max}\le \bar{\calL}},
\end{alignat}
\end{subequations}
where~$E_k^j$ is the uplink energy that each client $j\in [M]$ spends at each global iteration~$k$, $\mathcal{E}$ and $\bar{\calL}$ are the total energy and latency budget for uplink, $\ell_k^{\max}$ is the latency of the slowest client in uplink process at the global iteration~$k$, and $p_k^j$ is the uplink power coefficient of client~$j$ at each global iteration~$k$. We recall client uplink data rate~$r_k^j$ from~\eqref{eq: Rkj} and compute the uplink latency for each client $j\in[M]$, as~$\ell_k^j=b_{k}^j/r_k^j$~sec, and define~$\ell_k^{\max} := \max_j \ell_k^j$. Accordingly, the uplink energy $E_k^j$ of each client~$j~\in [M]$ is obtained as~$E_k^j := p_k^j p^u \ell_k^j$.





\section{Solution Approach}\label{section: Solution Approach}

The optimization problem in~\eqref{eq: optimization} is challenging to solve in advance (before training begins) because it requires information about parameters such as~$p_j^k$, $l_j^k$, $b_k^j$, and $\delta \bw_k^j$ for every client~$j~\in [M]$ and each iteration~$k$, none of which are available prior to training. To address this issue, similar to the approach in~\cite{fedcau}, we propose solving~\eqref{eq: optimization} at each iteration~$k$ without needing data from future iterations. As a result, for every iteration~$k$, the goal is to compute~$\bw_k$, $l_k^1,\ldots,l_k^M$, and $p_k^1,\ldots,p_k^M$ to minimize~$f(\bw_k)$ while considering our quantization scheme.
\begin{algorithm}[t]
\caption{Local updates with AdaDelta at each global FL iteration}
\label{alg: local adadelta}
\begin{algorithmic}[1]
\State \textbf{Inputs:} $\bw_0$, $M$, ${(\bx_{ij}, y_{ij})}_{i,j}$, $\alpha$, $\{|\calD_j|\}_{j\in[M]}$, $\rho$, $k$, $l^{\max}$, $\epsilon_a$, $\xi$, $\{l_{k-1}^j\}_{j\in[M]}$


\vspace{0.005\textheight}
\State \textbf{Initialize:} $E[\bg^2]_{0,k}^j = \boldsymbol{0}$

\vspace{0.0025\textheight}
\State Set~$\bw_{0,k}^j = \bw_{k-1}$
\vspace{0.0025\textheight}
\For{$l=1,\ldots, l^{\max}$}
\Comment{{\color{cyan} Local iterations}} 
\vspace{0.0025\textheight}
\State Compute~$\bg_{l-1,k}^{j}= \sum\limits_{n \in [ | \xi | ] }\nabla_{w} f(\bw_{l-1,k}^{j}; \bx_{nj}, y_{nj}) / | \xi |$
\State Update \Comment{{\color{cyan} Local AdaDelta}} $$E[\bg^2]_{l,k}^j = \rho E[\bg^2]_{l-1,k}^j + (1 - \rho) \bg_{l-1,k}^{j}\odot \bg_{l-1,k}^{j}$$

\State Compute $\textsc{RMS}[\bg]_{l,k}^j = \left( E[\bg^2]_{l,k}^j + \epsilon_a \right)^{\circ \frac{1}{2}} $

\State Update $\bw_{l, k}^j = \bw_{l-1, k}^j - {\alpha}~\bg_{l-1,k}^{j} \oslash{ \textsc{RMS}[\bg]_{l,k}^j}$

 \State Compute $\delta \bw_{k}^j = \bw_k^j - \bw_{k-1}$
 
\State Compute $u_{l,k}^j = \floor{ \log_{10} \|\delta\bw_{l,k}^j\|_{\infty}}$
 
\If{$l \ge l_{k-1}^j$, and $u_{l,k}^j \le u_{k-1}^j$} 
\Comment{{\color{cyan} Remark~\ref{remark: l_kj solve}}}
\vspace{0.0025\textheight}
\State Set~$l_k^j = l$,~$u_{k}^j = u_{l,k}^j$,~$\delta\bw_k^j = \delta \bw_{l,k}^j$
\State Break and go to line 16
\vspace{0.005\textheight}
\EndIf 

\EndFor
\vspace{0.005\textheight}
\State \textbf{Return}~$\delta\bw_k^j$,~$l_k^j$,~$u_k^j$ 
\end{algorithmic}
\end{algorithm}
To this aim, we decompose the optimization problem in~\eqref{eq: optimization} at each iteration~$k$ into two sub-problems. The first sub-problem focuses on determining~$\bw_k$ and $l_k^j$ for each client~$j=1,\ldots, M$, while the second sub-problem focuses on calculating~$p_k^j$ and $\ell_k^{\max}$ for every client~$j=1,\ldots, M$, minimizing both uplink energy consumption and latency. The rationale for this separation is that $p_k^j$ and $\ell_k^{\max}$ do not influence the training variables~$\bw_k$ and $l_k^j$,~$j~\in~[M]$, but rather they impact the overall number of iterations~$K$ due to the energy and latency budget. However, the quantization bits~$b_k^j$ for each client, which are computed locally after completing the local updates, are required to solve the power allocation sub-problem. Therefore, we propose solving the first sub-problem to obtain the local models~$\bw_k^j$ and consequently determine $b_k^j$ for each client~$j=1,\ldots, M$, followed by solving the second sub-problem to compute the optimal uplink power coefficients~$p_k^j$, $j~\in [M]$, and~$\ell_k^{\max}$.\footnote{Note that the number of quantization bits is not a direct optimization variable but is computed after $l_k^j$ iterations are completed. The first sub-problem obtains~$l_k^j$ for each client, after which the number of bits~$b_k^j$ is obtained.} In the following, we present the two sub-problems: $\calP 1$ and~$\calP 2$.

The first sub-problem~$\calP 1$ is as follows:
\begin{subequations}\label{eq: sub_optimization1}
\begin{alignat}{3}
\label{sub_genral1}
 \left(\calP 1\right) \quad \underset{{{ \bw_k, l_k^1,\ldots,l_k^M }}}{\mathrm{minimize}} & \quad f(\bw_k) \: \\ 
  \text{subject to} 
\label{sub_EFCAQ w_k}
 &\quad \bw_k = \bw_{k-1} + \frac{1}{M}\sum_{j=1}^{M}\delta \bq_k^j \: , \\
\label{sub_l_kj}
&\quad l_k^j \ge 1,\quad j\in [M].
\end{alignat}
\end{subequations}
To obtain~$\bw_k$ in~$\calP 1$, we perform the FL local training of~\eqref{eq: w_lkj_update} and global update of~\eqref{eq: EFCAQW} at every iteration~$k$. To obtain~$l_k^j$, $j~\in~[M]$, we analyze the behavior of the quantized local gradients obtained through the EMQ scheme, with a particular focus on the values of the local iterations~$l_k^j$, $\forall j$. The following proposition provides valuable insights into the necessary conditions for determining the optimal~$l_k^j$ in the context of the sub-problem~$\calP 1$ outlined in~\eqref{eq: sub_optimization1}.

\begin{prop}\label{prop: lkj and ukj}
    Let~$\bg_{l,k}^j$ be the gradient vector of client~$j \in [M]$ at local iteration~$l=1,\ldots,l_k^j$ and at every global iteration~$k$. We recall the AdaDelta updates according to~\eqref{eq: Eg2}-\eqref{eq: w_lkj_update}, and the definition of~$u_k^j$ from~\eqref{eq: ukj}. Let~$\alpha~\in~(0,1)$ be the primary step size. Then, for a sufficiently large~$l_{K^{\max}}$ and any~$\alpha~<1/\sqrt{10}$, there exists~$l_k^j$ such that~$l_{K^{\max}} > l_k^j \ge l_{k-1}^j \ge 2$ by which we obtain~$u_{k}^j \le u_{k-1}^j $, for all~$k~\in [K^{\max}]$, and $j~\in~[M]$.
\end{prop}
\begin{IEEEproof}\label{P:prop: lkj and ukj}
   See Appendix~\ref{P:prop: lkj and ukj}.   
\end{IEEEproof}
Proposition~\ref{prop: lkj and ukj} demonstrates that for each client~$j$, in the case of local AdaDelta updates, a non-decreasing sequence of~$l_k^j$ with respect to~$k$ leads to a corresponding non-increasing sequence of~$u_k^j$. This result is significant because the infinity norm of the quantization error~$\varepsilon_k^j$ is influenced by~$u_k^j$. As such, the quantization error in the EMQ scheme decreases or even vanishes as~$u_k^j$ decreases, particularly for large~$K$. This behavior facilitates the convergence of EFCAQ towards the standard FL method, reducing the impact of quantization as the global iterations increase. As a result, by utilizing the results of Proposition~\ref{prop: lkj and ukj}, we establish the following remark to obtain the optimal~$l_k^1,\ldots, l_k^M$ for solving the sub-problem~$\calP 1$.
\begin{remark}\label{remark: l_kj solve}
      Let us define~$\Delta \bw_{l,k}^j:= \bar{\Delta}\bw_{l,k}^j \times 10^{u_{l,k}^j} = \bw_{l,k}^j - \bw_{l-1,k}^j$, where~$u_{l,k}^j:= \floor{ \log_{10} |\Delta \bw_{l,k}^j |_{\infty}}$. Based on Proposition~\ref{prop: lkj and ukj}, which demonstrates that for each client $j \in [M]$, a non-increasing sequence of~$l_k^j$ w.r.t.~$k$ leads to a non-increasing sequence of~$u_k^j$, we obtain:
\begin{equation}\label{eq: l_kj solve}
  l_k^j:= \text{The first}~l~\Big|~l \ge l_{k-1}^j, \text{and}~ u_{l,k}^j \le u_{k-1}^j, 
\end{equation}
and obtain $u_k^j:= u_{l_k^j, k}^j$.    
\end{remark}
Remark~\ref{remark: l_kj solve} establishes an approach to obtain every~$l_k^j$, $j~\in~[M]$, solving optimization~\eqref{eq: sub_optimization1} of sub-problem~$\calP 1$. The steps of the local FL updates with AdaDelta, with a focus on Remark~\ref{remark: l_kj solve}, are summarized in Algorithm~\ref{alg: local adadelta}. 


Next, we present the second sub-problem that considers the uplink power allocation under constrained uplink energy and latency conditions:
\begin{subequations}\label{eq: sub_optimization2}
\begin{alignat}{3}
\label{sub_genral2}
 \left(\calP 2\right) \quad \underset{p_k^1,\ldots,p_k^M} {\mathrm{minimize}} & \quad  \theta_l \left( \max_{j~\in [M]} \ell_k^j \right) + \theta_E \sum_{j=1}^M E_k^j\: \\ 
  \text{subject to} 
\label{sub_p_kj}
&\quad 0 \le p_k^j \le 1,\quad j\in [M],
\end{alignat}
\end{subequations}
where~$\max_{j~\in [M]} \ell_k^j$ is the uplink latency of the slowest client, $0 \le \theta_E, \theta_l \le 1$ are scalarization weights between the energy and latency. Optimization problem~\eqref{eq: sub_optimization2} indicates the trade-off between the uplink latency\footnote{The uplink latency of the slowest client at every iteration~$k$ determines the uplink latency for that iteration. It is also referred to as the straggler effect.} and the total uplink energy at every global iteration~$k$. The trade-off is non-trivial
since the lowest energy is achieved by transmitting at very
low rates, which leads to high latency—and vice versa~\cite{wcnc}. To solve~\eqref{eq: sub_optimization2}, we consider the definition of~$\ell_k^{\max} = \max_j \ell_k^j$, resulting in~$\ell_j \le \ell_k^{\max}$, $j~\in [M]$. Then, we re-write~\eqref{eq: sub_optimization2} as
\begin{subequations}\label{eq: equiv_sub_optimization2}
\begin{alignat}{3}
\label{equiv_sub_genral2}
 \left(\calP 2\right) \quad \underset{p_k^1,\ldots,p_k^M, \ell_k^{\max}} {\mathrm{minimize}} & \quad  \theta_l \ell_k^{\max} + \theta_E \sum_{j=1}^M E_k^j\: \\ 
  \text{subject to} 
\label{equiv_sub_p_kj}
&\quad 0 \le p_k^j \le 1,\quad j\in [M], \: \\ 
\label{equiv_sub_ell_kj}
&\quad  \ell_k^j \le \ell_k^{\max},\quad j\in [M].
\end{alignat}
\end{subequations}
Optimization problem~\eqref{eq: equiv_sub_optimization2} is nonlinear because the objective function and constraints involve nonlinear relationships between the decision variables $p_k^j$ and the functions $\ell_k^j$ and $E_k^j$. Additionally, the constraint $\ell_k^j \le \ell_k^{\max}$ introduces further nonlinearity, as it involves the maximum of a set of nonlinear functions. Thus, it makes both the objective function and the feasible region nonlinear, classifying the problem as nonlinear optimization. Therefore, to solve~\eqref{eq: equiv_sub_optimization2}, we use the Sequential Quadratic Programming (SQP) approach~\cite{nocedal1999numerical}. To solve~\eqref{eq: equiv_sub_optimization2} with SQP, we first construct the Lagrangian function~$\calL$ that incorporates both the objective function and the constraints as
\begin{alignat}{3}\label{eq: Lagrangian}
& \calL(\bp_k, \ell_k^{\max}, \boldsymbol{\lambda}_k, \boldsymbol{\mu}_k, \boldsymbol{\nu}_k) :=   \theta_l \ell_k^{\max} + \theta_E \sum_{j=1}^M E_k^j +\: \\ 
 \nonumber
&  \sum_{j=1}^M \lambda_k^j (\ell_k^j - \ell_k^{\max} ) + \sum_{j=1}^M \mu_k^j(1-p_k^j)  - \sum_{j=1}^M \nu_k^j p_k^j, 
\end{alignat}
where~$\bp_k:= [p_k^1,\ldots, p_k^M]^\top$ is the vectorized power coefficients, $\boldsymbol{\lambda}_k:= [\lambda_k^1,\ldots, \lambda_k^M]^\top$, $\boldsymbol{\mu}_k:= [\mu_k^1,\ldots, \mu_k^M]^\top$ and~$\boldsymbol{\nu}_k:= [\nu_k^1,\ldots, \nu_k^M]^\top$ are the vectorized Lagrange multipliers associated with constraints of latencies~$\ell_k^j \le \ell_k^{\max}$, the power coefficients $p_k^j \le 1$ and $p_k^j \ge 0$, respectively. 

The next step is formulating the Quadratic Programming (QP) subproblem to iteratively determine the search direction~$\Delta \bx_k^i:= [\Delta p_k^{1,i}, \ldots,\Delta p_k^{M,i}, \Delta \ell_k^{\max, i}]^\top$, for $i=1,\ldots, I_{\max}$ rounds, that minimizes the quadratic model of the objective function~$F(\bx_k):= \theta_l \ell_k^{\max} + \theta_E \sum_{j=1}^M E_k^j$, $\bx_k := [p_k^{1}, \ldots,p_k^{M}, \ell_k^{\max}]^\top$, and a linear approximation of the constraints, as
\begin{subequations}\label{eq: QP}
\begin{alignat}{3}
    \underset{\Delta \bx_k^i}{\mathrm{minimize}} \quad & \frac{1}{2} {\Delta \bx_k^i}^\top \bH_k^i~\Delta \bx_k^i + \nabla_{x}~F(\bx_k^i)^\top \Delta \bx_k^i \: \\ 
     \label{QP_p0}
    \text{subject to} \quad & p_k^{j,i} + \Delta p_k^{j,i} \ge 0, \quad j~\in~[M] \: , \\ 
     \label{QP_p1}
    & p_k^{j,i} -1 + \Delta p_k^{j,i} \le 0, \quad j~\in~[M]  \: , \\ 
     \label{QP_ell}
    & \ell_k^{j,i} - \ell_k^{\max, i} + \frac{\partial \ell_k^j}{\partial p_k^j}\bigg|_{p_k^{j,i}} \Delta p_k^{j,i} \le \Delta \ell_k^{\max, i},
\end{alignat}
\end{subequations}
where~$\bH_k^i = \nabla_{xx}~\calL(\bp_k^i, \ell_k^{\max,i}, \boldsymbol{\lambda}_k, \boldsymbol{\mu}_k, \boldsymbol{\nu}_k) $ is the Hessian of the Lagrangian, $\bx_k^i := [p_k^{1,i}, \ldots,p_k^{M,i}, \ell_k^{\max, i}]^\top$, and
\begin{equation*}
  \nabla_{x}~F(\bx_k^i):= \left[\frac{\partial F}{\partial p_k^1}, \ldots, \frac{\partial F}{\partial p_k^M}, \frac{\partial F}{\partial \ell_k^{\max}} \right]^\top \bigg|_{\bx_k^i}.  
\end{equation*}

\begin{algorithm}[t]
\caption{Power allocation to jointly optimize uplink energy and latency}
\label{alg: power allocation}
\begin{algorithmic}[1]
\State \textbf{Inputs:} $M$, $k$, $\left\{ \bar{A_j}, \bar{B}_j, \Tilde{B}_j^{j'}, I_M^j, b_k^j \right\}_{j,j'} $, $\epsilon_{x}$, $\theta_E$, $\theta_l$

\vspace{0.0025\textheight}
\State \textbf{Initialize:} $\bx_k^{1}$, $\bar{\bH}_k^{1} $ 

\For{$i = 1,\ldots, I_{\max}$,}
\Comment{{\color{cyan} Iterative rounds of QP~\eqref{eq: QP_solve}}}

\State Obtain $\Delta \bx_k^{i*}$ from~\eqref{eq: Delta x*} with~$\bH_k^i = \bar{\bH}_k^i$

\State Update~$\bar{\alpha}_k^i$ using the Armijo rule~\cite{nocedal1999numerical}

\vspace{0.0025\textheight}
\State Update $\bx_k^{i+1} = \bx_k^i + \bar{\alpha}_k^i \Delta \bx_k^{i*}$

\State Calculate $\bar{\bs}_k^i$ and $ \bz_k^i$ from~\eqref{eq: s and z}

\vspace{0.0025\textheight}
\State  Update~$\bar{\bH}_k^{i+1}$ from~\eqref{eq: barH}
\Comment{{\color{cyan} Hessian approximation}}

\If{ $\| \bx_k^{i+1} - \bx_k^i\|_{\infty}  < \epsilon_x$,}
\Comment{{\color{cyan} Convergence check}}

\State Break and go to line~{\color{black} 13 }%
\EndIf

\EndFor 

\State Set $\bp_k = \left[ [\bx_k^{i+1}]_1,\ldots, [\bx_k^{i+1}]_M  \right]^\top$, $\ell_k^{\max} = \left[\bx_k^{i+1} \right]_{M+1}$

\vspace{0.0025\textheight}
\State \textbf{Return} $\bp_k$, $\ell_k^{\max}$
\end{algorithmic}
\end{algorithm}
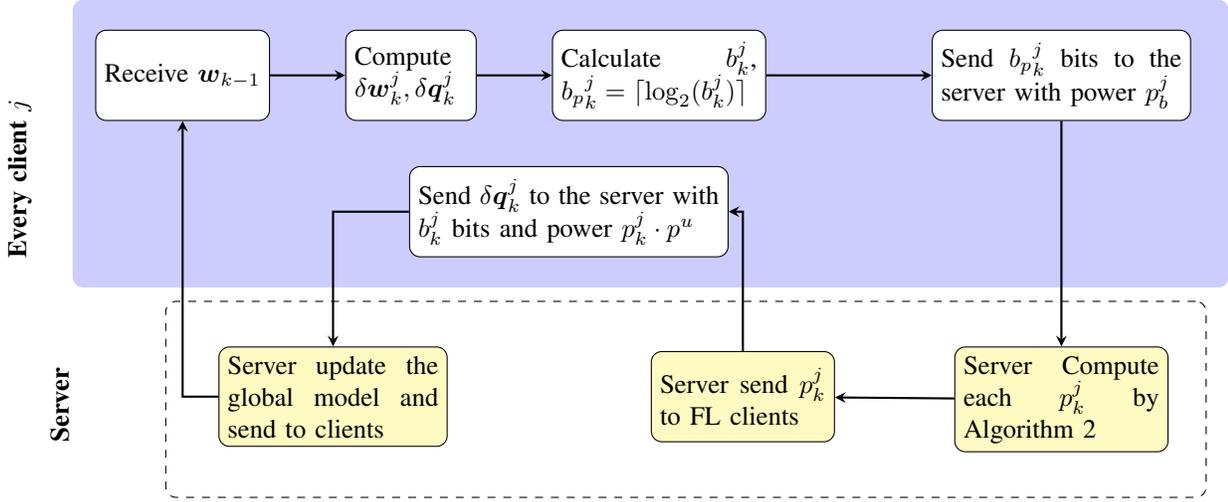
\begin{figure*}
\centering
\begin{minipage}{1.5\columnwidth}
 \centering

  \begin{center}
\begin{tikzpicture}[node distance= 1cm and 0.5cm]
\hspace{-2.1cm}
\node (receive) [process] {Receive $\bw_{k-1}$};

\node (compute) [process, right=of receive, xshift=0.5cm] {\parbox{1.5cm}{Compute $\delta \bw_k^j, \delta \bq_k^j$}};

\node (calculate) [process, right=of compute, xshift=0.5cm] {\parbox{2.6cm}{Calculate~$b_k^j$, ${b_p}_k^j = \lceil \log_2(b_k^j) \rceil$}};

\node (sendb) [process, right=of calculate, xshift=1.7cm] {\parbox{3.2cm}{Send ${b_p}_k^j$ bits to the server with power $p_b^j$}};

\node (sendr) [process, below=0.6cm of calculate, xshift=-1.2cm] {\parbox{4cm}{Send $\delta \bq_k^j$ to the server with $b_k^j$ bits and power $p_k^j\cdot p^u$}};

\node (servercompute) [server, below=3cm of receive, xshift=2cm] {\parbox{2.8cm}{Server update the global model and send to clients}};

\node (serversend) [server, right=of servercompute, xshift=2.2cm] {\parbox{2.2cm}{Server send $p_k^j$ to FL clients}};

\node (servercompute2) [server, right=of serversend, below=3cm of sendb] {\parbox{2.6cm}{Server Compute each $p_k^j$ by Algorithm~\ref{alg: power allocation}}};

\node (clientlabel) [text width=2.5cm, align=center, left=1cm of receive, rotate=90] {\textbf{Every client $j$}};

\node (serverlabel) [text width=2.5cm, align=center,  below =3cm of clientlabel, rotate=90] {\textbf{Server}};

\draw [arrow] (receive) -- (compute);
\draw [arrow] (compute) -- (calculate);
\draw [arrow] (calculate) -- (sendb);

\draw [arrow] (sendb) -- (servercompute2);
\draw [arrow] (sendr) -| (servercompute);
\draw [arrow] (servercompute2) -- (serversend);
\draw [arrow] (serversend) |- (sendr);
\draw [arrow] (servercompute) -| (receive);

\begin{scope}[on background layer]
\hspace{-2cm}
    \node [rectangle, fill=blue!20, rounded corners, fit=(receive) (sendb) (sendr), inner sep=0.4cm] {};
\end{scope}

\begin{scope}[on background layer]
\hspace{-2.2cm}
    \node [rectangle, draw=black, dashed, rounded corners, fit=(servercompute) (serversend) (servercompute2), inner sep=0.6cm] {};
\end{scope}

\end{tikzpicture}
\end{center}
\end{minipage}
   \caption{Overview of EFCAQ with power allocation for each client~$j$ at every iteration~$k$.}
   \label{powers}     
\end{figure*}
To solve~\eqref{eq: QP} and obtain~$\Delta \bx_k^{i*}$, we rewrite it in a general form, as
\begin{subequations}\label{eq: QP_solve}
\begin{alignat}{3}
    \underset{\Delta \bx_k^i}{\mathrm{minimize}} \quad & \frac{1}{2} {\Delta \bx_k^i}^\top \bH_k^i~\Delta \bx_k^i + \nabla_{x}~F(\bx_k^i)^\top \Delta \bx_k^i \: \\ 
     \label{QP_solve}
    \text{subject to} \quad & \bG_k^i \Delta \bx_k^i \le \Tilde{\bp}_k^i,
\end{alignat}
\end{subequations}
where~$\bG_k^i~\in \R^{3M \times (M+1) }$ is defined as
\begin{equation}
\bG_k^i =
\begin{bmatrix}
-\bI_M & \mathbf{0} \\
\bI_M & \mathbf{0} \\
{\boldsymbol{\partial} \ell_k^{j,i}} & -\mathbf{1}
\end{bmatrix}, 
\end{equation}
where
\begin{itemize}
    \item \( \bI_M \) is the \( M \times M \) identity matrix, corresponding to the constraint~\eqref{QP_p0},
    \item \(-bI_M \) is the negative \( M \times M \) identity matrix, corresponding to the constraint~\eqref{QP_p1},
    \item \( {\boldsymbol{\partial} \ell_k^j } \) is the \( M \times M \) diagonal matrix where each element on the diagonal is \( {\partial \ell_k^{j,i}}/{\partial p_k^j} \) at ${p_k^{j,i}} $, which corresponds to the linearized constraint~\eqref{QP_ell},
    \item \( \mathbf{0} \) is the \( M \times 1 \) column vector of zeros,
    \item \( \mathbf{1} \) is the \( M \times 1 \) column vector of ones, corresponding to \( \Delta \ell_k^{\max, i} \) in the linearized nonlinear constraint~\eqref{QP_ell}.
\end{itemize}
Accordingly, $\Tilde{\bp}_k^i \in \R^{3M}$ is obtained as
\begin{alignat}{3}
    \Tilde{\bp}_k^i&:= [\bp_k^i, \mathbf{1} - \bp_k^i, \boldsymbol{\delta}\ell_k^i ]^\top, \: \\ 
     \nonumber
     \bp_k^i&:= [ p_k^{1,i}, \ldots, p_k^{M,i}], \: \\ 
     \nonumber
    \boldsymbol{\delta}\ell_k^i &:= [\ell_k^{1,i}- \ell_k^{\max,i}, \ldots, \ell_k^{M,i}- \ell_k^{\max,i} ].
\end{alignat}
Afterward, we form the Lagrangian of~\eqref{eq: QP_solve} as
\begin{alignat}{3}
    \calL(\Delta \bx_i^k, \boldsymbol{\lambda}_Q ) & = \frac{1}{2} {\Delta \bx_k^i}^\top \bH_k^i~\Delta \bx_k^i + \: \\ 
     \nonumber
     & \nabla_{x}~F(\bx_k^i)^\top \Delta \bx_k^i + \boldsymbol{\lambda}_Q^\top \left(\bG_k^i \Delta \bx_k^i - \Tilde{\bp}_k^i \right),
\end{alignat}
where~$\boldsymbol{\lambda}_Q \ge 0$ is the vectorized Lagrangian multipliers. Next, we apply the first-order optimality conditions (KKT conditions) by setting~$\nabla_{\Delta x} \calL(\Delta \bx_i^k, \boldsymbol{\lambda}_Q ) = 0$, as
\begin{alignat}{3}
    \nabla_{\Delta x} \calL(\Delta \bx_i^k, \boldsymbol{\lambda}_Q ) = \bH_k^i~\Delta \bx_k^i + \nabla_{x}~F(\bx_k^i) + \boldsymbol{\lambda}_Q^\top \bG_k^i=0,
\end{alignat}
which gives us
\begin{alignat}{3}\label{eq: Delta x*}
    \Delta \bx_k^{i*} = -(\bH_k^i)^{-1} \left(\nabla_{x}~F(\bx_k^i) + \boldsymbol{\lambda}_Q^\top \bG_k^i \right). 
\end{alignat}
Then, for the complementary slackness and the dual feasibility of KKT conditions, we substitute $\Delta \bx_k^{i*}$  into the linear inequality constraints~$\bG_k^i \Delta \bx_k^{i*} \le \Tilde{\bp}_k^i$ and solve for the Lagrange multipliers~$\boldsymbol{\lambda}_Q$ that satisfy the complementary slackness (i.e., $\boldsymbol{\lambda}_Q^\top( \bG_k^i \Delta \bx_k^{i*} -\Tilde{\bp}_k^i) = 0$) and dual feasibility  (i.e., $\boldsymbol{\lambda}_Q \ge 0$) conditions. 

After obtaining~$\Delta \bx_k^{i*}$, we update~$\bx_k^{i+1}$ as
\begin{equation}
    \bx_k^{i+1} = \bx_k^i + \bar{\alpha}_k^i \Delta \bx_k^{i*},
\end{equation}
where $\bar{\alpha}_k^i > 0$ is the step length determined by a line search, ensuring a sufficient decrease in~$F$. To determine an appropriate step length $\bar{\alpha}_k^i$, we employ the Armijo rule, also known as backtracking line search \cite[Section 3.1]{nocedal1999numerical}. The steps for the Armijo line search are as follows. First, it computes the direction $\Delta \bx_k^i$ and initializes the step length $\bar{\alpha}_k^i = 1$. Then, it evaluates the values of $F(\bx_k^i + \bar{\alpha}_k^i \Delta \bx_k^{i*})$ and the gradient $\nabla F(\bx_k^i)$. Within a loop, it checks the condition of $F(\bx_k^i + \bar{\alpha}_k^i \Delta \bx_k^{i*}) \le F(\bx_k^i) + 0.1\bar{\alpha}_k^i  \nabla F(\bx_k^i)^{\top} \Delta \bx_k^{i*} $ to accept $\bar{\alpha}_k^i$; otherwise, it updates $\bar{\alpha}_k^i \leftarrow 0.5 \bar{\alpha}_k^i$ and repeats the loop until the condition is met. This process ensures that the step length $\bar{\alpha}_k^i$ leads to a sufficient decrease in $F$ while maintaining stability in the optimization process.

To solve~\eqref{eq: QP}, the Hessian of the Lagrangian, $\bH_k^i$, should be positive definite, which might not be fulfilled. Thus, to ensure a positive definite Hessian~$\bH_k^i$, we use the BFGS (Broyden-Fletcher-Goldfarb-Shanno)~\cite{nocedal1999numerical} update to approximate~$\bH_k^i$ with~$\bar{\bH}_k^i$. The BFGS update is a quasi-Newton method that updates the approximation of the Hessian matrix iteratively, ensuring it remains positive definite. The BFGS update for the Hessian approximation~$\bar{\bH}_k^i$ is
\begin{equation}\label{eq: barH}
     \bar{\bH}_k^{i+1} = \bar{\bH}_k^i - \frac{\bar{\bH}_k^i \bar{\bs}_k^i (\bar{\bs}_k^i)^{\top} \bar{\bH}_k^i}{ (\bar{\bs}_k^i)^{\top}\bar{\bH}_k^i \bar{\bs}_k^i} + \frac{\bz_k^i (\bz_k^i)^\top}{(\bz_k^i)^\top \bar{\bs}_k^i},
\end{equation}
where
\begin{alignat}{3}
\label{eq: s and z}
\bar{\bs}_k^i & = \bx_k^{i+1} - \bx_k^i,  \: \\ 
\nonumber
  \bz_k^i & = \nabla_{x}~\calL(\bp_k^{i+1}, \ell_k^{\max,i+1}, \boldsymbol{\lambda}_k, \boldsymbol{\mu}_k, \boldsymbol{\nu}_k)  \: \\ 
\nonumber
& - \nabla_{x}~\calL(\bp_k^i, \ell_k^{\max,i}, \boldsymbol{\lambda}_k, \boldsymbol{\mu}_k, \boldsymbol{\nu}_k).
\end{alignat}
Algorithm~\ref{alg: power allocation} summarizes the solution steps to obtain~$\bp_k$ and~$\ell_k^{\max}$ at every FL iteration~$k$. After initializing~$\bx_k^1$ and~$\bar{\bH}_k^1$, Algorithm~\ref{alg: power allocation} performs at most $I_{\max}$,~$i =1, \ldots, I_{\max}$, rounds of QP to iteratively update~$\Delta \bx_k^{i*}$, $\bar{\alpha}_k^i$ and~$\bx_k^{i+1}$ (lines~3-6). Then, it calculates $\bar{\bs}_k^i$ and $ \bz_k^i$ (line~7) to update the Hessian approximation~$\bar{\bH}_k^{i+1}$ (line~8), and then checks for the convergence (lines~9-11).

Finally, we calculate the total number of FL iterations $K$. After obtaining~$p_k^j$, $j~\in~[M]$, we need to evaluate if the energy budget~$\mathcal{E}$ and latency budget~$\bar{\calL}$ allow the training to continue. Therefore, upon obtaining $p_k^j$ from Algorithm~\ref{alg: power allocation}, the local uplink energy~$E_k^j$ and latency~$\ell_k^j$ that every client~$j~\in~[M]$ will spend on sending the quantized vectors~$\delta \bq_k^j$ can be calculated for every client $j~\in~[M]$. Thus, considering the limited budgets~$\bar{\calL}$ and $\mathcal{E}$, the server computes~$K$ as
\begin{equation}\label{eq: K}
    K = \min~\{ k_{\ell}^{\max}, k_{E}^{\max}  \} - 1,
\end{equation}
where
\begin{equation}\label{eq: k_l}
    k_{\ell}^{\max} := \text{The first global iteration}~{k'}~|~ \sum_{k=1}^{k'} \ell_{k}^{\max} > \bar{\calL},
\end{equation}
\begin{equation}\label{eq: k_e}
    k_{E}^{\max} := \text{The first global iteration}~{k'}~|~ \sum_{k=1}^{k'} \sum_{j=1}^M E_{k}^j > \mathcal{E}.
\end{equation}
Thus, $k_{\ell}^{\max}$ in~\eqref{eq: k_l} and $k_{E}^{\max}$ in~\eqref{eq: k_e} determine the number of FL iterations~$K$ in the considered resource-constrained setup. We desire to have a higher number of global iterations\footnote{This holds regardless of the computation latency and energy.} to achieve a better test accuracy from training. Therefore, spending lower energy and latency at each global iteration results in higher  $k_{\ell}^{\max}$ and $k_{E}^{\max}$, and consequently a higher~$K$, which we can achieve by Algorithm~\ref{alg: power allocation}.

Fig.~\ref{powers} outlines the steps executed by each client~$j \in [M]$ and the server during every global FL iteration. Upon receiving the global model~$\bw_k$ from the server, each client~$j$ performs~$l_k^j$ local updates to compute~$\delta\bw_k^j$. Subsequently, using the EMQ scheme, each client quantizes its local model update, resulting in~$\delta \bq_k^j$ and corresponding bit allocation~$b_k^j$. The server is then required to allocate uplink power coefficients~$p_k^j$ for each client~$j \in [M]$ as specified in Algorithm~\ref{alg: power allocation}, for which the values of~$b_k^j$ must be known. Consequently, each client~$j$ transmits the value of~$b_k^j$ to the server using ${b_p}_k^j:= \lceil \log_2(b_k^j) \rceil$ bits and sets the uplink power coefficient~$p_b^j = 1$, indicating full transmission power.\footnote{Note that ${b_p}k^j = \log_2(b_k^j) \le \log_2(8+ 6d) \ll b_k^j$, leading to negligible energy and latency overheads compared to $E_k^j$ and~$\ell_k^j$.} Once the server receives all~$b_k^j$ values from the clients, it executes Algorithm~\ref{alg: power allocation} to compute the uplink power allocation vector~$\bp_k$ and transmits it to the clients. Finally, each client~$j \in [M]$ sends its quantized update~$\delta \bq_k^j$ to the server using~$b_k^j$ bits, applying the allocated uplink power~$p_k^j \cdot p^u$ for updating the global FL model.

Algorithm~\ref{alg: FL+EMQ+power} summarizes the full process of FL training over a CFmMIMO system, incorporating the EMQ quantization scheme and power allocation within the EFCAQ framework. The algorithm includes global iterations (line~2), followed by local iterations and the EMQ scheme (lines~4-9), during which Algorithm~\ref{alg: local adadelta} is also employed. The server then executes power allocation using Algorithm~\ref{alg: power allocation} (lines~11-14) and sends back the powers to the clients. Afterward, clients transmit their quantized models to the server for global model updates (line~19). Finally, the server computes~$k_{\ell}^{\max}, k_E^{\max}$ and determines the corresponding~$K$ (lines~20-24).

\begin{algorithm}[t]
\caption{FL over CFmMIMO with EMQ scheme and power allocation}
\label{alg: FL+EMQ+power}
\begin{algorithmic}[1]
\State \textbf{Inputs:} $M$, $A_p$, $N$, $d$, ${(\bx, y)}$, $\left\{ \bar{A_j}, \bar{B}_j, \Tilde{B}_j^{j'}, I_M^j\right\}_{j,j'} $, $K^{\max}$,~$\bw_0$, $p^u$, $\bar{\calL}$, $\mathcal{E}$, Algorithms~\ref{alg: local adadelta},~\ref{alg: power allocation}

 \For{$k = 1, \ldots, K^{\max}$,}
 {\Comment{\color{cyan}FL global iterations}}

\State The APs receive~$\bw_{k-1}$ from the central server and send it to all clients via downlink transmission

\For{$j = 1, \ldots, M$,}
\Comment{{\color{cyan}Local iterations}}
    \State Set $\bw_{0,k}^j=\bw_{k-1} $ 
    \vspace{0.005\textheight}

    \State Obtain $l_k^j, u_k^j, \delta \bw_{k}^j$ from Algorithm~\ref{alg: local adadelta}
    \vspace{0.005\textheight} 
            
        \State Obtain~$\delta \bq_k^j$ by EMQ
        \Comment{{\color{cyan}Quantization}}
    

        \State Calculate~$b_k^j$ and ${b_p}_k^j= \left\lceil \log_2(b_k^j) \right\rceil$
        
        \vspace{0.005\textheight}
        \State Send~${b_p}_k^j$ bits to the APs via uplink transmission 
        \vspace{0.005\textheight}

\EndFor 

\vspace{0.005\textheight}
 \State {Central server do:}
 {\Comment{\color{cyan}Power allocation}}
  \vspace{0.005\textheight}
  \State Receive $b_k^j$ from all clients

  \vspace{0.005\textheight}
  \State Obtain~$\bp_k, \ell_k^{\max}$ from Algorithm~\ref{alg: power allocation}

\vspace{0.005\textheight}
\State Send~$\bp_k$ to all clients via downlink transmission
\For {$j=1,\ldots,M$,} in parallel:
{\Comment{\color{cyan}Uplink process}}
\vspace{0.005\textheight}
\State  Send $\delta \bq_k^j$ with~$b_k^j$ bits to the APs via uplink transmission and uplink power~$p_k^j \cdot p^u$
\vspace{0.005\textheight}
\EndFor

\vspace{0.005\textheight}
\State {Central server do:}
 {\Comment{\color{cyan}Global updates}}
 
\vspace{0.005\textheight}
\State Wait until receiving all~$\delta \bq_k^j$ and update the global model as $\bw_{k} \leftarrow \bw_{k-1} + \sum_{j =1}^{M} \delta \bq_k^j/M  $ 
\vspace{0.01\textheight}

\State Calculate~$k_{\ell}^{\max}$, $k_E^{\max}$ and $K$ according to~\eqref{eq: K}-\eqref{eq: k_e}


\If{$ k \ge K $},
\vspace{0.005\textheight}
\State Break and go to line~{\color{black} 25} 
\vspace{0.005\textheight}
\EndIf

\vspace{0.005\textheight}
\EndFor 
\vspace{0.005\textheight}
\State Set~$K = \min \left\{ k, K^{\max} \right\}$
\vspace{0.005\textheight}
\State \textbf{Output:} $\bw_{K}$, $K$

\end{algorithmic}
\end{algorithm}
\section{Convergence analysis}\label{section: Convergence analysis}
In this section, we analyze the convergence of FedAvg with local AdaDelta updates. To enable the analysis, we make the following assumptions, which are similar to the ones in~\cite{reddi2020adaptive}:
\newline
{\textbf{Assumption~1:}} (Unbiased Gradient). $\nabla_{w} f_j(\bw; \bx_{nj}, y_{nj})$ is an unbiased estimation of each client's true gradient~$\nabla_{w}f_j(\bw)$, for any~$j~\in[M], n~\in~[ |\xi| ]$. 
\newline
{\textbf{Assumption~2:}} (Lipschitz Gradient). The function~$f_j$ is~$\bar{L}$-smooth for all $j \in [M]$, $€0 < \bar{L} < \infty$, i.e.,~$\nrm{\nabla f_j(\bw_1)-\nabla f_j(\bw_2)} \le \bar{L} \left\| \bw_1 - \bw_2 \right\|$, for all $\bw_1,~\bw_2~\in~\R^d$.
\newline
{\textbf{Assumption~3:}} (Bounded Local Variance). The function~$f_j$ has~$\sigma_l$-bounded variance as $\E_{\xi}~{\nrm{ [\n f_j(\bw; \bx_{nj}, y_{nj})]_i - [\n f_j(\bw)]_i}}^2 = \sigma_{l_i}^2$, for all $j \in [M]$~$\bw~\in~\R^d$, $i~\in~[d]$, $n~\in~[ |\xi| ]$ . 
\newline
{\textbf{Assumption~4:}} (Bounded Global Variance). The gradient of the global loss function~$\n_{w} f(\bw)$ has a global variance bounded as $(1/M)\sum_{j=1}^M \nrm{[\n f_j(\bw)]_i-[\n f(\bw)]_i}^2 \le \sigma_{g_i}^2$, for all $j \in [M]$,~$\bw~\in~\R^d$, $i~\in~[d]$. 
\newline
{\textbf{Assumption~5:}} (Bounded Gradients). The function~$f_j$ has~$G$-bounded gradients as~$\left |[\n f_j(\bw; \bx_{nj}, y_{nj})]_i \right | \le G$ for any~$j~\in[M], n~\in~[\xi]$,~$\bw~\in~\R^d$ and~$i~\in~[d]$.

The bounded gradient assumption is widely used for
non-convex decentralized gradient optimization. Moreover, it holds for many neural network training tasks
due to the cross-entropy loss function~\cite{sun2022decentralized}.

Considering Assumptions~1-5, we provide a proposition establishing the convergence of FedAvg, where the clients employ AdaDelta for their local updates. 
\begin{prop}\label{prop: Fed_AdaDelta}
  Let $f$ be $\bar{L}$-smooth, $\alpha$ be the preliminary step size of FedAvg training, and the \textbf{Assumptions~1-5} hold. Assume that~$L := \max_j\{l_k^j\}$, $\rho_j = 1/M$, $\rho \neq 1$, and consider that the clients employ AdaDelta updates of~\eqref{eq: Eg2}-\eqref{eq: w_lkj_update} to obtain their local FL model. Then, for $\alpha \le\min(\alpha_0, \alpha_1, \alpha_2)$, where 
\begin{alignat}{3}\label{eq: alpha0}
\nonumber
& \alpha_0 := \left(6 L \bar{L}^2 \right)^{-0.5}, \: \\
\nonumber
    & \alpha \le \frac{\sqrt{M}}{ \sqrt{K \left( L^2 \bar{L}^2 d M + 6 L^3 \bar{L}^2 d \exp(6) \right) }} := \alpha_1, \quad &&\textbf{(I)} 
    \: \\
 & \alpha \le \sqrt{\frac{2(1-\rho)}{K L^2 \bar{L} d}}:= \alpha_2, \quad &&\textbf{(II)} 
\end{alignat}
and~$\sigma^2_k := \max_{i \in \{d\}} {\sigma_{l_i}^2}/{G^2(1-\rho)} +  \sigma_{g_i}^2 $, we obtain
\begin{equation}\label{eq: convergence_main}
  \E_k\left\{ f(\bw_{k+1}) \right\} \le   f(\bw_k) + \frac{1 + G^2 + \sigma_k^2 + \left(1-\rho^{L-1}\right)^{-1} }{ k+1 }. 
\end{equation}
\end{prop}
\begin{IEEEproof}
    See Appendix~\ref{P: prop: Fed_AdaDelta}.
\end{IEEEproof}
Proposition~\ref{prop: Fed_AdaDelta} demonstrates that when $\alpha \le \min (\alpha_0, \alpha_1, \alpha_2, 1)$, FedAvg with AdaDelta achieves a convergence rate of~\eqref{eq: convergence_main}, representing that for $k \gg 1$, $\E\left\{ f(\bw_{k+1}) \right\} - f(\bw_k) \rightarrow 0$. 

\begin{remark}\label{remark: overall convergence}
By solving the sub-problems \eqref{eq: sub_optimization1} and \eqref{eq: sub_optimization2} at each iteration \(k=1, \ldots, K\), and accounting for the bounded error of the proposed quantization scheme, EMQ, a sub-optimal solution to the main optimization problem \eqref{eq: optimization} is achieved.
\vspace{-0.008\textheight}
\end{remark}
\vspace{-0.008\textheight}
Remark~\ref{remark: overall convergence} highlights the main optimization problem \eqref{eq: optimization} achieves a sub-optimal solution. In the next section, we validate this through numerical results, showcasing the efficiency and superior training convergence of our proposed approach in this paper compared to benchmarks.
\begin{table}[t]
    \centering
    \caption{Simulation parameters.}
    \label{tab: parameters}
    \renewcommand{\arraystretch}{1.25} 
    \setlength{\extrarowheight}{1.25pt} 
    
    \resizebox{\columnwidth}{!}{%
    \selectfont
    \begin{tabular}{l l||l l}
        \hline
        \textbf{Parameter} &  \textbf{Value} & \textbf{Parameter} &  \textbf{Value}\\
        \hline
        Bandwidth & $B = 20$\,MHz & Number of clients & {$M \in \{20, 40\}$}\\
        \hline
        Area of interest (wrap around) & $1000 \times 1000$m & Pathloss exponent & {$\alpha_p= 3.67$}\\
        \hline
        Number of APs & $A_p = 16$ & Coherence block length & {$\tau_c = 200$}\\
        \hline
        Number of per-AP antennas  & $N = 4$ & Pilot length & {$\tau_p = 10$}\\
        \hline
        Uplink transmit power & {$p^u = 100$\,mW} & Uplink noise power & {$\sigma^2 = -94$\,dBm}\\
        \hline
        Noise figure & {$7$\,dB} & Size of FL model & { \color{black}{$d = 307, 498$  }}\\
        \hline
    \end{tabular}%
    }
\end{table}
\begin{figure}[t]
\centering
\begin{minipage}{0.49\columnwidth}
{\scriptsize\definecolor{amber}{rgb}{1.0, 0.49, 0.0}
\definecolor{taupe}{rgb}{0.28, 0.24, 0.2}
\definecolor{tealgreen}{rgb}{0.0, 0.51, 0.5}
\definecolor{britishracinggreen}{rgb}{0.0, 0.26, 0.15}
\definecolor{cerise}{rgb}{0.87, 0.19, 0.39}
\definecolor{mycolor1}{rgb}{0.00000,0.44700,0.74100}%
\definecolor{mycolor2}{rgb}{0.85000,0.32500,0.09800}%
\definecolor{mycolor3}{rgb}{0.92900,0.69400,0.12500}%
\definecolor{mycolor4}{rgb}{0.49400,0.18400,0.55600}%
\definecolor{bostonuniversityred}{rgb}{0.8, 0.0, 0.0}
\begin{tikzpicture}

\begin{axis}[%
width=0.8\columnwidth,
height=0.4\columnwidth,
at={(0,0)},
scale only axis,
xmin=0,
xmax=100,
ymin=9,
ymax=85,
grid style={dashed},
ymajorgrids,
grid=both,
ylabel near ticks,
ylabel={Test accuracy (\%) },
xlabel={Iteration~$k$},
axis background style={fill=white},
legend style={at={(0.98, 0.02)}, anchor=south east, legend cell align=left, inner sep=0.01pt, font = \scriptsize, text depth=0.25ex, legend image post style={scale=0.75}}, 
legend columns = 1,
every axis plot/.append style={line width=1.0pt}
]
\addplot [color=britishracinggreen, densely dashdotted, line width = 1.6]
  table[row sep=crcr]{%
0   10.08 \\
1   11.17 \\
2   27.52 \\
3   36.03 \\
4   44.67 \\
5   49.87 \\
6   55.72 \\
7   59.72 \\
8   62.22 \\
9   64.01 \\
10  66.17 \\
11  67.12 \\
12  68.15 \\
13  69.46 \\
14  70.15 \\
15  71.33 \\
16  72.68 \\
17  73.30 \\
18  73.17 \\
19  73.79 \\
20  74.05 \\
21  74.66 \\
22  74.76 \\
23  75.40 \\
24  75.70 \\
25  75.66 \\
26  76.00 \\
27  76.16 \\
28  76.67 \\
29  76.20 \\
30  76.50 \\
31  76.76 \\
32  76.98 \\
33  76.59 \\
34  77.17 \\
35  77.17 \\
36  77.10 \\
37  77.27 \\
38  77.25 \\
39  77.43 \\
40  77.63 \\
41  77.52 \\
42  77.23 \\
43  77.56 \\
44  78.01 \\
45  78.02 \\
46  77.96 \\
47  78.18 \\
48  78.16 \\
49  78.15 \\
50  78.19 \\
51  78.20 \\
52  78.20 \\
53  78.24 \\
54  78.34 \\
55  78.40 \\
56  78.41 \\
57  78.48 \\
58  78.50 \\
59  78.53 \\
60  78.60 \\
61  78.57 \\
62  78.61 \\
63  78.68 \\
64  78.70 \\
65  78.73 \\
66  78.72 \\
67  78.75 \\
68  78.79 \\
69  78.80 \\
70  78.81 \\
71  78.82 \\
72  78.83 \\
73  78.82 \\
74  78.85 \\
75  78.89 \\
76  79.02 \\
77  78.98 \\
78  79.02 \\
79  79.10 \\
80  79.24 \\
81  79.22 \\
82  79.26 \\
83  79.28 \\
84  79.31 \\
85  79.32 \\
86  79.35 \\
87  79.39 \\
88  79.40 \\
89  79.40 \\
90  79.39 \\
91  79.34 \\
92  79.25 \\
93  79.31 \\
94  79.39 \\
95  79.42 \\
96  79.44 \\
97  79.47 \\
98  79.48 \\
99  79.49 \\
100 79.51 \\
101 79.52 \\
};
\addlegendentry{AdaDelta} 
 \addplot [color=bostonuniversityred, mark = ball, mark options={scale=0.25}, line width = 0.75]
table[row sep=crcr]{%
0    10.0 \\
1    9.92 \\
2    10.0 \\
3    12.26 \\
4    14.45 \\
5    17.6 \\
6    21.42 \\
7    23.25 \\
8    22.93 \\
9    25.24 \\
10   27.47 \\
11   28.65 \\
12   30.35 \\
13   30.81 \\
14   32.11 \\
15   33.41 \\
16   34.67 \\
17   36.12 \\
18   37.7 \\
19   39.53 \\
20   39.78 \\
21   41.07 \\
22   42.55 \\
23   43.14 \\
24   44.16 \\
25   45.41 \\
26   45.7 \\
27   46.06 \\
28   46.94 \\
29   47.7 \\
30   47.71 \\
31   48.44 \\
32   49.43 \\
33   49.99 \\
34   49.97 \\
35   50.31 \\
36   51.39 \\
37   51.56 \\
38   52.02 \\
39   52.33 \\
40   52.73 \\
41   52.82 \\
42   53.61 \\
43   53.91 \\
44   54.28 \\
45   54.8 \\
46   55.05 \\
47   55.75 \\
48   56.29 \\
49   56.74 \\
50   56.86 \\
51   57.61 \\
52   58.02 \\
53   58.35 \\
54   58.8 \\
55   58.71 \\
56   59.14 \\
57   59.65 \\
58   59.98 \\
59   60.21 \\
60   60.64 \\
61   60.83 \\
62   61.04 \\
63   61.32 \\
64   62.04 \\
65   62.33 \\
66   62.56 \\
67   62.59 \\
68   63.03 \\
69   63.28 \\
70   63.77 \\
71   63.97 \\
72   63.76 \\
73   64.12 \\
74   64.39 \\
75   65.01 \\
76   64.89 \\
77   65.08 \\
78   65.43 \\
79   65.53 \\
80   65.79 \\
81   65.91 \\
82   66.29 \\
83   66.29 \\
84   66.35 \\
85   66.46 \\
86   66.59 \\
87   66.94 \\
88   67.08 \\
89   67.21 \\
90   67.14 \\
91   67.56 \\
92   67.8 \\
93   67.82 \\
94   68.04 \\
95   68.21 \\
96   68.4 \\
97   68.16 \\
98   68.54 \\
99   68.52 \\
100  68.93 \\
};
\addlegendentry{SGD} 




\end{axis}

\end{tikzpicture}%
}
\subcaption{$M = 20$, $L = 6$, non-IID.}
\label{subfig: AdaDelta vs SGD M20}
\end{minipage}
\begin{minipage}{0.49\columnwidth}
{\scriptsize\definecolor{amber}{rgb}{1.0, 0.49, 0.0}
\definecolor{taupe}{rgb}{0.28, 0.24, 0.2}
\definecolor{tealgreen}{rgb}{0.0, 0.51, 0.5}
\definecolor{britishracinggreen}{rgb}{0.0, 0.26, 0.15}
\definecolor{cerise}{rgb}{0.87, 0.19, 0.39}
\definecolor{mycolor1}{rgb}{0.00000,0.44700,0.74100}%
\definecolor{mycolor2}{rgb}{0.85000,0.32500,0.09800}%
\definecolor{mycolor3}{rgb}{0.92900,0.69400,0.12500}%
\definecolor{mycolor4}{rgb}{0.49400,0.18400,0.55600}%
\definecolor{bostonuniversityred}{rgb}{0.8, 0.0, 0.0}
\definecolor{brickred}{rgb}{0.8, 0.25, 0.33}
\begin{tikzpicture}

\begin{axis}[%
width=0.8\columnwidth,
height=0.4\columnwidth,
at={(0,0)},
scale only axis,
xmin=0,
xmax=100,
ymin=9,
ymax=85,
grid style={dashed},
ymajorgrids,
grid=both,
ylabel near ticks,
ylabel={Test accuracy (\%) },
xlabel={Iteration~$k$},
axis background style={fill=white},
legend style={at={(0.98, 0.02)}, anchor=south east, legend cell align=left, inner sep=0.01pt, font = \scriptsize, text depth=0.25ex, legend image post style={scale=0.75}}, 
legend columns = 1,
every axis plot/.append style={line width=1.0pt}
]
\addplot [color=black, densely dashdotted, line width = 1.75]
  table[row sep=crcr]{%
0   12.82 \\
1   28.23 \\
2   33.79 \\
3   39.04 \\
4   42.63 \\
5   47.12 \\
6   50.54 \\
7   53.07 \\
8   55.70 \\
9   57.31 \\
10  59.45 \\
11  61.26 \\
12  62.51 \\
13  63.30 \\
14  64.86 \\
15  65.66 \\
16  66.37 \\
17  67.61 \\
18  68.11 \\
19  68.70 \\
20  69.52 \\
21  70.07 \\
22  70.77 \\
23  71.06 \\
24  71.12 \\
25  71.75 \\
26  72.14 \\
27  72.42 \\
28  72.69 \\
29  73.18 \\
30  73.24 \\
31  73.27 \\
32  73.78 \\
33  73.96 \\
34  74.45 \\
35  74.30 \\
36  74.65 \\
37  74.81 \\
38  74.95 \\
39  75.06 \\
40  75.20 \\
41  75.50 \\
42  75.68 \\
43  75.65 \\
44  75.70 \\
45  76.11 \\
46  76.19 \\
47  76.34 \\
48  76.28 \\
49  76.42 \\
50  76.53 \\
51  76.61 \\
52  76.69 \\
53  76.73 \\
54  76.73 \\
55  76.73 \\
56  76.83 \\
57  76.85 \\
58  76.87 \\
59  76.89 \\
60  76.74 \\
61  76.80 \\
62  76.89 \\
63  76.91 \\
64  76.75 \\
65  77.01 \\
66  77.08 \\
67  77.021 \\
68  77.15 \\
69  77.15 \\
70  77.07 \\
71  77.19 \\
72  77.20 \\
73  77.22 \\
74  77.30 \\
75  77.30 \\
76  77.36 \\
77  77.39 \\
78  77.43 \\
79  77.52 \\
80  77.54 \\
81  77.54 \\
82  77.58 \\
83  77.61 \\
84  77.61 \\
85  77.64 \\
86  77.65 \\
87  77.66 \\
88  77.66 \\
89  77.68 \\
90  77.70 \\
91  77.71 \\
92  77.71 \\
93  77.72 \\
94  77.72 \\
95  77.73 \\
96  77.80 \\
97  77.81 \\
98  77.83 \\
99  77.85 \\
100 77.92 \\
101 78 \\
};
\addlegendentry{AdaDelta} 
 \addplot [color=brickred, mark = ball, mark options={scale=0.25}, line width = 0.75]
table[row sep=crcr]{%
0    10.13 \\
1    9.94 \\
2    9.74 \\
3    12.13 \\
4    15.68 \\
5    17.17 \\
6    17.34 \\
7    19.04 \\
8    20.15 \\
9    21.71 \\
10   23.68 \\
11   25.92 \\
12   27.06 \\
13   28.7 \\
14   30.06 \\
15   30.49 \\
16   30.96 \\
17   32.28 \\
18   32.98 \\
19   33.8 \\
20   34.56 \\
21   35.07 \\
22   35.84 \\
23   36.33 \\
24   37.07 \\
25   37.75 \\
26   38.46 \\
27   39.07 \\
28   39.49 \\
29   40.09 \\
30   40.57 \\
31   40.81 \\
32   41.23 \\
33   41.56 \\
34   42.07 \\
35   42.44 \\
36   42.94 \\
37   43.38 \\
38   43.53 \\
39   43.81 \\
40   44.13 \\
41   44.57 \\
42   44.98 \\
43   45.44 \\
44   46.01 \\
45   46.33 \\
46   46.66 \\
47   47.04 \\
48   47.52 \\
49   47.57 \\
50   47.96 \\
51   48.3 \\
52   48.69 \\
53   48.98 \\
54   49.31 \\
55   49.5 \\
56   50.1 \\
57   50.35 \\
58   50.6 \\
59   50.91 \\
60   51.4 \\
61   51.73 \\
62   51.76 \\
63   52.09 \\
64   52.58 \\
65   52.64 \\
66   52.95 \\
67   53.27 \\
68   53.38 \\
69   53.57 \\
70   53.72 \\
71   53.82 \\
72   54.05 \\
73   54.51 \\
74   54.52 \\
75   54.96 \\
76   55.07 \\
77   55.4 \\
78   55.68 \\
79   56.08 \\
80   56.29 \\
81   56.42 \\
82   56.78 \\
83   56.82 \\
84   57.06 \\
85   57.16 \\
86   57.59 \\
87   57.59 \\
88   57.88 \\
89   58.12 \\
90   58.25 \\
91   58.37 \\
92   58.75 \\
93   58.75 \\
94   59.02 \\
95   59.16 \\
96   59.49 \\
97   59.63 \\
98   59.88 \\
99   60.26 \\
100  59.89 \\
};
\addlegendentry{SGD } 




\end{axis}

\end{tikzpicture}%
}
\subcaption{$M = 40$, $L = 6$, non-IID.}
\label{subfig: AdaDelta vs SGD M40}
\end{minipage}


   

   

\caption{ Comparison of FedAvg with AdaDelta and SGD local updates.
}
\label{fig: AdaDelta vs SGD}
\end{figure}
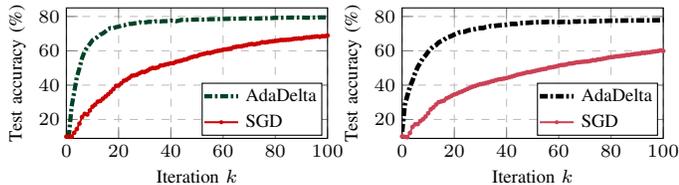


\section{Numerical Results}\label{section: numerical_results}
In this section, we will provide numerical results to demonstrate the performance of our proposed quantization scheme (EMQ) with the FedAvg local AdaDelta updates and our proposed power allocation method.
We consider a CFmMIMO network with $A_p=16$ APs, each with~$N=4$ antennas, a central server, and $M \in \{ 20,40\}$ single-antenna FL clients. Each client is assigned to a $\tau_p$-length pilot sequence according to the pilot assignment algorithm in~\cite{cellfreebook}. The simulation parameters and values are shown in TABLE~\ref{tab: parameters}. We will consider different values for the latency budget~$\bar{\calL}$ (in seconds) and energy budget (in Joule). The CIFAR-10 dataset is distributed among the clients in either an identical and independent distribution (IID) or non-IID manner, and each client trains the following CNN on its local dataset. The CNN architecture in~\cite{CNN_ref} is designed for image classification, consisting of three blocks of convolutional layers with increasing filter sizes (32, 64, 128), each followed by max-pooling and dropout layers to prevent overfitting. After feature extraction, the model is flattened and passed through a dense layer with a softmax activation to classify images into one of 10 categories, resulting in $d = 307,498$. This CNN architecture is chosen for its efficient parameter count ($d = 307,498$), balancing computational efficiency with strong classification accuracy. Its layered convolutional design, combined with max-pooling and dropout, enhances feature extraction and generalization, making it effective for image classification while minimizing overfitting and resource demands. In the following subsections, we evaluate the performance of our proposed EMQ, power allocation, and Algorithms~\ref{alg: local adadelta}-\ref{alg: FL+EMQ+power} considering the latency budget $\bar{\calL}$ (in second) and energy budget $\calE$ (in Joule).
\subsection{Comparison between FedAvg with AdaDelta and SGD}
This subsection evaluates the test accuracy of FL training using AdaDelta and SGD for local updates. Fig.~\ref{fig: AdaDelta vs SGD} presents test accuracy results for $M = 20$ clients in Fig.~\ref{subfig: AdaDelta vs SGD M20}, $M = 40$ clients in Fig.~\ref{subfig: AdaDelta vs SGD M40}, with $L = 6$ local iterations, $K = 100$ global iterations, and non-IID data distribution. In both scenarios (\(M = 20\) and \(M = 40\)), AdaDelta reaches a test accuracy of \(78\%\) in fewer than 39 iterations, showcasing its convergence efficiency. This performance is at least three times faster than that of SGD, attributed to AdaDelta's use of a per-parameter adaptive learning rate.


\subsection{Comparison between FedAvg with and without EMQ}
This subsection evaluates the convergence of the FedAvg with AdaDelta local updates training with the EMQ scheme from Algorithm~\ref{alg: local adadelta} and compares it to the FedAvg with full precision, as shown in Fig.~\ref{fig: Accs20} and Fig.~\ref{fig: Accs40}. We consider $M = 20, 40$, both IID and non-IID data distribution among the clients, the number of FedAvg local iterations $L = 2, 6$ corresponding to the minimum and maximum of the EMQ local iterations $l_k^j$, $j \in [M]$ and each global iteration $k \in [100]$. 

Figs.~\ref{subfig: M20iid_accs} and~\ref{subfig: M20non-iid_accs} compare the test accuracies, while Figs.~\ref{subfig: M20iid_locals}, and~\ref{subfig: M20iid_locals} at iteration~$k = 100$, for $M=20$, IID and non-IID distribution, respectively. We observe that in FedAvg+EMQ, although more than $55$\% of the clients have $l_k^j = 2$, the test accuracy outperforms the FedAvg with $L = 2$. This is an important result highlighting that the EMQ quantization error has a negligible effect on the training, which can be compensated with an adaptive number of local iterations for a few clients, thus achieving a better test accuracy than the full precision FedAvg (assuming 32 bits) with $L = 2$ while reducing the communication overhead due to the EMQ scheme. Moreover, the FedAvg+EMQ scheme achieves a comparable test accuracy compared to the full precision FedAvg with $L = 6$ at most after iteration $k = 40$ in the IID case and $k = 50$ in the non-IID case, resulting in saving at least $49$\% (IID) and $56$\% (non-IID) computation resources.

Similar arguments hold for $M = 40$, and the test accuracy comparison is shown in  Figs.~\ref{subfig: M40iid_accs} and~\ref{subfig: M40non-iid_accs} and the local iterations in Figs.~\ref{subfig: M40iid_locals}, and~\ref{subfig: M40iid_locals}. For both IID and non-IID cases, FedAvg+EMQ saves at least $56$\% of computation resources compared to the full precision FedAvg with $L = 6$.

\subsection{Power allocation with different $\theta_E$ and $\theta_l$}
In this subsection, we analyze the performance of the power allocation Algorithm~\ref{alg: power allocation} vs. the scalarization weights $\theta_E$ and $\theta_l$ for different energy and latency budgets while considering FedAvg+EMQ scheme, summarized in Algorithm~\ref{alg: FL+EMQ+power}. To this aim, Fig.~\ref{fig: theta_M20} and Fig.~\ref{fig: theta_M40} show the test accuracy and the stop iteration $K$ for non-IID data distribution between $M = 20$ and $M = 40$ clients, respectively. 
Fig.~\ref{subfig: theta_E_M20} shows the test accuracy vs. $\theta_E$ for $\theta_l = 0.5$ according to the corresponding stop iteration $K$ in Fig.~\ref{subfig: theta_E_M20_K}. We observe that the optimal value of $\theta_E$ depends on the values of the energy budget $\calE$ and latency budget $\bar{\calL}$. In the case of a fixed $\theta_l = 0.5$, the test accuracy and $K$ are non-decreasing with respect to $\theta_E$. Thus, higher values of $\theta_E$ are recommended for achieving a higher $K$ and consequently a better test accuracy. A similar argument holds for the test accuracy of Fig.~\ref{subfig: theta_E_M40} and the stop iteration $K$ of Fig~\ref{subfig: theta_E_M40_K} for $M = 40$ clients.

 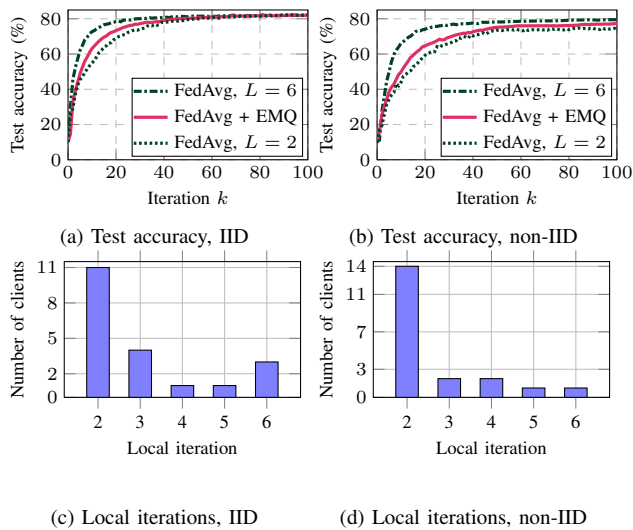
\begin{figure}[t]
\centering
\begin{minipage}{0.45\columnwidth}
{\scriptsize\definecolor{amber}{rgb}{1.0, 0.49, 0.0}
\definecolor{taupe}{rgb}{0.28, 0.24, 0.2}
\definecolor{tealgreen}{rgb}{0.0, 0.51, 0.5}
\definecolor{britishracinggreen}{rgb}{0.0, 0.26, 0.15}
\definecolor{cerise}{rgb}{0.87, 0.19, 0.39}
\definecolor{mycolor1}{rgb}{0.00000,0.44700,0.74100}%
\definecolor{mycolor2}{rgb}{0.85000,0.32500,0.09800}%
\definecolor{mycolor3}{rgb}{0.92900,0.69400,0.12500}%
\definecolor{mycolor4}{rgb}{0.49400,0.18400,0.55600}%
\begin{tikzpicture}

\begin{axis}[%
width=0.8\columnwidth,
height=0.5\columnwidth,
at={(0,0)},
scale only axis,
xmin=0,
xmax=100,
ymin=0,
ymax=85,
grid style={dashed},
ymajorgrids,
grid=both,
ylabel near ticks,
ylabel={Test accuracy (\%) },
xlabel={Iteration~$k$},
axis background style={fill=white},
legend style={at={(0.98, 0.02)}, anchor=south east, legend cell align=left, inner sep=0.01pt, font = \scriptsize, text depth=0.25ex, legend image post style={scale=0.75}}, 
legend columns = 1,
every axis plot/.append style={line width=1.0pt}
]
\addplot [color=britishracinggreen, densely dashdotted, line width = 1.2]
  table[row sep=crcr]{%
0    11.03 \\
1    30.02 \\
2    45.43 \\
3    52.06 \\
4    57.20 \\
5    62.03 \\
6    64.90 \\
7    68.11 \\
8    69.60 \\
9    71.50 \\
10   72.72 \\
11   73.33 \\
12   74.44 \\
13   75.12 \\
14   75.94 \\
15   76.65 \\
16   76.89 \\
17   77.20 \\
18   77.74 \\
19   78.01 \\
20   78.42 \\
21   78.74 \\
22   78.30 \\
23   79.22 \\
24   79.33 \\
25   79.41 \\
26   79.33 \\
27   79.71 \\
28   79.89 \\
29   80.07 \\
30   80.00 \\
31   80.38 \\
32   80.21 \\
33   80.24 \\
34   80.63 \\
35   80.31 \\
36   80.52 \\
37   80.48 \\
38   80.63 \\
39   80.63 \\
40   80.78 \\
41   80.88 \\
42   80.66 \\
43   81.00 \\
44   81.17 \\
45   81.02 \\
46   81.37 \\
47   81.38 \\
48   81.32 \\
49   81.23 \\
50   81.13 \\
51   81.41 \\
52   81.50 \\
53   81.46 \\
54   81.50 \\
55   81.41 \\
56   81.55 \\
57   81.38 \\
58   81.46 \\
59   81.29 \\
60   81.39 \\
61   81.40 \\
62   81.39 \\
63   81.59 \\
64   81.77 \\
65   81.65 \\
66   82.00 \\
67   81.58 \\
68   81.89 \\
69   82.08 \\
70   81.75 \\
71   81.77 \\
72   81.75 \\
73   81.58 \\
74   81.63 \\
75   81.90 \\
76   81.64 \\
77   81.65 \\
78   81.85 \\
79   81.75 \\
80   81.71 \\
81   82.23 \\
82   81.86 \\
83   81.94 \\
84   81.95 \\
85   81.88 \\
86   81.74 \\
87   81.92 \\
88   81.87 \\
89   82.02 \\
90   82.06 \\
91   82.08 \\
92   82.09 \\
93   81.86 \\
94   81.95 \\
95   81.97 \\
96   81.77 \\
97   81.82 \\
98   81.92 \\
99   82.07 \\
100  82.28 \\
101  82.34 \\
};
\addlegendentry{FedAvg, $L = 6$} 
\addplot [color=cerise, line width = 1.2]
  table[row sep=crcr]{%
0   10.14 \\
1   14.71 \\
2   32.92 \\
3   38.96 \\
4   44.00 \\
5   48.92 \\
6   52.54 \\
7   56.00 \\
8   58.28 \\
9   61.00 \\
10  63.11 \\
11  64.61 \\
12  66.05 \\
13  67.42 \\
14  68.60 \\
15  69.61 \\
16  70.27 \\
17  71.46 \\
18  72.27 \\
19  72.65 \\
20  73.38 \\
21  73.88 \\
22  74.42 \\
23  75.10 \\
24  75.43 \\
25  75.94 \\
26  76.25 \\
27  76.41 \\
28  76.94 \\
29  76.85 \\
30  77.26 \\
31  77.49 \\
32  77.87 \\
33  78.09 \\
34  77.93 \\
35  77.94 \\
36  78.16 \\
37  78.49 \\
38  78.26 \\
39  78.79 \\
40  79.01 \\
41  79.27 \\
42  79.46 \\
43  79.55 \\
44  79.78 \\
45  79.76 \\
46  80.09 \\
47  80.00 \\
48  79.98 \\
49  80.12 \\
50  80.36 \\
51  80.21 \\
52  80.59 \\
53  80.24 \\
54  80.56 \\
55  80.56 \\
56  80.79 \\
57  80.74 \\
58  80.70 \\
59  81.17 \\
60  80.80 \\
61  81.03 \\
62  80.86 \\
63  80.90 \\
64  81.27 \\
65  80.95 \\
66  81.13 \\
67  81.36 \\
68  81.57 \\
69  81.64 \\
70  81.42 \\
71  81.49 \\
72  81.64 \\
73  81.55 \\
74  81.28 \\
75  81.67 \\
76  81.28 \\
77  81.70 \\
78  81.78 \\
79  81.67 \\
80  81.73 \\
81  81.59 \\
82  81.99 \\
83  81.62 \\
84  82.17 \\
85  81.62 \\
86  81.77 \\
87  82.05 \\
88  81.63 \\
89  81.87 \\
90  81.78 \\
91  81.82 \\
92  81.95 \\
93  82.02 \\
94  82.19 \\
95  82.11 \\
96  81.93 \\
97  81.89 \\
98  82.10 \\
99  81.86 \\
100 82.17 \\
101 82.23 \\
};
\addlegendentry{FedAvg + EMQ } 

\addplot [color=britishracinggreen, densely dotted, line width = 1.2]
  table[row sep=crcr]{%
0    10.08 \\
1    25.68 \\
2    29.16 \\
3    37.05 \\
4    42.29 \\
5    45.38 \\
6    47.15 \\
7    49.57 \\
8    51.53 \\
9    53.38 \\
10   54.95 \\
11   56.29 \\
12   58.66 \\
13   61.03 \\
14   62.22 \\
15   63.84 \\
16   64.70 \\
17   66.19 \\
18   66.46 \\
19   68.07 \\
20   68.86 \\
21   69.85 \\
22   70.27 \\
23   70.83 \\
24   71.23 \\
25   72.29 \\
26   72.70 \\
27   73.36 \\
28   73.65 \\
29   73.52 \\
30   74.19 \\
31   74.46 \\
32   74.73 \\
33   75.87 \\
34   76.33 \\
35   75.99 \\
36   76.83 \\
37   76.36 \\
38   76.60 \\
39   77.92 \\
40   77.96 \\
41   78.11 \\
42   78.22 \\
43   78.55 \\
44   78.63 \\
45   78.65 \\
46   78.71 \\
47   79.02 \\
48   78.96 \\
49   79.24 \\
50   79.56 \\
51   79.21 \\
52   79.83 \\
53   79.75 \\
54   80.00 \\
55   79.87 \\
56   80.54 \\
57   80.37 \\
58   80.26 \\
59   80.51 \\
60   80.61 \\
61   80.74 \\
62   80.73 \\
63   80.36 \\
64   80.67 \\
65   81.08 \\
66   81.15 \\
67   81.08 \\
68   81.10 \\
69   81.18 \\
70   81.03 \\
71   81.15 \\
72   81.45 \\
73   81.18 \\
74   81.44 \\
75   81.57 \\
76   81.60 \\
77   81.86 \\
78   81.58 \\
79   81.93 \\
80   81.52 \\
81   81.73 \\
82   81.64 \\
83   81.68 \\
84   81.99 \\
85   81.70 \\
86   81.89 \\
87   81.69 \\
88   81.63 \\
89   82.05 \\
90   82.08 \\
91   82.12 \\
92   82.05 \\
93   82.25 \\
94   82.11 \\
95   82.08 \\
96   82.04 \\
97   81.89 \\
98   82.05 \\
99   82.37 \\
100  82.18 \\
};
\addlegendentry{FedAvg, $L = 2$}




\end{axis}

\end{tikzpicture}%
}
\subcaption{Test accuracy, IID}
\label{subfig: M20iid_accs}
\end{minipage}
\begin{minipage}{0.45\columnwidth}
{\scriptsize\definecolor{amber}{rgb}{1.0, 0.49, 0.0}
\definecolor{taupe}{rgb}{0.28, 0.24, 0.2}
\definecolor{tealgreen}{rgb}{0.0, 0.51, 0.5}
\definecolor{britishracinggreen}{rgb}{0.0, 0.26, 0.15}
\definecolor{cerise}{rgb}{0.87, 0.19, 0.39}
\definecolor{mycolor1}{rgb}{0.00000,0.44700,0.74100}%
\definecolor{mycolor2}{rgb}{0.85000,0.32500,0.09800}%
\definecolor{mycolor3}{rgb}{0.92900,0.69400,0.12500}%
\definecolor{mycolor4}{rgb}{0.49400,0.18400,0.55600}%
\begin{tikzpicture}

\begin{axis}[%
width=0.8\columnwidth,
height=0.5\columnwidth,
at={(0,0)},
scale only axis,
xmin=0,
xmax=100,
ymin=0,
ymax=85,
grid style={dashed},
ymajorgrids,
grid=both,
ylabel near ticks,
ylabel={Test accuracy (\%) },
xlabel={Iteration~$k$},
axis background style={fill=white},
legend style={at={(0.98, 0.02)}, anchor=south east, legend cell align=left, inner sep=0.01pt, font = \scriptsize, text depth=0.25ex, legend image post style={scale=0.75}}, 
legend columns = 1,
every axis plot/.append style={line width=1.0pt}
]
\addplot [color=britishracinggreen, densely dashdotted, line width = 1.2]
  table[row sep=crcr]{%
0   10.08 \\
1   11.17 \\
2   27.52 \\
3   36.03 \\
4   44.67 \\
5   49.87 \\
6   55.72 \\
7   59.72 \\
8   62.22 \\
9   64.01 \\
10  66.17 \\
11  67.12 \\
12  68.15 \\
13  69.46 \\
14  70.15 \\
15  71.33 \\
16  72.68 \\
17  73.30 \\
18  73.17 \\
19  73.79 \\
20  74.05 \\
21  74.66 \\
22  74.76 \\
23  75.40 \\
24  75.70 \\
25  75.66 \\
26  76.00 \\
27  76.16 \\
28  76.67 \\
29  76.20 \\
30  76.50 \\
31  76.76 \\
32  76.98 \\
33  76.59 \\
34  77.17 \\
35  77.17 \\
36  77.10 \\
37  77.27 \\
38  77.25 \\
39  77.43 \\
40  77.63 \\
41  77.52 \\
42  77.23 \\
43  77.56 \\
44  78.01 \\
45  78.02 \\
46  77.96 \\
47  78.18 \\
48  78.16 \\
49  78.15 \\
50  78.19 \\
51  78.20 \\
52  78.20 \\
53  78.24 \\
54  78.34 \\
55  78.40 \\
56  78.41 \\
57  78.48 \\
58  78.50 \\
59  78.53 \\
60  78.60 \\
61  78.57 \\
62  78.61 \\
63  78.68 \\
64  78.70 \\
65  78.73 \\
66  78.72 \\
67  78.75 \\
68  78.79 \\
69  78.80 \\
70  78.81 \\
71  78.82 \\
72  78.83 \\
73  78.82 \\
74  78.85 \\
75  78.89 \\
76  79.02 \\
77  78.98 \\
78  79.02 \\
79  79.10 \\
80  79.24 \\
81  79.22 \\
82  79.26 \\
83  79.28 \\
84  79.31 \\
85  79.32 \\
86  79.35 \\
87  79.39 \\
88  79.40 \\
89  79.40 \\
90  79.39 \\
91  79.34 \\
92  79.25 \\
93  79.31 \\
94  79.39 \\
95  79.42 \\
96  79.44 \\
97  79.47 \\
98  79.48 \\
99  79.49 \\
100 79.51 \\
101 79.52 \\
};
\addlegendentry{FedAvg, $L = 6$} 

\addplot [color=cerise, line width = 1.2]
  table[row sep=crcr]{%
0   10.04 \\
1   15.93 \\
2   25.32 \\
3   30.10 \\
4   34.33 \\
5   38.63 \\
6   41.21 \\
7   42.89 \\
8   45.68 \\
9   48.43 \\
10  49.88 \\
11  52.19 \\
12  54.11 \\
13  56.06 \\
14  58.15 \\
15  59.20 \\
16  60.60 \\
17  61.65 \\
18  62.71 \\
19  64.22 \\
20  64.58 \\
21  65.23 \\
22  65.82 \\
23  66.29 \\
24  66.89 \\
25  67.56 \\
26  68.39 \\
27  68.00 \\
28  68.07 \\
29  69.02 \\
30  69.53 \\
31  69.95 \\
32  70.45 \\
33  70.87 \\
34  71.21 \\
35  71.48 \\
36  71.82 \\
37  71.72 \\
38  72.28 \\
39  72.00 \\
40  72.69 \\
41  72.98 \\
42  73.30 \\
43  73.60 \\
44  73.43 \\
45  73.95 \\
46  74.20 \\
47  74.60 \\
48  74.80 \\
49  75.10 \\
50  75.20 \\
51  75.25 \\
52  75.28 \\
53  75.38 \\
54  75.40 \\
55  75.55 \\
56  75.68 \\
57  75.75 \\
58  75.94 \\
59  75.99 \\
60  76.03 \\
61  76.00 \\
62  76.00 \\
63  76.01 \\
64  76.02 \\
65  76.02 \\
66  76.03 \\
67  76.03 \\
68  76.04 \\
69  76.04 \\
70  76.04 \\
71  76.05 \\
72  75.99 \\
73  76.08 \\
74  76.06 \\
75  75.92 \\
76  76.08 \\
77  76.08 \\
78  76.18 \\
79  76.20 \\
80  76.19 \\
81  76.26 \\
82  76.26 \\
83  76.38 \\
84  76.44 \\
85  76.70 \\
86  76.74 \\
87  76.89 \\
88  76.67 \\
89  76.84 \\
90  76.84 \\
91  76.90 \\
92  76.95 \\
93  76.88 \\
94  76.96 \\
95  77.00 \\
96  77.02 \\
97  76.98 \\
98  77.11 \\
99  77.24 \\
100 77.20 \\
101 77.24 \\
};
\addlegendentry{FedAvg + EMQ } 

\addplot [color=britishracinggreen, densely dotted, line width = 1.2]
  table[row sep=crcr]{%
0   10.06 \\
1   17.23 \\
2   19.35 \\
3   25.74 \\
4   31.47 \\
5   34.31 \\
6   36.71 \\
7   38.07 \\
8   43.24 \\
9   42.01 \\
10  44.08 \\
11  47.52 \\
12  48.49 \\
13  49.69 \\
14  50.52 \\
15  52.80 \\
16  53.58 \\
17  54.96 \\
18  56.54 \\
19  57.89 \\
20  59.30 \\
21  60.10 \\
22  60.86 \\
23  61.48 \\
24  61.33 \\
25  64.17 \\
26  63.60 \\
27  64.77 \\
28  65.18 \\
29  65.63 \\
30  65.96 \\
31  66.57 \\
32  67.37 \\
33  67.56 \\
34  69.14 \\
35  68.98 \\
36  69.91 \\
37  69.67 \\
38  68.92 \\
39  70.81 \\
40  70.25 \\
41  71.03 \\
42  71.00 \\
43  71.42 \\
44  72.05 \\
45  71.20 \\
46  71.86 \\
47  72.67 \\
48  73.16 \\
49  72.81 \\
50  73.48 \\
51  73.61 \\
52  73.29 \\
53  73.72 \\
54  73.82 \\
55  73.76 \\
56  73.90 \\
57  73.60 \\
58  74.03 \\
59  73.67 \\
60  73.63 \\
61  73.21 \\
62  74.11 \\
63  73.38 \\
64  73.84 \\
65  73.80 \\
66  73.75 \\
67  73.74 \\
68  73.96 \\
69  73.68 \\
70  73.81 \\
71  73.50 \\
72  74.30 \\
73  74.26 \\
74  73.97 \\
75  74.29 \\
76  73.79 \\
77  73.56 \\
78  73.74 \\
79  74.27 \\
80  73.84 \\
81  74.30 \\
82  74.54 \\
83  74.47 \\
84  74.77 \\
85  74.39 \\
86  74.56 \\
87  74.51 \\
88  74.20 \\
89  74.71 \\
90  74.71 \\
91  74.19 \\
92  73.88 \\
93  73.95 \\
94  74.78 \\
95  74.21 \\
96  74.57 \\
97  74.11 \\
98  74.23 \\
99  74.61 \\
100 74.16 \\
101 74.53 \\
};
\addlegendentry{FedAvg, $L = 2$ }




\end{axis}

\end{tikzpicture}%
}
\subcaption{Test accuracy, non-IID}
\label{subfig: M20non-iid_accs}
\end{minipage}


\begin{minipage}{0.45\columnwidth}
{\scriptsize
\begin{tikzpicture}[scale=1]
    \begin{axis}[
        width=1.185\columnwidth,
        height=0.85\columnwidth,
        ybar,
        xlabel={Local iteration},
        ylabel={Number of clients},
        symbolic x coords=  {2, 3, 4, 5, 6},
        xtick = {2, ..., 6},
        ytick = {0, 2, 5, 8, 11},
        ymin=0, ymax=11.5,  
        bar width=0.3cm,
        enlarge x limits=0.2,
        grid=major
    ]
    \addplot[fill=blue!50] coordinates {
    (2,11) (3,4) (4,1) (5,1) (6,3)
        
    };
    \end{axis}
\end{tikzpicture}
}
\subcaption{Local iterations, IID}
\label{subfig: M20iid_locals}
\end{minipage}
\begin{minipage}{0.45\columnwidth}
{\scriptsize
\begin{tikzpicture}[scale=1]
    \begin{axis}[
        width=1.185\columnwidth,
        height=0.85\columnwidth,
        ybar,
        xlabel={Local iteration},
        ylabel={Number of clients},
        symbolic x coords=  {2, 3, 4, 5, 6},
        xtick = {2, ..., 6},
        ytick = {0, 3, 7, 11, 14},
        ymin=0, ymax=14.5,  
        bar width=0.3cm,
        enlarge x limits=0.2,
        grid=major
    ]
    \addplot[fill=blue!50] coordinates {
    (2,14) (3,2) (4,2) (5,1) (6,1)
       
    };
    \end{axis}
\end{tikzpicture}
}
\subcaption{Local iterations, non-IID}
\label{subfig: M20non-iid_locals}
\end{minipage}

\caption{Convergence analysis of FedAvg with the EMQ scheme and adaptive local iterations compared to FedAvg with full 32-bit precision, and $L = 2, 6$ constant number of local iterations, and $M = 20$ clients. a) and b) Test accuracy. c) and d) Corresponding adaptive number of local iterations, $l_k ^j$ $j \in [M]$, for $k = 100$, obtained from Algorithm~\ref{alg: local adadelta}.  
}
\label{fig: Accs20}
\end{figure}

\begin{figure}[t]
\centering
\begin{minipage}{0.45\columnwidth}
{\scriptsize\definecolor{amber}{rgb}{1.0, 0.49, 0.0}
\definecolor{taupe}{rgb}{0.28, 0.24, 0.2}
\definecolor{tealgreen}{rgb}{0.0, 0.51, 0.5}
\definecolor{britishracinggreen}{rgb}{0.0, 0.26, 0.15}
\definecolor{cerise}{rgb}{0.87, 0.19, 0.39}
\definecolor{mycolor1}{rgb}{0.00000,0.44700,0.74100}%
\definecolor{mycolor2}{rgb}{0.85000,0.32500,0.09800}%
\definecolor{mycolor3}{rgb}{0.92900,0.69400,0.12500}%
\definecolor{mycolor4}{rgb}{0.49400,0.18400,0.55600}%
\definecolor{crimsonglory}{rgb}{0.75, 0.0, 0.2}
\begin{tikzpicture}

\begin{axis}[%
width=0.8\columnwidth,
height=0.5\columnwidth,
at={(0,0)},
scale only axis,
xmin=0,
xmax=100,
ymin=0,
ymax=85,
grid style={dashed},
ymajorgrids,
grid=both,
ylabel near ticks,
ylabel={Test accuracy (\%) },
xlabel={Iteration~$k$},
axis background style={fill=white},
legend style={at={(0.98, 0.02)}, anchor=south east, legend cell align=left, inner sep=0.01pt, font = \scriptsize, text depth=0.25ex, legend image post style={scale=0.75}}, 
legend columns = 1,
every axis plot/.append style={line width=1.0pt}
]
\addplot [color=black, densely dashdotted, line width = 1.2]
  table[row sep=crcr]{%
0   11.84 \\
1   31.55 \\
2   40.40 \\
3   46.49 \\
4   50.99 \\
5   54.31 \\
6   56.47 \\
7   58.72 \\
8   61.22 \\
9   63.42 \\
10  64.82 \\
11  65.95 \\
12  67.50 \\
13  68.52 \\
14  69.46 \\
15  70.20 \\
16  71.12 \\
17  72.04 \\
18  72.54 \\
19  73.09 \\
20  73.63 \\
21  74.11 \\
22  74.29 \\
23  74.95 \\
24  75.48 \\
25  75.97 \\
26  76.13 \\
27  76.36 \\
28  76.78 \\
29  76.92 \\
30  77.16 \\
31  77.49 \\
32  77.66 \\
33  77.74 \\
34  77.99 \\
35  78.41 \\
36  78.53 \\
37  78.57 \\
38  78.92 \\
39  78.92 \\
40  79.10 \\
41  79.04 \\
42  79.14 \\
43  79.28 \\
44  79.50 \\
45  79.58 \\
46  79.61 \\
47  79.68 \\
48  79.80 \\
49  79.89 \\
50  80.05 \\
51  79.90 \\
52  80.12 \\
53  80.17 \\
54  80.47 \\
55  80.55 \\
56  80.49 \\
57  80.62 \\
58  80.69 \\
59  80.60 \\
60  80.65 \\
61  80.79 \\
62  80.70 \\
63  80.84 \\
64  80.73 \\
65  80.91 \\
66  80.73 \\
67  81.04 \\
68  80.91 \\
69  81.29 \\
70  80.91 \\
71  80.91 \\
72  81.03 \\
73  81.14 \\
74  81.20 \\
75  81.22 \\
76  81.11 \\
77  81.42 \\
78  81.25 \\
79  81.17 \\
80  81.43 \\
81  81.47 \\
82  81.66 \\
83  81.63 \\
84  81.65 \\
85  81.73 \\
86  81.63 \\
87  81.51 \\
88  81.72 \\
89  81.75 \\
90  81.69 \\
91  81.61 \\
92  81.62 \\
93  81.61 \\
94  81.92 \\
95  81.85 \\
96  81.93 \\
97  81.90 \\
98  82.00 \\
99  82.06 \\
100 82.03 \\
101 82.05 \\
};
\addlegendentry{FedAvg, $L = 6$} 

\addplot [color=crimsonglory, line width = 1.2]
  table[row sep=crcr]{%
0    10.79 \\
1    19.89 \\
2    29.97 \\
3    35.47 \\
4    40.17 \\
5    42.97142857 \\
6    45.065 \\
7    46.88666667 \\
8    48.84 \\
9    50.71727273 \\
10   52.38 \\
11   53.78923077 \\
12   55.38571429 \\
13   57.09 \\
14   58.2125 \\
15   59.55352941 \\
16   60.42333333 \\
17   61.92210526 \\
18   63.0 \\
19   64.08714286 \\
20   64.84363636 \\
21   66.08956522 \\
22   66.965 \\
23   68.22 \\
24   69.29461538 \\
25   69.70888889 \\
26   69.63285714 \\
27   70.70655172 \\
28   71.78 \\
29   72.70322581 \\
30   72.51625 \\
31   73.34909091 \\
32   73.97176471 \\
33   74.53428571 \\
34   75.08666667 \\
35   75.72891892 \\
36   76.29105263 \\
37   76.55307692 \\
38   77.015 \\
39   76.51682927 \\
40   77.00857143 \\
41   77.10023256 \\
42   76.82181818 \\
43   76.96333333 \\
44   76.91478261 \\
45   77.09617021 \\
46   77.2975 \\
47   77.48877551 \\
48   77.65 \\
49   77.76 \\
50   77.87 \\
51   77.98 \\
52   77.96 \\
53   77.9 \\
54   78.06 \\
55   77.91 \\
56   78.23 \\
57   78.12 \\
58   78.37 \\
59   78.53 \\
60   78.86 \\
61   78.92 \\
62   78.93 \\
63   78.88 \\
64   78.85 \\
65   78.91 \\
66   78.99 \\
67   79.01 \\
68   79.24 \\
69   79.37 \\
70   79.28 \\
71   79.35 \\
72   79.34 \\
73   79.39 \\
74   79.47 \\
75   79.38 \\
76   79.48 \\
77   79.57 \\
78   79.41 \\
79   79.65 \\
80   79.76 \\
81   79.74 \\
82   79.75 \\
83   79.7 \\
84   79.86 \\
85   79.97 \\
86   80.0 \\
87   80.09 \\
88   80.08 \\
89   79.99 \\
90   80.14 \\
91   80.0 \\
92   80.04 \\
93   80.02 \\
94   80.08 \\
95   80.01 \\
96   80.04 \\
97   80.03 \\
98   80.06 \\
99   80.05 \\
100  80.06 \\
};
\addlegendentry{FedAvg + EMQ } 

\addplot [color=black, densely dotted, line width = 1.2]
  table[row sep=crcr]{%
0   11.50 \\
1   26.66 \\
2   28.39 \\
3   29.99 \\
4   31.90 \\
5   34.86 \\
6   37.50 \\
7   40.54 \\
8   42.94 \\
9   45.31 \\
10  46.99 \\
11  48.11 \\
12  49.42 \\
13  50.74 \\
14  51.80 \\
15  52.96 \\
16  54.19 \\
17  55.04 \\
18  56.49 \\
19  57.84 \\
20  58.73 \\
21  59.95 \\
22  61.37 \\
23  62.48 \\
24  63.01 \\
25  64.27 \\
26  65.36 \\
27  65.85 \\
28  66.50 \\
29  66.87 \\
30  67.31 \\
31  68.00 \\
32  68.36 \\
33  69.07 \\
34  69.52 \\
35  70.20 \\
36  70.44 \\
37  70.91 \\
38  71.35 \\
39  71.85 \\
40  72.07 \\
41  72.67 \\
42  73.44 \\
43  73.80 \\
44  73.98 \\
45  74.39 \\
46  74.74 \\
47  74.68 \\
48  75.13 \\
49  75.31 \\
50  75.49 \\
51  75.68 \\
52  75.79 \\
53  75.91 \\
54  75.98 \\
55  76.02 \\
56  75.93 \\
57  76.089 \\
58  75.98 \\
59  76.26 \\
60  76.01 \\
61  75.99 \\
62  76.05 \\
63  76.00 \\
64  76.07 \\
65  76.08 \\
66  76.10 \\
67  76.10 \\
68  76.12 \\
69  76.11 \\
70  76.13 \\
71  76.19 \\
72  76.13 \\
73  76.25 \\
74  76.11 \\
75  76.18 \\
76  76.19 \\
77  76.30 \\
78  75.981 \\
79  76.27 \\
80  76.29 \\
81  76.42 \\
82  76.06 \\
83  76.42 \\
84  76.24 \\
85  76.43 \\
86  76.51 \\
87  76.54 \\
88  76.47 \\
89  76.50 \\
90  76.12 \\
91  76.57 \\
92  76.65 \\
93  76.58 \\
94  76.72 \\
95  76.28 \\
96  76.84 \\
97  76.85 \\
98  76.84 \\
99  76.89 \\
100 76.97 \\
};
\addlegendentry{FedAvg, $L = 2$ }




\end{axis}

\end{tikzpicture}%

}
\subcaption{Test accuracy, IID}
\label{subfig: M40iid_accs}
\end{minipage}
\begin{minipage}{0.45\columnwidth}
{\scriptsize\definecolor{amber}{rgb}{1.0, 0.49, 0.0}
\definecolor{taupe}{rgb}{0.28, 0.24, 0.2}
\definecolor{tealgreen}{rgb}{0.0, 0.51, 0.5}
\definecolor{britishracinggreen}{rgb}{0.0, 0.26, 0.15}
\definecolor{cerise}{rgb}{0.87, 0.19, 0.39}
\definecolor{mycolor1}{rgb}{0.00000,0.44700,0.74100}%
\definecolor{mycolor2}{rgb}{0.85000,0.32500,0.09800}%
\definecolor{mycolor3}{rgb}{0.92900,0.69400,0.12500}%
\definecolor{mycolor4}{rgb}{0.49400,0.18400,0.55600}%
\definecolor{crimsonglory}{rgb}{0.75, 0.0, 0.2}
\begin{tikzpicture}

\begin{axis}[%
width=0.8\columnwidth,
height=0.5\columnwidth,
at={(0,0)},
scale only axis,
xmin=0,
xmax=100,
ymin=0,
ymax=85,
grid style={dashed},
ymajorgrids,
grid=both,
ylabel near ticks,
ylabel={Test accuracy (\%) },
xlabel={Iteration~$k$},
axis background style={fill=white},
legend style={at={(0.98, 0.02)}, anchor=south east, legend cell align=left, inner sep=0.01pt, font = \scriptsize, text depth=0.25ex, legend image post style={scale=0.75}}, 
legend columns = 1,
every axis plot/.append style={line width=1.0pt}
]
\addplot [color=black, densely dashdotted, line width = 1.2]
  table[row sep=crcr]{%
0   12.82 \\
1   28.23 \\
2   33.79 \\
3   39.04 \\
4   42.63 \\
5   47.12 \\
6   50.54 \\
7   53.07 \\
8   55.70 \\
9   57.31 \\
10  59.45 \\
11  61.26 \\
12  62.51 \\
13  63.30 \\
14  64.86 \\
15  65.66 \\
16  66.37 \\
17  67.61 \\
18  68.11 \\
19  68.70 \\
20  69.52 \\
21  70.07 \\
22  70.77 \\
23  71.06 \\
24  71.12 \\
25  71.75 \\
26  72.14 \\
27  72.42 \\
28  72.69 \\
29  73.18 \\
30  73.24 \\
31  73.27 \\
32  73.78 \\
33  73.96 \\
34  74.45 \\
35  74.30 \\
36  74.65 \\
37  74.81 \\
38  74.95 \\
39  75.06 \\
40  75.20 \\
41  75.50 \\
42  75.68 \\
43  75.65 \\
44  75.70 \\
45  76.11 \\
46  76.19 \\
47  76.34 \\
48  76.28 \\
49  76.42 \\
50  76.53 \\
51  76.61 \\
52  76.69 \\
53  76.73 \\
54  76.73 \\
55  76.73 \\
56  76.83 \\
57  76.85 \\
58  76.87 \\
59  76.89 \\
60  76.74 \\
61  76.80 \\
62  76.89 \\
63  76.91 \\
64  76.75 \\
65  77.01 \\
66  77.08 \\
67  77.021 \\
68  77.15 \\
69  77.15 \\
70  77.07 \\
71  77.19 \\
72  77.20 \\
73  77.22 \\
74  77.30 \\
75  77.30 \\
76  77.36 \\
77  77.39 \\
78  77.43 \\
79  77.52 \\
80  77.54 \\
81  77.54 \\
82  77.58 \\
83  77.61 \\
84  77.61 \\
85  77.64 \\
86  77.65 \\
87  77.66 \\
88  77.66 \\
89  77.68 \\
90  77.70 \\
91  77.71 \\
92  77.71 \\
93  77.72 \\
94  77.72 \\
95  77.73 \\
96  77.80 \\
97  77.81 \\
98  77.83 \\
99  77.85 \\
100 77.92 \\
101 78 \\
};
\addlegendentry{FedAvg, $L = 6$} 

\addplot [color=crimsonglory, line width = 1.2]
  table[row sep=crcr]{%
0   10.32 \\
1   18.09 \\
2   27.75 \\
3   30.99 \\
4   33.94 \\
5   36.24 \\
6   37.60 \\
7   39.53 \\
8   41.57 \\
9   42.95 \\
10  45.82 \\
11  46.80 \\
12  47.71 \\
13  48.38 \\
14  49.59 \\
15  50.54 \\
16  51.60 \\
17  52.03 \\
18  52.77 \\
19  53.46 \\
20  54.41 \\
21  55.25 \\
22  56.20 \\
23  56.99 \\
24  57.51 \\
25  58.76 \\
26  59.37 \\
27  60.29 \\
28  61.10 \\
29  61.92 \\
30  62.40 \\
31  62.79 \\
32  63.94 \\
33  64.12 \\
34  64.96 \\
35  65.49 \\
36  66.29 \\
37  66.58 \\
38  67.13 \\
39  67.52 \\
40  67.96 \\
41  68.26 \\
42  68.48 \\
43  68.83 \\
44  69.04 \\
45  69.36 \\
46  69.81 \\
47  70.020 \\
48  70.52 \\
49  70.79 \\
50  71.10 \\
51  71.40 \\
52  71.76 \\
53  71.81 \\
54  72.05 \\
55  72.13 \\
56  72.15 \\
57  72.29 \\
58  72.30 \\
59  72.31 \\
60  72.36 \\
61  72.38 \\
62  72.43 \\
63  72.63 \\
64  72.69 \\
65  72.83 \\
66  72.88 \\
67  72.97 \\
68  73.09 \\
69  73.18 \\
70  73.27 \\
71  73.20 \\
72  73.25 \\
73  73.39 \\
74  73.39 \\
75  73.49 \\
76  73.52 \\
77  73.62 \\
78  73.62 \\
79  73.76 \\
80  73.77 \\
81  73.78 \\
82  73.81 \\
83  73.93 \\
84  73.95 \\
85  73.96 \\
86  73.98 \\
87  74.09 \\
88  74.11 \\
89  74.12 \\
90  74.12 \\
91  74.13 \\
92  74.13 \\
93  74.14 \\
94  74.18 \\
95  74.19 \\
96  74.22 \\
97  74.26 \\
98  74.27 \\
99  74.27 \\
100 74.29 \\
101  74.32\\
};
\addlegendentry{FedAvg + EMQ } 

\addplot [color=black, densely dotted, line width = 1.2]
  table[row sep=crcr]{%
0   10.00 \\
1   15.09 \\
2   24.84 \\
3   26.90 \\
4   28.78 \\
5   29.92 \\
6   32.79 \\
7   34.72 \\
8   36.95 \\
9   38.15 \\
10  40.15 \\
11  40.87 \\
12  42.94 \\
13  44.28 \\
14  44.92 \\
15  46.00 \\
16  47.07 \\
17  47.93 \\
18  48.60 \\
19  49.69 \\
20  50.51 \\
21  51.08 \\
22  52.09 \\
23  52.62 \\
24  53.58 \\
25  54.34 \\
26  55.52 \\
27  56.21 \\
28  56.90 \\
29  57.32 \\
30  57.67 \\
31  58.02 \\
32  59.19 \\
33  59.76 \\
34  60.31 \\
35  61.20 \\
36  61.65 \\
37  62.07 \\
38  62.89 \\
39  62.84 \\
40  63.83 \\
41  63.94 \\
42  64.36 \\
43  65.19 \\
44  65.46 \\
45  65.75 \\
46  66.33 \\
47  66.43 \\
48  67.05 \\
49  67.02 \\
50  68.00 \\
51  68.29 \\
52  68.53 \\
53  68.82 \\
54  68.71 \\
55  69.08 \\
56  69.25 \\
57  69.21 \\
58  69.22 \\
59  69.45 \\
60  69.68 \\
61  69.51 \\
62  69.79 \\
63  69.88 \\
64  69.63 \\
65  70.46 \\
66  70.03 \\
67  70.40 \\
68  70.85 \\
69  70.63 \\
70  70.98 \\
71  71.05 \\
72  70.87 \\
73  71.03 \\
74  71.18 \\
75  71.29 \\
76  71.49 \\
77  71.23 \\
78  71.44 \\
79  71.48 \\
80  71.69 \\
81  71.66 \\
82  71.83 \\
83  71.62 \\
84  71.75 \\
85  71.98 \\
86  72.08 \\
87  72.00 \\
88  71.93 \\
89  72.10 \\
90  72.05 \\
91  72.08 \\
92  72.01 \\
93  72.13 \\
94  72.10 \\
95  72.10 \\
96  72.13 \\
97  72.15 \\
98  72.16 \\
99  72.15 \\
100 72.12 \\
101 72.16\\
};
\addlegendentry{FedAvg, $L = 2$}




\end{axis}

\end{tikzpicture}%

}
\subcaption{Test accuracy, non-IID}
\label{subfig: M40non-iid_accs}
\end{minipage}
\begin{minipage}{0.46\columnwidth}
{\scriptsize
\begin{tikzpicture}[scale=1]
    \begin{axis}[
        width=1.185\columnwidth,
        height=0.85\columnwidth,
        ybar,
        xlabel={Local iteration},
        ylabel={Number of clients},
        symbolic x coords=  {2, 3, 4, 5, 6},
        xtick = {2, ..., 6},
        ytick = {2, 7, 13, 19, 25},
        ymin=0, ymax=25.5,  
        bar width=0.3cm,
        enlarge x limits=0.2,
        grid=major
    ]
    \addplot[fill=blue!50] coordinates {
    (2,25) (3,8) (4,2) (5,1) (6,4)
        
    };
    \end{axis}
\end{tikzpicture}
}
\subcaption{Local iterations, IID}
\label{subfig: M40iid_locals}
\end{minipage}
\begin{minipage}{0.46 \columnwidth}
{\scriptsize
\begin{tikzpicture}[scale=1]
    \begin{axis}[
        width=1.185\columnwidth,
        height=0.85\columnwidth,
        ybar,
        xlabel={Local iteration},
        ylabel={Number of clients},
        symbolic x coords=  {2, 3, 4, 5, 6},
        xtick = {2, ..., 6},
        ytick = {2, 8, 15, 22, 29},
        ymin=0, ymax=29.5,  
        bar width=0.3cm,
        enlarge x limits=0.2,
        grid=major
    ]
    \addplot[fill=blue!50] coordinates {
    (2,29) (3,4) (4,2) (5,2) (6,3)
        
    };
    \end{axis}
\end{tikzpicture}
}
\subcaption{Local iterations, non-IID}
\label{subfig: M40non-iid_locals}
\end{minipage}

\caption{Convergence analysis of FedAvg with the EMQ scheme and adaptive local iterations compared to FedAvg with full 32-bit precision, and $L = 2, 6$ constant number of local iterations, and $M = 40$ clients. a) and b) Test accuracy. c) and d) Corresponding adaptive number of local iterations, $l_k ^j$ $j \in [M]$, for $k = 100$, obtained from Algorithm~\ref{alg: local adadelta}. 
}
\label{fig: Accs40}
\end{figure}
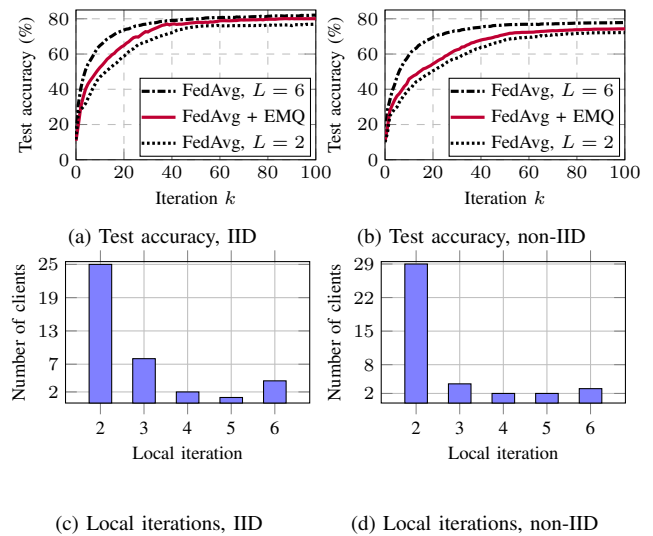

Figs.~\ref{subfig: theta_l_M20} and~\ref{subfig: theta_l_M20_K} show the test accuracy and the corresponding stop iteration $K$ vs. $\theta_l$ for $M = 20$ and $\theta_E = 0.5$. We observe that for a lower energy budget $\calE = 0.01$, increasing $\theta_l$ results in a decreasing $K$ and test accuracy until they become constant due to the limited energy budget. The reason is that when $\theta_l$ is small, and $\theta_E$ is fixed, Algorithm~\ref{alg: power allocation} gives more weight to energy minimization; thus, we obtain a higher $K$ and test accuracy. However, by increasing $\theta_l$, we give the latency minimization more weight, resulting in a less energy-efficient power allocation, which causes a drop in $K$ and test accuracy for lower energy budget situations. The similar arguments hold for $M = 40$ in Figs.~\ref{subfig: theta_l_M40} and~\ref{subfig: theta_l_M40_K}, highlighting the importance of the values of $\calE$ and $\calL$ on the choice of optimal $\theta_E$ and $\theta_l$.


\begin{table*}[t]
\centering
\tiny
\caption{Comparison analysis of EMQ and proposed power allocation of Algorithm~2 with the benchmarks, non-IID $M = 20$, $\calE=1$, $\theta_E = 0.5$, $\theta_l = 1$.}
\renewcommand{\arraystretch}{1.2} 
    \setlength{\extrarowheight}{1.3pt} 
   \setlength{\tabcolsep}{1.3pt}    
     \resizebox{\linewidth}{!}{
\begin{tabular}{c|c ||  c|c|c| c|c|c || c|c|c| c|c|c  || c|c|c| c|c|c }

\hline

\multicolumn{2}{c||}{\textbf{Quantization scheme}}  & \multicolumn{6}{c||}{\textbf{LAQ}} &\multicolumn{6}{c||}{\textbf{AQUILA}} &\multicolumn{6}{c}{\textbf{EMQ}} \\

\hline

\multicolumn{2}{c||}{ \textbf{ Latency budget $\bar{\calL}$} }&  $ 0.5 $ & $ 1 $ & $ 1.5 $ &   $ 2 $ & $ 2.5 $ & $ 3 $ &  $ 0.5 $ & $ 1 $ & $ 1.5 $ &   $ 2 $ & $ 2.5 $ & $ 3 $ & $ 0.5 $ & $ 1 $ & $ 1.5 $ &   $ 2 $ & $ 2.5 $ & $ 3 $   \\ 
\hline
\multirow{2}{*}{\textbf{Algorithm~2}} &  \textbf{Accuracy (\%)} & $25.22$ & $43.8$ & $51.23$ & $56.4$ & $60.4$ & $63.3$ 

& $34.25$& $46.14$ & $54.64$ & $59.65$ & $63.48$ & $66.46$

& \cellcolor{yellow}$49.88$ & \cellcolor{yellow}$62.71$ & \cellcolor{yellow}$68$ & \cellcolor{yellow}$71.82$ & \cellcolor{yellow}$73.43$ & \cellcolor{yellow}$75.28$  \\
                        & $K$  & $6$ & $13$ & $19$ & $26$ & $32$ & $39
 $& $8$& $14$ & $21$ & $27$ & $34$ & $41$ & 
 $\cellcolor{yellow}10$ & $\cellcolor{yellow}18$ & $\cellcolor{yellow}27$ & $\cellcolor{yellow}36$ & $\cellcolor{yellow}44$ & $\cellcolor{yellow}52$   \\
\hline
\multirow{2}{*}{\textbf{Dinkelbach}} &  \textbf{Accuracy (\%)} &  $25.22$ & $41.6$ & $49.7$ & $54$ & $56.67$ & $63.1$ &

$27.2$& $41.92$ & $52$ & $58.55$ & $62.44$ & $65.3$

& $45.68$ & $59.2$ & $65.82$ & $69.02$ & $71.72$ & $73.6$  \\
& $K$  &  $6$ & $12$ & $18$ & $24$ & $30$ & $38$ 
                        & $6$& $12$ & $19$ & $25$ & $32$ & $39$
                        &  $8$ & $15$ & $22$ & $29$ & $37$ & $43$   \\
\hline
\multirow{2}{*}{\textbf{Max-sum rate}} &  \textbf{Accuracy (\%)} & $10.8$ & $22.63$ & $25.22$ & $34.65$ & $39.76$ & $43.8$ &

$21.26$ & $25.64$ & $34.35$ & $37.14$ & $41.92$ & $47.24$

& $30.1$ & $42.89$ & $49.88$ & $58.15$ & $61.65$ & $64.58$  \\
                        & $K$  & $2$ & $4$ & $6$ & $9$ & $11$ & $13$ &
                        
                   $3$ & $5$ & $8$ & $10$ & $12$ & $15$
                    
                    & $3$ & $7$ & $10$ & $14$ & $17$ & $20$   \\
\hline

\end{tabular} 
}

\label{tab: noniid20_ELL_vs_benchmarks}
\end{table*}

\begin{table*}[t]
\centering
\caption{Comparison analysis of EMQ and proposed power allocation of Algorithm~2 with the benchmarks, non-IID $M = 20$, $\bar{\calL}=4$, $\theta_E = 0.5$, $\theta_l = 1$.}
\renewcommand{\arraystretch}{1.2} 
    \setlength{\extrarowheight}{1.3pt} 
   \setlength{\tabcolsep}{1.3pt}    
     \resizebox{\linewidth}{!}{
\begin{tabular}{c|c ||c|c|c| c|c|c|| c|c|c| c|c|c|| c|c|c| c|c|c }

\hline

\multicolumn{2}{c||}{\textbf{Quantization scheme}}  & \multicolumn{6}{c||}{\textbf{LAQ}} &\multicolumn{6}{c||}{\textbf{AQUILA}} &\multicolumn{6}{c}{\textbf{EMQ}} \\

\hline

\multicolumn{2}{c||}{ \textbf{ Energy budget $\calE$} }& $0.05$ &~$0.075$~& $0.1$ &   $0.15$ & $0.2$ & $0.25$ & $0.05$ & $0.075$ & $0.1$ &   $0.15$ & $0.2$ & $0.25$ & $0.05$ & $0.075$ & $0.1$ &   $0.15$ & $0.2$ & $0.25$  \\ 
\hline
\multirow{2}{*}{\textbf{Algorithm~2}} &  \textbf{Accuracy (\%)} & $34.65$ & $44.73$ & $49.7$ & $58.7$ & $62.6$ & $65.58$  &

$48.62$ & $54.64$ & $62.44$ & $67.09$ & $67.09$ & $67.09$

& \cellcolor{yellow}$66.89$  & \cellcolor{yellow}$72.28$ &\cellcolor{yellow} $75.1$  & \cellcolor{yellow}$76.1$ & \cellcolor{yellow}$76.1$  & \cellcolor{yellow}$76.1 $    \\
                        & $K$  & $9$ & $14$ & $18$ & $28$ & $37$ & $46 $ &
                        $16$ & $21$ & $32$ & $44$ & $44$ & $44$ 
                        
                        & \cellcolor{yellow}$24$  & \cellcolor{yellow}$38$ & \cellcolor{yellow}$49$  & \cellcolor{yellow}$69$ & \cellcolor{yellow}$69$  & \cellcolor{yellow}$69 $    \\
\hline
\multirow{2}{*}{\textbf{Dinkelbach}} &  \textbf{Accuracy (\%)} &$10.8 $ & $16.7$ & $22.63$ &  $25.22$ & $34.65$ & $39.76$ &

$21.26$ & $24.33$ & $27.2$ & $34.35$ & $37.14$ & $41.92$

& $30.1$ & $38.63$ & $42.89$ & $49.88$  & $58.15$ & $61.65$   \\
                        & $K$  &$2 $ & $3$ & $4$ &  $6$ & $9$ & $11$ &
                        $3$ & $4$ & $6$ & $8$ & $10$ & $12$
                        & $3$ & $5$ & $7$ & $10$  & $14$ & $17$  \\
\hline
\multirow{2}{*}{\textbf{Max-sum rate}} &  \textbf{Accuracy (\%)} &$10.8$ & $22.63$ & $24.9$ & $31.6$ & $39.76$ & $43.8$ &

$24.33$ &$25.64$& $27.2$ & $36.33$ & $41.92$ & $47.24$

& $34.33$ & $41.21$ & $45.68$ & $56.06$ & $61.65$ & $65.23$\\
                        & $K$  &$2$ & $4$ & $5$ & $8$ & $11$ & $13$ &
                       $4$ &$5$& $6$ & $9$ & $12$ & $15$
                        & $4$ & $6$ & $8$ & $13$ & $17$ & $21$   \\
\hline

\end{tabular} 
}

\label{tab: noniid20_E_vs_benchmarks}
\vspace{-0.015\textheight}
\end{table*}

\subsection{Power allocation and benchmark comparisons}
 In this subsection, we analyze the test accuracy and $K$ obtained from FedAvg+EMQ of Algorithm~\ref{alg: FL+EMQ+power}, focusing on the power allocation of Algorithm~\ref{alg: power allocation}. Furthermore, we compare the test accuracy and $K$ after applying power allocation of Algorithm~\ref{alg: power allocation} with two power allocation benchmarks: the max-min energy efficiency of Dinkelbach~\cite{EE_dinkel} and the max-sum rate~\cite{cellfreebook}. Moreover, we consider two quantization schemes of LAQ~\cite{LAQ} with a fixed number of bits and AQUILA~\cite{AQUILA}, which is an adaptive bit allocation scheme. TABLE~\ref{tab: noniid20_ELL_vs_benchmarks} considers FedAvg with AdaDelta local updates, $\calE = 1$, $M = 20$, non-IID data distribution, $\theta_E = 0.5$, $\theta_l=1$, and different latency budgets to compare the corresponding test accuracy and $K$ results. For the sake of a fair comparison, we define $\bar{b}:= [\E_{k,j} \bb_k^j]$ as the number of bits assigned to the LAQ scheme. Furthermore, for LAQ and AQUILA, we consider the number of local iterations similar to the adaptive $l_k^j$, $\forall j, k$ of EMQ. Comparing the results in TABLE~\ref{tab: noniid20_ELL_vs_benchmarks} for each latency budget, we observe that for all three quantization approaches, our proposed power allocation scheme in Algorithm~\ref{alg: power allocation} outperforms the Dinkelbach and max-sum rate methods by increasing the test accuracy up to $7$\% and $19$\%, respectively, as a consequence of increasing $K$. Moreover, for every power allocation method, the EMQ scheme outperforms the AQUILA and LAQ by increasing the test accuracy up to $19$\% and $24$\%, respectively. Similarly, TABLE~\ref{tab: noniid20_E_vs_benchmarks} compares the test accuracy and $K$ of the for different energy budgets~$\calE$, considering $\bar{\calL} = 4$. We observe that the power allocation in Algorithm~\ref{alg: power allocation} outperforms the Dinkelbach and max-sum rate methods by increasing the test accuracy up to $36$\% and $35$\%, respectively. Furthermore, for every power allocation method, the EMQ scheme outperforms the AQUILA and LAQ by increasing the test accuracy up to $21$\% and $32$\%, respectively. Finally, the FedAvg + EMQ combined with our proposed power allocation of Algorithm~\ref{alg: power allocation} (highlighted rows) achieves the highest test accuracy in the same energy and latency budgets compared to FedAvg + LAQ or AQUILA with Dinkelbach or max-sum rate methods.





\definecolor{amber}{rgb}{1.0, 0.49, 0.0}
\definecolor{taupe}{rgb}{0.28, 0.24, 0.2}
\definecolor{tealgreen}{rgb}{0.0, 0.51, 0.5}
\definecolor{britishracinggreen}{rgb}{0.0, 0.26, 0.15}
\definecolor{cerise}{rgb}{0.87, 0.19, 0.39}
\definecolor{coolblack}{rgb}{0.0, 0.18, 0.39}
\definecolor{denim}{rgb}{0.08, 0.38, 0.74}
\definecolor{crimsonglory}{rgb}{0.75, 0.0, 0.2}
\definecolor{napiergreen}{rgb}{0.16, 0.5, 0.0}
\definecolor{pumpkin}{rgb}{1.0, 0.46, 0.09}

\begin{figure}[t]
\centering

\begin{minipage}{0.49\columnwidth}
    {\scriptsize\definecolor{amber}{rgb}{1.0, 0.49, 0.0}
\definecolor{taupe}{rgb}{0.28, 0.24, 0.2}
\definecolor{tealgreen}{rgb}{0.0, 0.51, 0.5}
\definecolor{britishracinggreen}{rgb}{0.0, 0.26, 0.15}
\definecolor{cerise}{rgb}{0.87, 0.19, 0.39}
\definecolor{mycolor1}{rgb}{0.00000,0.44700,0.74100}%
\definecolor{mycolor2}{rgb}{0.85000,0.32500,0.09800}%
\definecolor{mycolor3}{rgb}{0.92900,0.69400,0.12500}%
\definecolor{mycolor4}{rgb}{0.49400,0.18400,0.55600}%
\begin{tikzpicture}

\begin{axis}[%
width=0.8\columnwidth,
height=0.5\columnwidth,
at={(0,0)},
scale only axis,
xmin=0,
xmax=1,
ymin=0,
ymax=85,
grid style={dashed},
ymajorgrids,
grid=both,
ylabel near ticks,
ylabel={Test accuracy (\%) },
xlabel={$\theta_E$},
axis background style={fill=white},
legend style={at={(0.98, 0.02)}, anchor=south east, legend cell align=left, inner sep=0.01pt, font = \scriptsize, text depth=0.25ex, legend image post style={scale=0.75}}, 
legend columns = 1,
every axis plot/.append style={line width=1.0pt}
]
 

\addplot [color=tealgreen, mark = square*, mark options={scale=0.6}, line width = 0.8]
  table[row sep=crcr]{%
0.001	41.21 \\
0.02	42.89 \\
0.04	45.68 \\
0.2	   59.20 \\
0.4	   68 \\
0.6	   68 \\
0.8	   68 \\
1	  68 \\
};

\addplot [color=blue, mark = pentagon*, mark options={scale=0.8}, line width = 0.8]
  table[row sep=crcr]{%
0.001	41.21 \\
0.02	42.89 \\
0.04	45.68 \\
0.2	   48.43 \\
0.4	   48.43 \\
0.6	   48.43 \\
0.8	   48.43 \\
1	  48.43 \\
};

\addplot [color=black, mark = diamond, mark options={scale=0.8}, line width = 0.8]
  table[row sep=crcr]{%
0.001	10.04 \\
0.02	10.04 \\
0.04	10.04 \\
0.2	   10.04 \\
0.4	   30.10 \\
0.6	   34.33 \\
0.8	   48.43 \\
1	  61.65 \\
};

\end{axis}

\end{tikzpicture}%

 }
    \subcaption{$\theta_l = 0.5$}
    \label{subfig: theta_E_M20}
\end{minipage}
\begin{minipage}{0.49\columnwidth}
    {\scriptsize\definecolor{amber}{rgb}{1.0, 0.49, 0.0}
\definecolor{taupe}{rgb}{0.28, 0.24, 0.2}
\definecolor{tealgreen}{rgb}{0.0, 0.51, 0.5}
\definecolor{britishracinggreen}{rgb}{0.0, 0.26, 0.15}
\definecolor{cerise}{rgb}{0.87, 0.19, 0.39}
\definecolor{mycolor1}{rgb}{0.00000,0.44700,0.74100}%
\definecolor{mycolor2}{rgb}{0.85000,0.32500,0.09800}%
\definecolor{mycolor3}{rgb}{0.92900,0.69400,0.12500}%
\definecolor{mycolor4}{rgb}{0.49400,0.18400,0.55600}%
\begin{tikzpicture}

\begin{axis}[%
width=0.8\columnwidth,
height=0.5\columnwidth,
at={(0,0)},
scale only axis,
xmin=0,
xmax=1,
ymin=0,
ymax=85,
grid style={dashed},
ymajorgrids,
grid=both,
ylabel near ticks,
ylabel={Test accuracy (\%) },
xlabel={$\theta_l$},
axis background style={fill=white},
legend style={at={(0.98, 0.02)}, anchor=south east, legend cell align=left, inner sep=0.01pt, font = \scriptsize, text depth=0.25ex, legend image post style={scale=0.75}}, 
legend columns = 1,
every axis plot/.append style={line width=1.0pt}
]
 

\addplot [color=tealgreen, mark = square*, mark options={scale=0.6}, line width = 0.8]
  table[row sep=crcr]{%
0.001	61.65 \\
0.02	66.29 \\
0.04	67.56 \\
0.2	   67.8 \\
0.4	   68 \\
0.6	   68 \\
0.8	   68 \\
1	  68 \\
};

\addplot [color=blue, mark = pentagon*, mark options={scale=0.8}, line width = 0.8]
  table[row sep=crcr]{%
0.001	34.33 \\
0.02	42.89 \\
0.04	45.68 \\
0.2	   48.43 \\
0.4	   48.43 \\
0.6	   48.43 \\
0.8	   48.43 \\
1	  48.43 \\
};

\addplot [color=black, mark = diamond, mark options={scale=0.8}, line width = 0.8]
  table[row sep=crcr]{%
0.001	10.04 \\
0.02	42.89 \\
0.04	42.89 \\
0.2	   42.89 \\
0.4	   34.33 \\
0.6	   34.3 \\ 
0.8	   34.33 \\
1	  34.33 \\
};

\end{axis}

\end{tikzpicture}%

 }
    \subcaption{$\theta_E = 0.5$}
    \label{subfig: theta_l_M20}
\end{minipage}
\begin{minipage}{0.49\columnwidth}
    {\scriptsize\definecolor{amber}{rgb}{1.0, 0.49, 0.0}
\definecolor{taupe}{rgb}{0.28, 0.24, 0.2}
\definecolor{tealgreen}{rgb}{0.0, 0.51, 0.5}
\definecolor{britishracinggreen}{rgb}{0.0, 0.26, 0.15}
\definecolor{cerise}{rgb}{0.87, 0.19, 0.39}
\definecolor{mycolor1}{rgb}{0.00000,0.44700,0.74100}%
\definecolor{mycolor2}{rgb}{0.85000,0.32500,0.09800}%
\definecolor{mycolor3}{rgb}{0.92900,0.69400,0.12500}%
\definecolor{mycolor4}{rgb}{0.49400,0.18400,0.55600}%
\begin{tikzpicture}

\begin{axis}[%
width=0.8\columnwidth,
height=0.5\columnwidth,
at={(0,0)},
scale only axis,
xmin=0,
xmax=1,
ymin=0,
ymax=30,
grid style={dashed},
ymajorgrids,
grid=both,
ylabel near ticks,
ylabel={Stop iteration~$K$ },
xlabel={$\theta_E$},
axis background style={fill=white},
legend style={at={(0.98, 0.02)}, anchor=south east, legend cell align=left, inner sep=0.01pt, font = \scriptsize, text depth=0.25ex, legend image post style={scale=0.75}}, 
legend columns = 1,
every axis plot/.append style={line width=1.0pt}
]
 

\addplot [color=tealgreen, mark = square*, mark options={scale=0.6}, line width = 0.8]
  table[row sep=crcr]{%
0.001	6\\
0.02	7\\
0.04	8\\
0.2	15\\
0.4	27\\
0.6	27\\
0.8	27\\
1	27\\
};

\addplot [color=blue, mark = pentagon*, mark options={scale=0.8}, line width = 0.8]
  table[row sep=crcr]{%
0.001	6\\
0.02	7\\
0.04	8\\
0.2	9\\
0.4	9\\
0.6	9\\
0.8	9\\
1	  9 \\
};

\addplot [color=black, mark = diamond, mark options={scale=0.8}, line width = 0.8]
  table[row sep=crcr]{%
0.001	1\\
0.02	1\\
0.04	1\\
0.2	1\\
0.4	3\\
0.6	4\\
0.8	9\\
1	  17 \\
};




\end{axis}

\end{tikzpicture}%

 }
    \subcaption{$\theta_l = 0.5$}
    \label{subfig: theta_E_M20_K}
\end{minipage}
\begin{minipage}{0.49\columnwidth}
    {\scriptsize\definecolor{amber}{rgb}{1.0, 0.49, 0.0}
\definecolor{taupe}{rgb}{0.28, 0.24, 0.2}
\definecolor{tealgreen}{rgb}{0.0, 0.51, 0.5}
\definecolor{britishracinggreen}{rgb}{0.0, 0.26, 0.15}
\definecolor{cerise}{rgb}{0.87, 0.19, 0.39}
\definecolor{mycolor1}{rgb}{0.00000,0.44700,0.74100}%
\definecolor{mycolor2}{rgb}{0.85000,0.32500,0.09800}%
\definecolor{mycolor3}{rgb}{0.92900,0.69400,0.12500}%
\definecolor{mycolor4}{rgb}{0.49400,0.18400,0.55600}%
\begin{tikzpicture}

\begin{axis}[%
width=0.8\columnwidth,
height=0.5\columnwidth,
at={(0,0)},
scale only axis,
xmin=0,
xmax=1,
ymin=0,
ymax=30,
grid style={dashed},
ymajorgrids,
grid=both,
ylabel near ticks,
ylabel={Stop iteration~$K$},
xlabel={$\theta_l$},
axis background style={fill=white},
legend style={at={(0.98, 0.02)}, anchor=south east, legend cell align=left, inner sep=0.01pt, font = \scriptsize, text depth=0.25ex, legend image post style={scale=0.75}}, 
legend columns = 1,
every axis plot/.append style={line width=1.0pt}
]
 

\addplot [color=tealgreen, mark = square*, mark options={scale=0.6}, line width = 0.8]
  table[row sep=crcr]{%
0.001	17\\
0.02	23\\
0.04	25\\
0.2	26\\
0.4	27\\
0.6	27\\
0.8	27\\
1	27\\
};

\addplot [color=blue, mark = pentagon*, mark options={scale=0.8}, line width = 0.8]
  table[row sep=crcr]{%
0.001	4\\
0.02	7\\
0.04	8\\
0.2	9\\
0.4	9\\
0.6	9\\
0.8	9\\
1	  9 \\
};

\addplot [color=black, mark = diamond, mark options={scale=0.8}, line width = 0.8]
  table[row sep=crcr]{%
0.001	1\\
0.02	7\\
0.04	7\\
0.2	7\\
0.4	4\\
0.6	4\\ 
0.8	4\\
1	 4\\ 
};

\end{axis}

\end{tikzpicture}%

 }
    \subcaption{$\theta_E = 0.5$}
    \label{subfig: theta_l_M20_K}
\end{minipage}
\begin{tikzpicture}
\begin{axis}[
    hide axis, 
    xmin=0, xmax=1, ymin=0, ymax=1,
    legend columns=3,
    legend style={at={(0.665, -0.01)}, anchor=north, draw=black, font=\scriptsize, /tikz/every even column/.append style={column sep=0.05cm}}
]
\addplot [color=tealgreen, mark = square*, mark options={scale=0.6}] coordinates {(-1,-1)};
\addlegendentry{$\calE = 0.1$, $\bar{\cal L} = 1.5 $}

\addplot [color=blue, mark = pentagon*, mark options={scale=0.8}]coordinates {(-1,-1)};
\addlegendentry{$\calE = 0.1$, $\bar{\cal L} = 0.5 $}

\addplot [color=black, mark = diamond, mark options={scale=0.8}] coordinates {(-1,-1)};
\addlegendentry{$\calE = 0.01$, $\bar{\cal L} = 1.5 $}

\end{axis}
\end{tikzpicture}

\caption{Performance analysis of FedAvg + EMQ with power allocation scheme in Algorithm~\ref{alg: FL+EMQ+power} vs. $\theta_E$ and $\theta_l$, non-IID data distribution for $M = 20$ clients.}
\label{fig: theta_M20}
\end{figure}
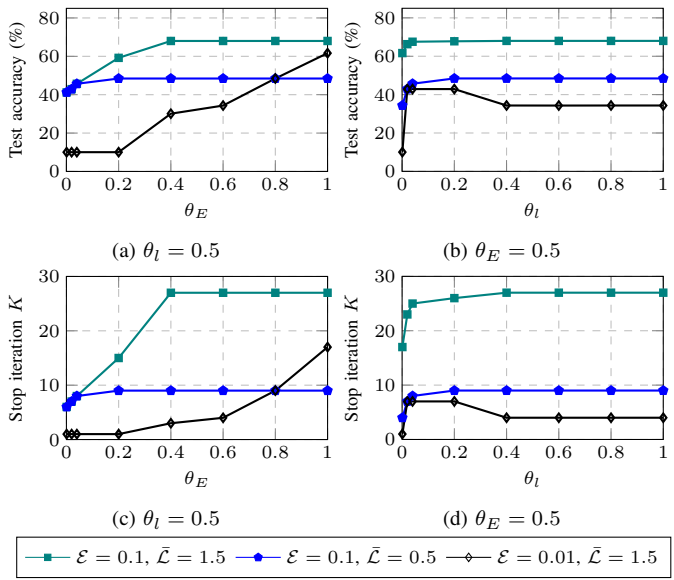

\definecolor{amber}{rgb}{1.0, 0.49, 0.0}
\definecolor{taupe}{rgb}{0.28, 0.24, 0.2}
\definecolor{tealgreen}{rgb}{0.0, 0.51, 0.5}
\definecolor{britishracinggreen}{rgb}{0.0, 0.26, 0.15}
\definecolor{cerise}{rgb}{0.87, 0.19, 0.39}
\definecolor{coolblack}{rgb}{0.0, 0.18, 0.39}
\definecolor{denim}{rgb}{0.08, 0.38, 0.74}
\definecolor{crimsonglory}{rgb}{0.75, 0.0, 0.2}
\definecolor{napiergreen}{rgb}{0.16, 0.5, 0.0}
\definecolor{pumpkin}{rgb}{1.0, 0.46, 0.09}
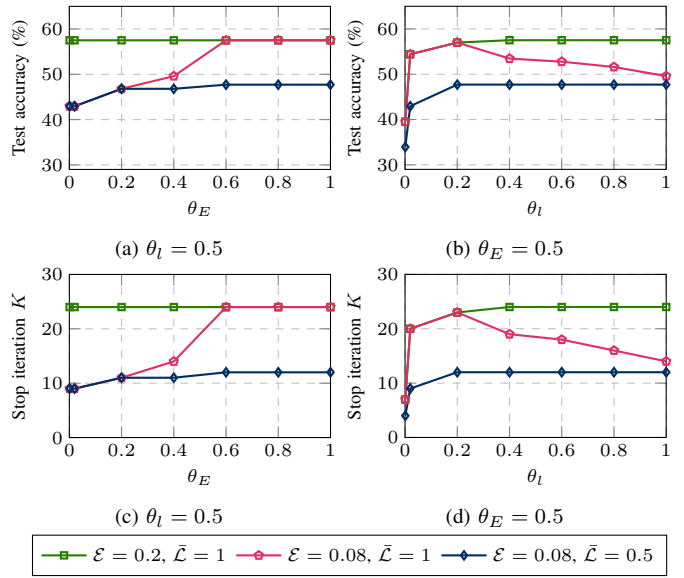
\begin{figure}[t]
\centering
\begin{minipage}{0.49\columnwidth}
{\scriptsize\definecolor{amber}{rgb}{1.0, 0.49, 0.0}
\definecolor{taupe}{rgb}{0.28, 0.24, 0.2}
\definecolor{tealgreen}{rgb}{0.0, 0.51, 0.5}
\definecolor{britishracinggreen}{rgb}{0.0, 0.26, 0.15}
\definecolor{cerise}{rgb}{0.87, 0.19, 0.39}
\definecolor{coolblack}{rgb}{0.0, 0.18, 0.39}
\definecolor{denim}{rgb}{0.08, 0.38, 0.74}
\definecolor{crimsonglory}{rgb}{0.75, 0.0, 0.2}
\definecolor{napiergreen}{rgb}{0.16, 0.5, 0.0}
\definecolor{pumpkin}{rgb}{1.0, 0.46, 0.09}

\begin{tikzpicture}

\begin{axis}[%
width=0.8\columnwidth,
height=0.5\columnwidth,
at={(0,0)},
scale only axis,
xmin=0,
xmax=1,
ymin=29,
ymax=65,
grid style={dashed},
ymajorgrids,
grid=both,
ylabel near ticks,
ylabel={Test accuracy (\%) },
xlabel={$\theta_E$},
axis background style={fill=white},
legend style={at={(0.98, 0.02)}, anchor=south east, legend cell align=left, inner sep=0.01pt, font = \scriptsize, text depth=0.25ex, legend image post style={scale=0.75}}, 
legend columns = 1,
every axis plot/.append style={line width=1.0pt}
]
 

\addplot [color=napiergreen, mark = square, mark options={scale=0.6}, line width = 0.8]
  table[row sep=crcr]{%
0.001	 57.51\\
0.02	57.51\\
0.2	   57.51 \\
0.4	   57.51 \\
0.6	   57.51 \\
0.8	   57.51\\
1	     57.51 \\
};

\addplot [color=cerise, mark = pentagon, mark options={scale=0.8}, line width = 0.8]
  table[row sep=crcr]{%
0.001	  42.95 \\
0.02	 42.95 \\
0.2	   46.80 \\
0.4	   49.59 \\
0.6	   57.51 \\
0.8	   57.51\\
1	    57.51 \\
};

\addplot [color=coolblack, mark = diamond, mark options={scale=0.8}, line width = 0.8]
  table[row sep=crcr]{%
0.001	 42.95\\
0.02	42.95\\
0.2	   46.8 \\
0.4	   46.8 \\
0.6	   47.71 \\
0.8	   47.71 \\
1	      47.71 \\
};

\end{axis}

\end{tikzpicture}%

}
\subcaption{$\theta_l = 0.5$}
\label{subfig: theta_E_M40}
\end{minipage}
\begin{minipage}{0.49\columnwidth}
{\scriptsize\definecolor{amber}{rgb}{1.0, 0.49, 0.0}
\definecolor{taupe}{rgb}{0.28, 0.24, 0.2}
\definecolor{tealgreen}{rgb}{0.0, 0.51, 0.5}
\definecolor{britishracinggreen}{rgb}{0.0, 0.26, 0.15}
\definecolor{cerise}{rgb}{0.87, 0.19, 0.39}
\definecolor{coolblack}{rgb}{0.0, 0.18, 0.39}
\definecolor{denim}{rgb}{0.08, 0.38, 0.74}
\definecolor{crimsonglory}{rgb}{0.75, 0.0, 0.2}
\definecolor{napiergreen}{rgb}{0.16, 0.5, 0.0}
\definecolor{pumpkin}{rgb}{1.0, 0.46, 0.09}
\begin{tikzpicture}

\begin{axis}[%
width=0.8\columnwidth,
height=0.5\columnwidth,
at={(0,0)},
scale only axis,
xmin=0,
xmax=1,
ymin=29,
ymax=65,
grid style={dashed},
ymajorgrids,
grid=both,
ylabel near ticks,
ylabel={Test accuracy (\%) },
xlabel={$\theta_l$},
axis background style={fill=white},
legend style={at={(0.98, 0.02)}, anchor=south east, legend cell align=left, inner sep=0.01pt, font = \scriptsize, text depth=0.25ex, legend image post style={scale=0.75}}, 
legend columns = 1,
every axis plot/.append style={line width=1.0pt}
]
 

\addplot [color=napiergreen, mark = square, mark options={scale=0.6}, line width = 0.8]
  table[row sep=crcr]{%
0.001	39.53 \\
0.02	54.41 \\
0.2	   56.99 \\
0.4	   57.51 \\
0.6	   57.51 \\
0.8	   57.51 \\
1	  57.51 \\
};

\addplot [color=cerise, mark = pentagon, mark options={scale=0.8}, line width = 0.8]
  table[row sep=crcr]{%
0.001	39.53\\
0.02	54.41 \\
0.2	   56.99\\
0.4	   53.46 \\
0.6	   52.77 \\
0.8	   51.6 \\
1	  49.59 \\
};

\addplot [color=coolblack, mark = diamond, mark options={scale=0.8}, line width = 0.8]
  table[row sep=crcr]{%
0.001	33.94\\
0.02	42.95 \\
0.2	   47.71 \\
0.4	   47.71 \\
0.6	   47.71 \\
0.8	   47.71 \\
1	  47.71 \\
};

\end{axis}

\end{tikzpicture}%

}
\subcaption{$\theta_E = 0.5$}
\label{subfig: theta_l_M40}
\end{minipage}
\begin{minipage}{0.49\columnwidth}
{\scriptsize\definecolor{amber}{rgb}{1.0, 0.49, 0.0}
\definecolor{taupe}{rgb}{0.28, 0.24, 0.2}
\definecolor{tealgreen}{rgb}{0.0, 0.51, 0.5}
\definecolor{britishracinggreen}{rgb}{0.0, 0.26, 0.15}
\definecolor{cerise}{rgb}{0.87, 0.19, 0.39}
\definecolor{coolblack}{rgb}{0.0, 0.18, 0.39}
\definecolor{denim}{rgb}{0.08, 0.38, 0.74}
\definecolor{crimsonglory}{rgb}{0.75, 0.0, 0.2}
\definecolor{napiergreen}{rgb}{0.16, 0.5, 0.0}
\definecolor{pumpkin}{rgb}{1.0, 0.46, 0.09}
\begin{tikzpicture}

\begin{axis}[%
width=0.8\columnwidth,
height=0.5\columnwidth,
at={(0,0)},
scale only axis,
xmin=0,
xmax=1,
ymin=0,
ymax=30,
grid style={dashed},
ymajorgrids,
grid=both,
ylabel near ticks,
ylabel={Stop iteration~$K$ },
xlabel={$\theta_E$},
axis background style={fill=white},
legend style={at={(0.98, 0.02)}, anchor=south east, legend cell align=left, inner sep=0.01pt, font = \scriptsize, text depth=0.25ex, legend image post style={scale=0.75}}, 
legend columns = 1,
every axis plot/.append style={line width=1.0pt}
]
 

\addplot [color=napiergreen, mark = square, mark options={scale=0.6}, line width = 0.8]
  table[row sep=crcr]{%
0.001	24\\
0.02	24\\
0.2	    24\\
0.4	    24\\
0.6	    24\\
0.8	    24\\
1	    24\\
};

\addplot [color=cerise, mark = pentagon, mark options={scale=0.8}, line width = 0.8]
  table[row sep=crcr]{%
0.001	9\\
0.02	9\\
0.2	  11\\
0.4	   14 \\
0.6	    24\\
0.8	    24\\
1	    24 \\
};

\addplot [color=coolblack, mark = diamond, mark options={scale=0.8}, line width = 0.8]
  table[row sep=crcr]{%
0.001	9\\
0.02	9\\
0.2	    11\\
0.4	   11 \\
0.6	    12 \\
0.8	     12\\
1	    12\\
};

\end{axis}

\end{tikzpicture}%

}
\subcaption{$\theta_l = 0.5$}
\label{subfig: theta_E_M40_K}
\end{minipage}
\begin{minipage}{0.49\columnwidth}
{\scriptsize\definecolor{amber}{rgb}{1.0, 0.49, 0.0}
\definecolor{taupe}{rgb}{0.28, 0.24, 0.2}
\definecolor{tealgreen}{rgb}{0.0, 0.51, 0.5}
\definecolor{britishracinggreen}{rgb}{0.0, 0.26, 0.15}
\definecolor{cerise}{rgb}{0.87, 0.19, 0.39}
\definecolor{coolblack}{rgb}{0.0, 0.18, 0.39}
\definecolor{denim}{rgb}{0.08, 0.38, 0.74}
\definecolor{crimsonglory}{rgb}{0.75, 0.0, 0.2}
\definecolor{napiergreen}{rgb}{0.16, 0.5, 0.0}
\definecolor{pumpkin}{rgb}{1.0, 0.46, 0.09}
\begin{tikzpicture}

\begin{axis}[%
width=0.8\columnwidth,
height=0.5\columnwidth,
at={(0,0)},
scale only axis,
xmin=0,
xmax=1,
ymin=0,
ymax=30,
grid style={dashed},
ymajorgrids,
grid=both,
ylabel near ticks,
ylabel={Stop iteration~$K$ },
xlabel={$\theta_l$},
axis background style={fill=white},
legend style={at={(0.98, 0.02)}, anchor=south east, legend cell align=left, inner sep=0.01pt, font = \scriptsize, text depth=0.25ex, legend image post style={scale=0.75}}, 
legend columns = 1,
every axis plot/.append style={line width=1.0pt}
]
 

\addplot [color=napiergreen, mark = square, mark options={scale=0.6}, line width = 0.8]
  table[row sep=crcr]{%
0.001	7\\
0.02	20\\
0.2	    23\\
0.4	    24\\
0.6	    24\\
0.8	    24\\
1	    24\\
};

\addplot [color=cerise, mark = pentagon, mark options={scale=0.8}, line width = 0.8]
  table[row sep=crcr]{%
0.001	7\\
0.02	20\\
0.2	    23\\
0.4	    19\\
0.6	    18\\
0.8	    16\\
1	    14 \\
};

\addplot [color=coolblack, mark = diamond, mark options={scale=0.8}, line width = 0.8]
  table[row sep=crcr]{%
0.001	4\\
0.02	9\\
0.2	 12\\
0.4	 12\\
0.6	  12\\
0.8	  12\\
1	  12 \\
};

\end{axis}

\end{tikzpicture}%

}
\subcaption{$\theta_E = 0.5$}
\label{subfig: theta_l_M40_K}
\end{minipage}
\begin{tikzpicture}
\begin{axis}[
    hide axis, 
    xmin=0, xmax=1, ymin=0, ymax=1,
    legend columns=3,
    legend style={at={(0.665, -0.01)}, anchor=north, draw=black, font=\scriptsize, /tikz/every even column/.append style={column sep=0.05cm}}
]
\addplot [color=napiergreen, mark = square, mark options={scale=0.6}, line width = 0.8]
coordinates {(-1,-1)};
\addlegendentry{$\calE = 0.2$, $\bar{\cal L} = 1 $}

\addplot [color=cerise, mark = pentagon, mark options={scale=0.8}, line width = 0.8]coordinates {(-1,-1)};
\addlegendentry{$\calE = 0.08$, $\bar{\cal L} = 1$}

\addplot [color=coolblack, mark = diamond, mark options={scale=0.8}, line width = 0.8]coordinates {(-1,-1)};
\addlegendentry{$\calE = 0.08$, $\bar{\cal L} = 0.5 $}

\end{axis}
\end{tikzpicture}

\caption{Performance analysis of FedAvg + EMQ with power allocation scheme in Algorithm~\ref{alg: FL+EMQ+power} vs. $\theta_E$ and $\theta_l$, non-IID data distribution for $M = 40$ clients.
}
\label{fig: theta_M40}
\end{figure}



\section{Conclusions and Future works}\label{Section: Conclusions}
This paper proposed an energy-efficient and low-latency framework that enables FL training over CFmMIMO networks with minimal resource needs. To achieve this, we introduced the EMQ quantization scheme, which dynamically allocates bits to each local gradient vector. Additionally, we formulated an FL training problem that optimizes the uplink powers and the number of local and global iterations during training. We then solved an optimization problem to determine the power allocation that minimizes a weighted sum of the straggler latency and total uplink energy consumption at each global iteration. Numerical results on IID and non-IID data distributions demonstrated that the proposed EMQ scheme, with an adaptive number of local iterations, achieved similar test accuracy to the full precision FL with the maximum number of local iterations, saving at least $49$\% of computational resources. Comparisons of the EMQ scheme with LAQ, AQUILA with Dinkelbach, and max-sum rate power allocation approaches further showed that, across all power allocation methods, EMQ outperformed AQUILA and LAQ by increasing test accuracy by up to $36$\% and $35$\%, respectively. Ultimately, FedAvg + EMQ with the proposed power allocation achieved the highest test accuracy within the same energy and latency budgets, outperforming FedAvg combined with LAQ, AQUILA with Dinkelbach, or max-sum rate methods.

For future work, we will focus on obtaining the optimal scalarization weights w.r.t. the total energy and latency budgets. Moreover, the optimal downlink power allocation to minimize downlink energy and latency, along with uplink powers, is one of the potential extensions of this paper.


\appendices

\section{Proof of Lemma~\ref{lemma: local_error}}\label{P:lemma: local_error}
Considering each element~$i = 1, \ldots, d$ of the vector~$\varepsilon_k^j$, we express it as
\begin{alignat}{3}
\nonumber
&\left[\varepsilon_k^j\right]_i = \left[\delta \bw_k^j\right]_i - \left[\delta \bq_k^j\right]_i = s_k^j \cdot \left( \Bar{\delta q}_{i,k}^j - [\Bar{\delta q}_{i,k}^j] \right) \cdot 10^{u_k^j}, \: \\
&\nrm{\varepsilon_k^j}_{\infty} = \max_{i \in [d]} \bigg| s_k^j \cdot \left( \Bar{\delta q}_{i,k}^j - [\Bar{\delta q}_{i,k}^j] \right) \cdot 10^{u_k^j} \bigg| = \: \\
\nonumber
& 10^{u_k^j} \cdot \max_{i \in [d]} \bigg| \Bar{\delta q}_{i,k}^j - [\Bar{\delta q}_{i,k}^j] \bigg| \leq 10^{u_k^j} \cdot 0.5 = 5 \cdot 10^{u_k^j-1} ,
\end{alignat}
where the last inequality arises from the fact that the absolute difference between any value~$a$ and its rounded value~$[a]$ is bounded by~$0.5$ because rounding maps $a$ to the nearest integer, which is at most 0.5 units away...

\section{Proof of Lemma~\ref{lemma: global_error}}\label{P:lemma: global_error}
We consider the update of~$\bw_k$ with EMQ quantization as
\begin{alignat}{3}
\nonumber
 & \bw_k = \bw_{k-1} + \sum_{j=1}^M \frac{\delta \bq_k^j}{M} =  \bw_{k-2} + \frac{1}{M} \sum_{j=1}^M \left( \delta \bq_k^j +  \delta \bq_{k-1}^j\right) \: \\
\nonumber
& =\bw_{0} + \sum_{k'=1}^k \sum_{j=1}^M \frac{\delta \bq_{k'}^j}{M} = \bw_{0} + \sum_{k'=1}^k \frac{1}{M}  \sum_{j=1}^M \left(\delta \bw_{k'}^j - \varepsilon_{k'}^j \right)\: \\
& =\underbrace{\bw_{0} + \sum_{k'=1}^k \frac{1}{M}  \sum_{j=1}^M \delta \bw_{k'}^j}_{:= \bw_k^F} - \sum_{k'=1}^k \frac{1}{M}  \sum_{j=1}^M \varepsilon_{k'}^j,
\end{alignat}
where~$\bw_k^F$ is the FedAvg global model with full precision. Defining~$u_{k'}^{\max}:= \max_{j} u_{k'}^j$, we have
\begin{alignat}{3}
  &\nrm{\bw_k}_{\infty} = \nrm{\bw_k^F - \sum_{k'=1}^k \frac{1}{M}  \sum_{j=1}^M \varepsilon_{k'}^j }_{\infty} \le  \nrm{\bw_k^F}_{\infty} + \: \\
 \nonumber
& \nrm{\sum_{k'=1}^k \frac{1}{M}  \sum_{j=1}^M \varepsilon_{k'}^j }_{\infty} \le  \nrm{\bw_k^F}_{\infty} + \sum_{k'=1}^k \frac{1}{M}  \sum_{j=1}^M \nrm{\varepsilon_{k'}^j}_{\infty} \le \: \\
 \nonumber
& \nrm{\bw_k^F}_{\infty} + \sum_{k'=1}^k M \cdot \frac{1}{M} \cdot 0.5\cdot 10^{u_{k'}^{\max}}.
\end{alignat}

\section{Proof of Proposition~\ref{prop: lkj and ukj}}\label{P:prop: lkj and ukj}
Defining~$\Delta \bw_{l,k}^j:= \bar{\Delta}\bw_{l,k}^j \times 10^{u_{l,k}^j} = \bw_{l,k}^j - \bw_{l-1,k}^j $, and\footnote{For simplifying the notation and reducing the space, we eliminate the Hadamard product.}~$\bg_{l,k+1}^j~\odot~\bg_{l,k+1}^j:=(\bg_{l,k+1}^j)^2$,  we have 
\begin{equation}\label{eq: DW-lemma}
   \bar{\Delta}\bw_{l,k}^j \times 10^{u_{l,k}^j} =  \frac{- \alpha~\bg_{l,k}^j}{\sqrt{\rho^l E_0^2 + (1 - \rho) \sum_{i=0}^{l-1}\rho^i \left( \bg_{l-i,k}^j \right)^2 } }, 
\end{equation}
where~$u_{l,k}^j:= \floor{ \log_{10} \|\Delta \bw_{l,k}^j \|_{\infty}}$. Similarly, we define~$\bg_{l,k}^j:= \bar{\bg}_{l,k}^j \times 10^{{v}_{l,k}^j}$, and re-write~\eqref{eq: DW-lemma} as
\begin{alignat}{3}\label{eq: DW-lemma1}
   \left(\bar{\Delta}\bw_{l,k}^j \right)^2 &\times 10^{2 u_{l,k}^j} =  ~{\frac{10~\alpha^2 ~\left(\bar{\bg}_{l,k}^j\right)^2 \times 10^{2{v}_{l,k}^j-1} }{\rho^l E_0^2 + (1 - \rho) \sum_{i=0}^{l-1}\rho^i \left( \bg_{l-i,k}^j \right)^2 } }\: \\
 \nonumber
 & \le~{\frac{ ~\left(\bar{\bg}_{l,k}^j\right)^2 \times 10^{2{v}_{l,k}^j-1} }{ \left(\bar{\bg}_{l,k}^j\right)^2 \times 10^{2{v}_{l,k}^j-1} + \sum_{i=1}^{l-1}10^{2{v}_{l-i,k}^j-1-i} } }.  
\end{alignat}
Since the absolute value of every element of~$\bar{\Delta}\bw_{l,k}^j$ and $\bar{\bg}_{l,k}^j$ are between~$1$ and~$10$, we have
\begin{alignat}{3}\label{eq: DW-lemma2}
   10^{2u_{l,k}^j} \le {\frac{10^{2{v}_{l,k}^j-1} }{ 10^{2{v}_{l,k}^j-1} + \sum_{i=1}^{l-1}10^{2{v}_{l-i,k}^j-3-i} } }.
\end{alignat}
 Then, we apply~$\log_{10}(.)$ to both sides of~\eqref{eq: DW-lemma2} and obtain
 \begin{alignat}{3}\label{eq: DW-lemma3}
   {u_{l,k}^j} \le -\frac{1}{2}\log_{10}\left( 1 + \sum_{i=1}^{l-1}10^{~2{v}_{l-i,k}^j -2{v}_{l,k}^j-2-i} \right).
\end{alignat}
Since the function~$f_j$, for all $j \in [M]$ is~$\bar{L}$-smooth (see Assumption~2), we apply the monotone mapping in~\cite{boydcnvx} on the local gradients~$\bg_{l,k}^j$ as
\begin{equation}\label{eq: monotone mapping}
  \left( \bg_{l,k}^j - \bg_{l-1,k}^j  \right)^\top  \left( \bw_{l,k}^j - \bw_{l-1,k}^j  \right) \ge 0,
\end{equation}
where
\begin{equation*}
  \bw_{l,k}^j - \bw_{l-1,k}^j = -\frac{\alpha \bg_{l-1,k}^j}{\sqrt{\rho E_{l-2}^2 + (1-\rho)(\bg_{l-1,k}^j)^2 }},
\end{equation*}
and replace it in~\eqref{eq: monotone mapping} to obtain
\begin{alignat}{3}\label{eq: monotone mapping2}
 & \left( \bg_{l,k}^j - \bg_{l-1,k}^j  \right)^\top  \left( \bw_{l,k}^j - \bw_{l-1,k}^j  \right) =  \: \\
 \nonumber
 & \left( \bg_{l-1,k}^j - \bg_{l,k}^j  \right)^\top \frac{\alpha \bg_{l-1,k}^j}{\sqrt{\rho E_{l-2}^2 + (1-\rho)(\bg_{l-1,k}^j)^2 }} \ge 0. 
\end{alignat}
We simplify~\eqref{eq: monotone mapping2}, considering~$\alpha > 0$, and obtain
\begin{alignat}{3}\label{eq: monotone mapping3}
  & \left( \bg_{l-1,k}^j - \bg_{l,k}^j  \right)^\top \bg_{l-1,k}^j \ge 0,   \: \\
 \nonumber
 & \left\| \bg_{l-1,k}^j \right\|_2^2  \ge \left(\bg_{l,k}^j\right)^\top \bg_{l-1,k}^j.
\end{alignat}
Replacing~$\bg_{l,k}^j= \bar{\bg}_{l,k}^j \times 10^{{v}_{l,k}^j}$, and $\bg_{l-1,k}^j= \bar{\bg}_{l-1,k}^j \times 10^{{v}_{l-1,k}^j}$ in~\eqref{eq: monotone mapping3}, we obtain
\begin{alignat}{3}\label{eq: monotone mapping4}
  & 10^{{v}_{l-1,k}^j - {v}_{l,k}^j} \ge \frac{\left(\bar{\bg}_{l,k}^j\right)^\top \bar{\bg}_{l-1,k}^j}{\left\| \bar{\bg}_{l-1,k}^j \right\|_2^2 } := \bar{v}_{l,k}^j \ge \bar{v}^j ,   \: \\
 \nonumber
& {v}_{l,k}^j \le {v}_{l-1,k}^j - \log_{10} \left(\bar{v}^j\right).
\end{alignat}
Considering~\eqref{eq: monotone mapping3}, we recall the bound for~$u_{l,k}^j$ in~\eqref{eq: DW-lemma3} 
 \begin{alignat}{3}\label{eq: DW-lemma4}
   {u_{l,k}^j} & \le -\frac{1}{2}\log_{10}\left( 1 + \sum_{i=1}^{l-1}10^{~2{v}_{l-i,k}^j -2{v}_{l,k}^j-2-i} \right)\: \\
 \nonumber
& \le  -\frac{1}{2}\log_{10}\left( 1 + \frac{10^{2\bar{v}^j-2}}{9} \left( 1 - (0.1)^l \right) \right),
\end{alignat}
which is a decreasing function of~$l$. Thus, for a sufficiently large~$l_{K^{\max}}$ and by considering a set of non-decreasing~$l_1^j < l_2^j < \ldots < l_k^j < \ldots \le l_{K^{\max}}$, 
the expression on the right side of~\eqref{eq: DW-lemma4} decreases, and we obtain that~$u_{l_k^j,k}^j = u_k^j \le u_{l_{k-1}^j,k-1}^j = u_{k-1}^j$.

\section{Proof of Proposition~\ref{prop: Fed_AdaDelta}}\label{P: prop: Fed_AdaDelta}
Considering that~$f$ is $\bar{L}$-smooth, defining~$E_l^2:=E[\bg^2]_{l,k}^j$, we have
\begin{alignat}{3}\label{eq: bound}
&f(\bw_{k+1}) \le f(\bw_k) + \frac{\bar{L}}{2}\nrm{\bw_{k+1} - \bw_k}^2 \: \\
 \nonumber
 & + \inner{\n f(\bw_k), \bw_{k+1} - \bw_k } \: \\
 \nonumber
 & = f(\bw_k) + \frac{\alpha^2 \bar{L}}{2M^2}\nrm{\sum_{j=1}^M \sum_{l=1}^L - \frac{\bg_{l,k+1}^j}{\sqrt{\rho E_{l-1}^2 + (1-\rho)(\bg_{l,k+1}^j)^2} }}^2  \: \\
 \nonumber
 & + \frac{\alpha}{M}\inner{\n f(\bw_k), \sum_{j=1}^M \sum_{l=1}^L - \frac{\bg_{l,k+1}^j}{\sqrt{\rho E_{l-1}^2 + (1-\rho)(\bg_{l,k+1}^j)^2 }  }  }.
\end{alignat}
Considering that $\nrm{\sum_{i=1}^n~\ba_i}^2~\le~n~\sum_{i=1}^n~\nrm{\ba_i}^2$, for any vector~$\ba_i$, we have
\begin{alignat}{3}\label{eq: norm_ws}
&\frac{\alpha^2 \bar{L}}{2M^2}\nrm{\sum_{j=1}^M \sum_{l=1}^L - \frac{\bg_{l,k+1}^j}{\sqrt{\rho E_{l-1}^2 + (1-\rho)(\bg_{l,k+1}^j)^2} }}^2 \le \: \\
 \nonumber
 &\frac{M \alpha^2 \bar{L}}{2M^2} \sum_{j=1}^M \nrm{\sum_{l=1}^L \frac{\bg_{l,k+1}^j}{\sqrt{\rho E_{l-1}^2 + (1-\rho)(\bg_{l,k+1}^j)^2} }}^2 \le \: \\
 \nonumber
 &\frac{ L\alpha^2 \bar{L}}{2M} \sum_{j=1}^M \sum_{l=1}^L\nrm{\frac{\bg_{l,k+1}^j}{\sqrt{\rho E_{l-1}^2 + (1-\rho)(\bg_{l,k+1}^j)^2} }}^2 \le \: \\
 \nonumber
 &\frac{ L\alpha^2 \bar{L}}{2M}  \sum_{j=1}^M \sum_{l=1}^L \sum_{i=1}^d \left[ \frac{\bg_{l,k+1}^j}{\sqrt{\rho E_{l-1}^2 + (1-\rho)(\bg_{l,k+1}^j)^2} } \right]_i^2 = \: \\
 \nonumber
 &\frac{ L\alpha^2 \bar{L}}{2M}  \sum_{j=1}^M \sum_{l=1}^L \sum_{i=1}^d  \frac{\left[\bg_{l,k+1}^j\right]_i^2}{{\left[\rho E_{l-1}^2 + (1-\rho)(\bg_{l,k+1}^j)^2\right]_i}}. 
\end{alignat}
Since~$\rho~\in(0,1)$, each element of the term~$\rho E_{l-1}^2 = \rho( \rho E_{l-2}^2 + (1-\rho)(\bg_{l-1,k+1}^j)^2) = (1-\rho)\sum_{i=1}^{l-1} \rho^i (\bg_{l-i, k+1})^2 + \rho^lE_0^2\sim\mathcal{O}(\rho^{2})$, thus~$\mathcal{O}\left(\rho E_{l-1}^2\right) < \mathcal{O}\left(((1-\rho)(\bg_{l,k+1}^j)^2\right)$, and we simplify the inequality~\eqref{eq: norm_ws} as
\begin{alignat}{3}\label{eq: norm_ws1}
\nonumber
&\frac{ L\alpha^2 \bar{L}}{2M}  \sum_{j=1}^M \sum_{l=1}^L \sum_{i=1}^d  \frac{\left[\bg_{l,k+1}^j\right]_i^2}{{\left[\rho E_{l-1}^2 + (1-\rho)(\bg_{l,k+1}^j)^2\right]_i}} \le  \: \\
 \nonumber
 &  \frac{ L\alpha^2 \bar{L}}{2M}  \sum_{j=1}^M \sum_{l=1}^L \sum_{i=1}^d  \frac{\left[\bg_{l,k+1}^j\right]_i^2}{{(1-\rho) \left|\left[\bg_{l,k+1}^j\right]_i^2 \right|}} \le  \: \\
 &  \frac{ L\alpha^2 \bar{L}}{2M(1-\rho)} M L d = \frac{ \alpha^2 L^2\bar{L} d }{2(1-\rho)}.
\end{alignat}
Using~\eqref{eq: norm_ws1}, we re-write~\eqref{eq: bound} with~$\E$ w.r.t. the global iteration~$k+1$ from both sides of the inequality as
\begin{alignat}{3}\label{eq: E_bound}
&\E\left\{ f(\bw_{k+1}) \right\} \le f(\bw_k) +  \frac{ \alpha^2 L^2\bar{L} d }{2(1-\rho)} +  \: \\
 \nonumber
 & \frac{\alpha}{M} \E \left\{ \! \inner{\! \n f(\bw_k), \sum_{j=1}^M \sum_{l=1}^L \frac{-\bg_{l,k+1}^j}{\sqrt{\rho E_{l-1}^2 + (1-\rho)(\bg_{l,k+1}^j)^2 }  } \!   } \! \right\}.
\end{alignat}
Next, we define~$\n_k:=  \n f(\bw_k) = ({1}/{M}) \sum_{j=1}^M \bg_k^j$, where~$\bg_k^j := \n_k^j(L)$, and focus on the last term in inequality~\eqref{eq: E_bound}, we have
\begin{alignat}{3}\label{eq: bound_inner1}
  &  \frac{\alpha}{M} \inner{\n f(\bw_k), \sum_{j=1}^M \sum_{l=1}^L \frac{-\bg_{l,k+1}^j}{\sqrt{\rho E_{l-1}^2 + (1-\rho)(\bg_{l,k+1}^j)^2 }  }   }  =  \: \\
 \nonumber
 &  \inner{\! \!  \n_k, \alpha L(\n_k -\n_k) -\sum_{j=1}^M \sum_{l=1}^L  \frac{ (\alpha/M) \bg_{l,k+1}^j }{\sqrt{\rho E_{l-1}^2 + (1-\rho)(\bg_{l,k+1}^j)^2 }  }  \! \!} \: \\
 \nonumber
 & = \inner{\n_k, - \alpha L \n_k} + \: \\
 \nonumber
 &  \inner{\n_k, \alpha L \n_k  - \frac{\alpha}{M} \sum_{j=1}^M \sum_{l=1}^L  \frac{ \bg_{l,k+1}^j }{\sqrt{\rho E_{l-1}^2 + (1-\rho)(\bg_{l,k+1}^j)^2 } } \! } \: \\
 \nonumber
 & = - \alpha L \nrm{\n_k}^2 +\: \\
 \nonumber
 & \! \inner{\n_k, \!  \frac{\alpha}{M} \sum_{j=1}^M \sum_{l=1}^L \bg_k^j - \frac{ \bg_{l,k+1}^j }{\sqrt{\rho E_{l-1}^2 + (1-\rho)(\bg_{l,k+1}^j)^2 }  }  }.
\end{alignat}
According to Cauchy–Schwarz inequality,~$\inner{\ba,\bb}\le\nrm{\ba}\nrm{\bb}$, we re-write the last inner product in~\eqref{eq: bound_inner1} as
\begin{alignat}{3}\label{eq: inner_Cauchy–Schwarz}
 &  \inner{\n_k,  \frac{\alpha}{M} \sum_{j=1}^M \sum_{l=1}^L \bg_k^j - \frac{ \bg_{l,k+1}^j }{\sqrt{\rho E_{l-1}^2 + (1-\rho)(\bg_{l,k+1}^j)^2 }  }  } \le \: \\
 \nonumber
 & \nrm{\n_k} \nrm{\frac{\alpha}{M} \sum_{j=1}^M \sum_{l=1}^L \bg_k^j - \frac{ \bg_{l,k+1}^j }{\sqrt{\rho E_{l-1}^2 + (1-\rho)(\bg_{l,k+1}^j)^2 }  }}.
\end{alignat}
Afterward, we use the inequality of~$ x y \le (x^2+y^2)/2$ for~\eqref{eq: inner_Cauchy–Schwarz}
\begin{alignat}{3}\label{eq: inner_simplify}
& \nrm{\n_k} \nrm{\frac{\alpha}{M} \sum_{j=1}^M \sum_{l=1}^L \bg_k^j - \frac{ \bg_{l,k+1}^j }{\sqrt{\rho E_{l-1}^2 + (1-\rho)(\bg_{l,k+1}^j)^2 }  }} \le \: \\
 \nonumber
 & \frac{\nrm{\n_k}^2}{2} + \frac{\alpha^2}{2M^2}\nrm{\sum_{j=1}^M \sum_{l=1}^L \bg_k^j - \frac{ \bg_{l,k+1}^j }{\sqrt{\rho E_{l-1}^2 + (1-\rho)(\bg_{l,k+1}^j)^2 } }}^2.
\end{alignat}
By inserting~\eqref{eq: inner_simplify} into~\eqref{eq: E_bound}, we have
\begin{alignat}{3}\label{eq: bound_inner2}
\nonumber
 &\E\left\{ f(\bw_{k+1}) \right\} \le f(\bw_k) +  \frac{ \alpha^2 L^2\bar{L} d }{2(1-\rho)} +\: \\
 \nonumber
 & \left(\frac{1}{2}-\alpha L\right) \E \left\{ \nrm{ \n_k }^2 \right\} +  \: \\
 \nonumber
 & \frac{\alpha^2}{2M^2} \E \left\{\nrm{\sum_{j=1}^M \sum_{l=1}^L \bg_k^j - \frac{ \bg_{l,k+1}^j }{\sqrt{\rho E_{l-1}^2 + (1-\rho)(\bg_{l,k+1}^j)^2 } }}^2\right\}\: \\
 & \le f(\bw_k) +  \frac{ \alpha^2 L^2\bar{L} d }{2(1-\rho)} + \left(\frac{1}{2}-\alpha L\right) \E \left\{ \nrm{ \n_k }^2 \right\} +  \: \\
 \nonumber
 & \frac{\alpha^2 M L }{2M^2}\E \! \left\{\sum_{j=1}^M \sum_{l=1}^L \nrm{\bg_k^j - \frac{ \bg_{l,k+1}^j }{\sqrt{\rho E_{l-1}^2 + (1-\rho)(\bg_{l,k+1}^j)^2 } }}^2 \!  \right\},
\end{alignat}
where the last inequality results from~$\E\left\{\nrm{z_1+\ldots+z_r}^2 \right\} \le r\E\left\{\nrm{z_1}^2+\ldots+\nrm{z_r}^2 \right\}$, which~$z_i,~{i~\in~[r]}$ are random variables. We simplify the last term in~\eqref{eq: bound_inner2} as
\begin{alignat}{3}\label{eq: bound_inner3}
 \nonumber
& \E \nrm{\bg_k^j \!-\! \frac{ \bg_{l,k+1}^j }{\sqrt{\rho E_{l-1}^2 \!+\! (1\!-\!\rho)(\bg_{l,k+1}^j)^2 } }}^2 = \: \\
 \nonumber
 & \E \nrm{\bg_k^j\!-\!\bg_{l,k+1}^j \!+\!\bg_{l,k+1}^j \!-\! \frac{ \bg_{l,k+1}^j }{\sqrt{\rho E_{l-1}^2 \!+\! (1\!-\!\rho)(\bg_{l,k+1}^j)^2 } }}^2 \: \\
 & \le 2\E \left\{ \nrm{\bg_k^j -\bg_{l,k+1}^j}^2 \right\} +  \: \\
 \nonumber
 &  2\E \left\{ \nrm{\bg_{l,k+1}^j\left(1 - \frac{1}{\sqrt{\rho E_{l-1}^2 + (1-\rho)(\bg_{l,k+1}^j)^2 } }  \right) }^2 \right\}. 
\end{alignat}
Afterward, we use the \textbf{Assumption~2}, Lipschitz Gradient, in~\eqref{eq: bound_inner3} and obtain
\begin{alignat}{3}\label{eq: bound_inner4}
\nonumber
 & \E\left\{ f(\bw_{k+1}) \right\} \le f(\bw_k) +  \frac{ \alpha^2 L^2\bar{L} d }{2(1-\rho)} \: \\
 \nonumber
 & + \left(\frac{1}{2}-\alpha L\right) \E \left\{ \nrm{ \n_k }^2 \right\} +  \: \\
 \nonumber
 & \frac{\alpha^2 L \bar{L}^2}{M} \E \left\{\sum_{j=1}^M \sum_{l=1}^L \nrm{\bw_{l,k+1}^j - \bw_k}^2\right\}+ \frac{\alpha^2 L \bar{L}^2}{M} \cdot \sum_{j=1}^M \sum_{l=1}^L \: \\
 \nonumber
 &  \E \left\{ \nrm{\bg_{l,k+1}^j\left(1 - \frac{1}{\sqrt{\rho E_{l-1}^2 + (1-\rho)(\bg_{l,k+1}^j)^2 } }  \right) }^2 \right\} \le \: \\
 \nonumber
 & f(\bw_k) + \left(\frac{1}{2}-\alpha L\right) \E \left\{ \nrm{ \n_k }^2 \right\} +  \sum_{l=1}^L \alpha^2 L \bar{L}^2 d G^2 T(l) \: \\
 & + \frac{\alpha^2 L \bar{L}^2}{M} \E \left\{\sum_{j=1}^M \sum_{l=1}^L \nrm{\bw_{l,k+1}^j - \bw_k}^2\right\} + \frac{ \alpha^2 L^2\bar{L} d }{2(1-\rho)}, 
\end{alignat}
where~$\bw_{l,k+1}^j$ is the local FL parameter at global iteration~$k+1$ after~$l$ local iterations. Using~\textbf{Assumption~5}, we obtain~$T(l)$ as
\begin{equation}\label{eq: T(l)}
    T(l) := \left(1-\frac{1}{\sqrt{G^2\left(1 -\rho^{l} \right) + \rho^lE_0^2}} \right)^2.
\end{equation}
Next, we focus on~$\E \left\{\sum_{j=1}^M \sum_{l=1}^L \nrm{\bw_{l,k+1}^j - \bw_k}^2\right\}$, and we start from
\begin{alignat}{3}\label{eq: bound_ws0}
\nonumber
    & \bw_{l,k+1}^j - \bw_k  \stackrel{ \text{According to~} \eqref{eq: w_lkj_update}}{=}   \: \\
 \nonumber
 & \bw_{l-1,k+1}^j -\frac{\alpha g_{l-1,k+1}^j}{\sqrt{\rho E_{l-2}^2 + (1-\rho)(\bg_{l-1,k+1}^j)^2 }} - \bw_k =  \: \\
 \nonumber
 & \bw_{l-1,k+1}^j - \bw_k -\frac{\alpha \left(g_{l-1,k+1}^j - \n_{l-1,k+1}^j \right)}{\sqrt{\rho E_{l-2}^2 + (1-\rho)(\bg_{l-1,k+1}^j)^2 }} +  \: \\
 \nonumber
 &\alpha \n_{l-1,k+1}^j -\frac{\alpha \n_{l-1,k+1}^j }{\sqrt{\rho E_{l-2}^2 + (1-\rho)(\bg_{l-1,k+1}^j)^2 }} - \alpha \n_{l-1,k+1}^j   \: \\
 &+ \alpha \n_k^j - \alpha \n_k^j + \alpha \n_k - \alpha \n_k,
\end{alignat}
where $\n_{l,k+1}^j$ is the true local gradient over the whole local dataset at~$\bw_{l,k+1}^j$ in client~$j$, and~$g_{l,k+1}^j$ is the unbiased estimator of it, see~\textbf{Assumption~1}. Moreover,~$\n_k^j$ is the true local gradient of client~$j$ at~$\bw_k$. Then, we take the expectation over~\eqref{eq: bound_ws0} as
\begin{alignat}{3}\label{eq: bound_ws1}
\nonumber
 & \E \left\{ \nrm{\bw_{l,k+1}^j - \bw_k }^2 \right\} \le  6\E \left\{ \nrm{\bw_{l-1,k+1}^j - \bw_k }^2 \right\} +  \: \\
 \nonumber
 & 6 \alpha^2 \E \left\{ \nrm{ \frac{ g_{l-1,k+1}^j - \n_{l-1,k+1}^j }{\sqrt{\rho E_{l-2}^2 + (1-\rho)(\bg_{l-1,k+1}^j)^2 }} }^2 \right\} +   \: \\
 \nonumber
 & 6 \alpha^2 T(l-1)\E \left\{ \nrm{\n_{l-1,k+1}^j }^2 \right\}  +    \: \\
 \nonumber
 & 6 \alpha^2 \E \left\{ \nrm{\n_k^j - \n_{l-1,k+1}^j }^2 \right\} + \: \\
 & 6 \alpha^2 \E \left\{ \nrm{ \n_k - \n_k^j }^2 \right\} +6 \alpha^2 \E \left\{ \nrm{  \n_k }^2 \right\}.
\end{alignat}
We simplify~\eqref{eq: bound_ws1} utilizing~\textbf{Assumptions~2-5}, and obtain
\begin{alignat}{3}\label{eq: bound_ws2}
\nonumber
 & \E \left\{ \nrm{\bw_{l,k+1}^j - \bw_k }^2 \right\} \le  6\E \left\{ \nrm{\bw_{l-1,k+1}^j - \bw_k }^2 \right\} +  \: \\
 \nonumber
 & \frac{6 \alpha^2}{G^2(1-\rho)} \E \left\{\sum_{i=1}^d \sigma_{l_i}^2\right\} + 6 \alpha^2 T(l-1)\E \left\{ \nrm{ \n_{l-1,k+1}^j }^2 \right\}+  \: \\
 \nonumber
 & 6 \bar{L}^2 \alpha^2 \E \left\{ \nrm{ \bw_{l-1,k+1}^j - \bw_k }^2 \right\}+   \: \\
 & 6 \alpha^2 \E \left\{ \nrm{ \n_k - \n_k^j }^2 \right\} +6 \alpha^2 \E \left\{ \nrm{  \n_k }^2 \right\}.
\end{alignat}
Next, we take the summations over~all $j$ and~$l$ as
\begin{alignat}{3}\label{eq: bound_ws3}
 & \E \left\{\sum_{j=1}^M \sum_{l=1}^L \nrm{\bw_{l,k+1}^j - \bw_k }^2 \right\} \le \: \\
 \nonumber
 &  (6 + 6 \bar{L}^2 \alpha^2  )\sum_{j=1}^M \sum_{l=1}^L\E \left\{ \nrm{\bw_{l-1,k+1}^j - \bw_k }^2 \right\}  \: \\
 \nonumber
 & +\frac{6 \alpha^2 M L}{G^2(1-\rho)} \E \left\{\sum_{i=1}^d \sigma_{l_i}^2\right\} \: \\
 \nonumber
 & + 6 \alpha^2 \sum_{j=1}^M \sum_{l=1}^L T(l-1) \E \left\{ \nrm{ \n_{l-1,k+1}^j }^2 \right\}  \: \\
 \nonumber
 & + 6 \alpha^2 L \E \left\{\sum_{j=1}^M  \nrm{ \n_k - \n_k^j }^2 \right\} +6 \alpha^2 L M \E \left\{ \nrm{ \n_k }^2 \right\},
\end{alignat}
where after applying~\textbf{Assumption~4} and multiplying~$1/M$ into~\eqref{eq: bound_ws3}, we have
\begin{alignat}{3}\label{eq: bound_ws4}
 &\frac{1}{M} \E \left\{\sum_{j=1}^M \sum_{l=1}^L \nrm{\bw_{l,k+1}^j - \bw_k }^2 \right\} \le \: \\
 \nonumber
 &  \frac{1}{M}(6 + 6\bar{L}^2 \alpha^2  )\sum_{j=1}^M \sum_{l=1}^L\E \left\{ \nrm{\bw_{l-1,k+1}^j - \bw_k }^2 \right\} +  \: \\
 \nonumber
 & \frac{6\alpha^2 L} {G^2(1-\rho)} \E \left\{\sum_{i=1}^d \sigma_{l_i}^2\right\} + \: \\
 \nonumber
 &\frac{6 \alpha^2}{M} \sum_{j=1}^M \sum_{l=1}^L T(l-1) \E \left\{ \nrm{ \n_{l-1,k+1}^j }^2 \right\}  \: \\
 \nonumber
 & + 6 \alpha^2 L \E \left\{\sum_{i=1}^d \sigma_{g_i}^2 \right\} +6 \alpha^2 L \E \left\{ \nrm{ \n_k }^2 \right\}.
\end{alignat}
We consider~$\sum_{l=1}^L T(l)$ and $\sum_{l=1}^L T(l-1)$ as
\begin{alignat}{3}\label{eq: T(l)_simp}
\nonumber
    & \sum_{l=1}^L T(l) = \: \\
 \nonumber
 &\sum_{l=1}^L 1 + \frac{1}{G^2\left(1 -\rho^{l} \right) + \rho^lE_0^2} -\frac{2}{\sqrt{G^2\left(1 -\rho^{l} \right) + \rho^lE_0^2}} \le  \: \\
 &\sum_{l=1}^L 1 + \frac{1}{G^2\left(1 -\rho^{l} \right) + \rho^lE_0^2} \le L + \frac{L}{G^2(1-\rho^L)}, \: \\
 & \sum_{l=1}^L T(l-1) \le L + \frac{L}{G^2(1-\rho^{L-1})}.
\end{alignat}
Next, we simplify~\eqref{eq: bound_ws4} as
\begin{alignat}{3}\label{eq: bound_ws5}
 &\frac{1}{M} \sum_{j=1}^M \E \left\{\sum_{l=1}^L \nrm{\bw_{l,k+1}^j - \bw_k }^2 \right\} \le \: \\
 \nonumber
 & \left(6 + 6\bar{L}^2 \alpha^2 \right)  \frac{1}{M} \sum_{j=1}^M\E \left\{  \sum_{l=1}^L\nrm{\bw_{l-1,k+1}^j - \bw_k }^2 \right\} +  \: \\
 \nonumber
 & 6\alpha^2 L \sum_{i=1}^d \E \left\{ \frac{\sigma_{l_i}^2} {G^2(1-\rho)} +  \sigma_{g_i}^2 \right\} + \: \\
 \nonumber
 &6 \alpha^2 L d G^2\left(1 + \frac{1}{G^2(1-\rho^{L-1})}\right) + 6 \alpha^2 L \E \left\{ \nrm{ \n_k }^2 \right\}.
\end{alignat}
We define the series of~$X_l:= (1/M) \sum_{j=1}^M \E \left\{\sum_{l=1}^L \nrm{\bw_{l,k+1}^j - \bw_k }^2 \right\} $ and replace in~\eqref{eq: bound_ws5}, we have
\begin{alignat}{3}\label{eq: bounds_series0}
\nonumber
 & X_l \le \left(6 + 6\bar{L}^2 \alpha^2 \right)  X_{l-1} + 6\alpha^2 L \sum_{i=1}^d \E \left\{ \frac{\sigma_{l_i}^2} {G^2(1-\rho)} +  \sigma_{g_i}^2 \right\} \: \\
 \nonumber
 & +6 \alpha^2 L d G^2\left(1 + \frac{1}{G^2(1-\rho^{L-1})}\right) + 6 \alpha^2 L \E \left\{ \nrm{ \n_k }^2 \right\} = \: \\
 \nonumber
 & \left(6 + 6\bar{L}^2 \alpha^2 \right) X_{l-1} + \bar{B} \le \left(6 + 6\bar{L}^2 \alpha^2 \right)^2 X_{l-2}  +\: \\
 \nonumber
 &  \bar{B} \left(1+ \left(6 + 6\bar{L}^2 \alpha^2 \right) \right) \le  \left(6 + 6\bar{L}^2 \alpha^2 \right)^l X_{0} +  \: \\
 &\bar{B} \sum_{s=0}^{l-1}\left(6 + 6\bar{L}^2 \alpha^2 \right)^s,
\end{alignat}
where~$X_0 = 0$ since~$\bw_{0,k+1}^j = \bw_k$, and
\begin{alignat}{3}\label{eq: bar_B}
&\bar{B}:=  6\alpha^2 L \sum_{i=1}^d \E \left\{ \frac{\sigma_{l_i}^2} {G^2(1-\rho)} +  \sigma_{g_i}^2 \right\} \: \\
 \nonumber
 & +6 \alpha^2 L d G^2\left(1 + \frac{1}{G^2(1-\rho^{L-1})}\right) + 6 \alpha^2 L \E \left\{ \nrm{ \n_k }^2 \right\}.    
\end{alignat}
Considering that~$\alpha \le \left(6 L \bar{L}^2 \right)^{-0.5}$, we obtain~$6+6\bar{L}^2~\alpha^2 \le 6 + 1/L$, and for any~$1 \le l \le L$,
\begin{alignat}{3}\label{eq: sum_bar_B}
& \bar{B} \sum_{s=0}^{l-1}\left(6 + 6\bar{L}^2 \alpha^2 \right)^s \le \bar{B} L \left( 6 + \frac{1}{L}\right)^L \le \bar{B} L \exp(6).
\end{alignat}
Now, we consider~\eqref{eq: bound_inner4}, we have
\begin{alignat}{3}\label{eq: bound_kogether0}
\nonumber
 & \E\left\{ f(\bw_{k+1}) \right\} \le   f(\bw_k) + \left(\frac{1}{2}-\alpha L\right) \E \left\{ \nrm{ \n_k }^2 \right\} + \: \\
 \nonumber
 &  \alpha^2 L \bar{L}^2 d G^2 \left(L + \frac{L}{G^2(1-\rho^L)} \right) + \frac{ \alpha^2 L^2\bar{L} d }{2(1-\rho)} \: \\
 & + \frac{\alpha^2 L^2 \bar{L}^2 \exp(6)}{M} \bar{B},   
\end{alignat}
where we replace~$\bar{B}$ from~\eqref{eq: bar_B} into~\eqref{eq: bound_kogether0}, and obtain
\begin{alignat}{3}\label{eq: bound_kogether1}
\nonumber
 & \E\left\{ f(\bw_{k+1}) \right\} \le   f(\bw_k) + \left(\frac{1}{2}-\alpha L\right) \E \left\{ \nrm{ \n_k }^2 \right\}  \: \\
 \nonumber
 & + \alpha^2 L \bar{L}^2 d G^2 \left(L + \frac{L}{G^2(1-\rho^L)} \right) + \frac{ \alpha^2 L^2\bar{L} d }{2(1-\rho)} \: \\
 & + \frac{6\alpha^4 L^3 \bar{L}^2 \exp(6)}{M}\sum_{i=1}^d \E \left\{ \frac{\sigma_{l_i}^2} {G^2(1-\rho)} +  \sigma_{g_i}^2 \right\} \: \\
 \nonumber
 & + \frac{6\alpha^4 L^3 d G^2 \bar{L}^2 \exp(6)}{M} \left(1 + \frac{1}{G^2(1-\rho^{L-1})}\right) \: \\
 \nonumber
 & +  \frac{6 \alpha^4 L^3 \bar{L}^2 \exp(6)}{M} \E \left\{ \nrm{ \n_k }^2 \right\}  \stackrel{\eqref{eq: alpha0}}{\le} \: \\
 \nonumber
 & f(\bw_k) + \frac{1}{k+1}\left( 1 + G^2 + \sigma_k^2 + \frac{1}{(1-\rho^{L-1})} \right).
\end{alignat}

\bibliographystyle{./MetaFiles/IEEEtran}
\bibliography{./MetaFiles/References}
\end{document}
***********